\documentclass[11pt]{article}

\usepackage{amsmath}
\usepackage{amssymb}
\usepackage{booktabs}

\usepackage{graphicx}
\usepackage{xcolor}
\usepackage{caption}
\usepackage[letterpaper,margin=1in]{geometry}
\usepackage[hidelinks]{hyperref}

\renewenvironment{abstract}
	{\quotation}
	{\endquotation}

\date{}

\makeatletter
\renewcommand{\fnum@figure}{\textbf{Figure \thefigure}}
\renewcommand{\fnum@table}{\textbf{Table \thetable}}
\makeatother

\usepackage{scicite}

\usepackage{url}
\usepackage{cleveref}
\usepackage{array}
\usepackage{makecell}

\def\scititle{
	Partial recovery of meter-scale surface weather
}
\title{\bfseries \boldmath \scititle}

\author{
	Jonathan~Giezendanner$^{1\dagger}$,
	Qidong~Yang$^{1\dagger}$,
	Ruizhe~Huang$^{1\dagger}$,\and
    Eric~Schmitt$^{2}$,
    Anirban~Chandra$^{2}$,
    Yawen~Zhang$^{3}$,
    Jeremy~Vila$^{2}$,\and
    Detlef~Hohl$^{2}$,
    Campbell~Watson$^{4}$,
    Sherrie~Wang$^{1\ast}$\\[0.6em]
	\small$^{1}$Massachusetts Institute of Technology\\
    \small$^{2}$Shell Information Technology\\
    \small$^{3}$Allen Institute for AI\\
    \small$^{4}$IBM Research\\
	\small$^\ast$Corresponding author. Email: sherwang@mit.edu\and
	\small$^\dagger$These authors contributed equally to this work.
}

\begin{document} 

\maketitle

\begin{abstract} 
Near-surface weather varies over tens to hundreds of meters, yet remains unresolved in analyses and forecasts. We test whether this variation can be inferred without resolving atmospheric dynamics. Combining sparse weather stations, high-resolution Earth observation, and coarse atmospheric dynamics, we infer temperature, dewpoint, and wind at 30-m resolution across the contiguous United States. Against measurements held out in space and time, estimates reduce error by 11--28\% relative to the strongest baseline. Within held-out $0.25^\circ$ grid cells, we recover more spatial variance than baselines, explaining nearly half of temperature variability in the median cell. The method captures time-varying differences between locations and produces coherent patterns associated with topography and land cover. Beyond weather, our findings illustrate how sparse observations of a dynamical system can be combined with dense observations of persistent environmental structure to recover otherwise unresolved spatial variability.
\end{abstract}


\noindent

\newpage 

\section{Introduction}

\begin{figure}
\begin{center}
\vspace{-40pt}
\includegraphics[width=1\textwidth]{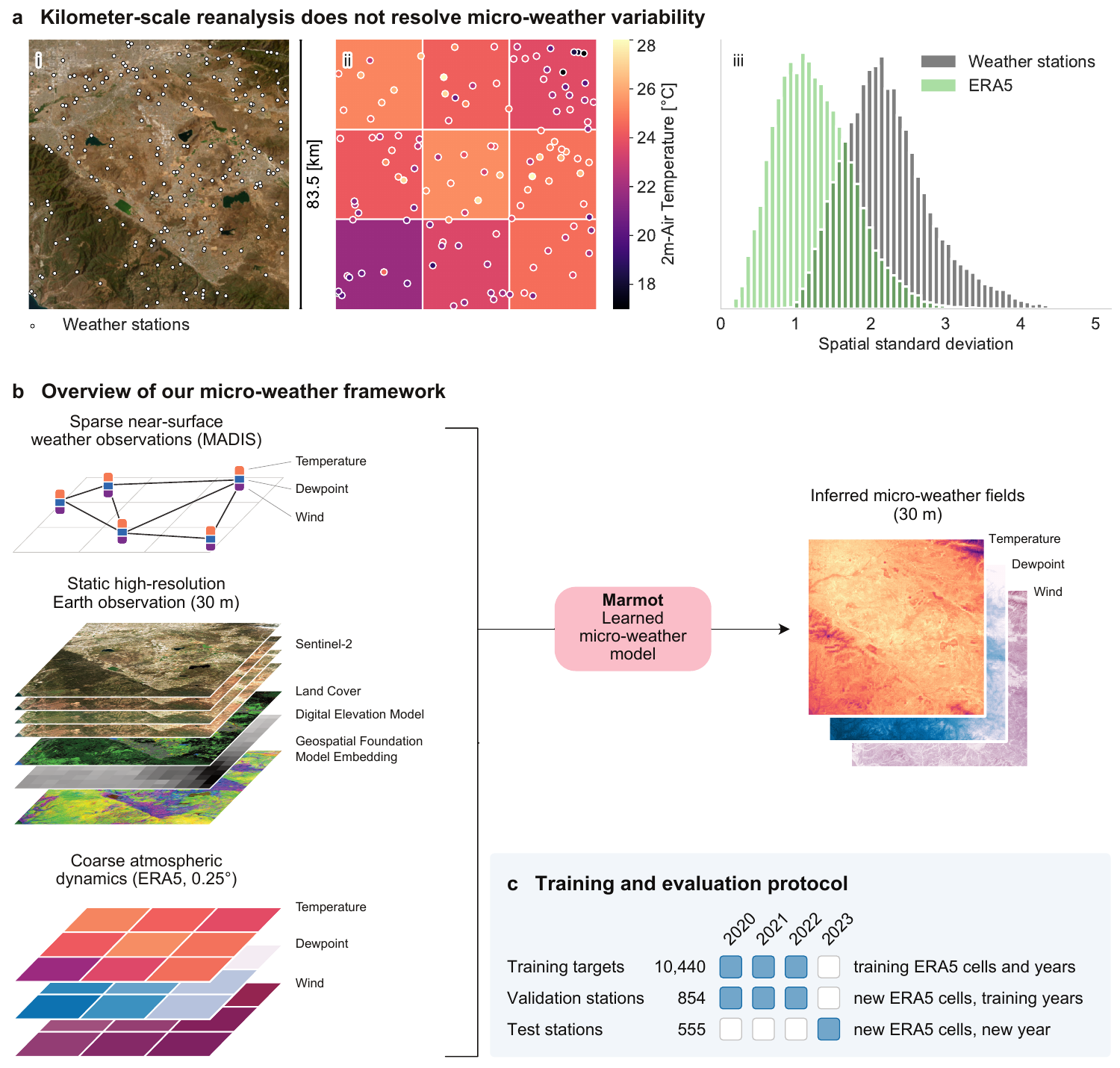}
\vspace{-35pt}
\end{center}
\caption{
    \textbf{Framework combining surface stations, Earth observation, and reanalysis to recover sub-kilometer weather variability.}
    \textbf{(a)} (i) Example over the Los Angeles metropolitan area illustrating that kilometer-scale reanalysis (ERA5) does not resolve meter-scale weather variability. (ii) 2 m air temperature from ERA5 (grid) and as measured by surface weather stations (points). (iii) Histogram of the spatial standard deviation of 2 m air temperature, 2020--2023, across the weather stations versus across the nine ERA5 grid cells.
    \textbf{(b)} Overview of the inference framework. The model combines observations from sparse surface stations, high-resolution Earth observation data, and coarse atmospheric dynamics from ERA5 to generate spatially continuous 30 m resolution fields of 2 m air temperature, dewpoint temperature, and 10 m wind.
    \textbf{(c)} Training and evaluation protocol. Stations are split into backbone, training, validation, and test sets by whole 0.25\textdegree{} ERA5 cells; backbone and training stations serve as training targets. Filled squares mark the years each set contains; test stations are new to the model in both space and time.
}
\label{fig:Methods:OverviewCombined}
\end{figure}

Weather at the Earth's surface can change across meter scales: think of stepping from a shaded park into a sunlit street or breaking above the treeline into exposed wind. These small-scale variations --- which we here refer to as ``micro-weather'' --- are created by terrain, vegetation, buildings, and other surface features, and in turn impact human heat exposure \cite{wang2025dual, li2023divergent}, renewable energy performance \cite{giebel2011wind}, wildfire propagation \cite{eghdami2025sensitivity, cruz2019fire}, and pollutant dispersion \cite{baklanov2009air}. Despite their importance, micro-weather variations are not captured by numerical weather predictions (NWP) and reanalysis products at regional or global scales~\cite{giorgi1991, rummukainen2010, chandler2020, schar2020, jijon2021}, since NWP, along with their recent machine-learning emulators~\cite{lam2023, kurth2023fourcastnet, nguyen2023climax, bi2023accurate, bodnar2024, schmude2024, price2025, alet2025}, are run at grid spacings of several kilometers (Fig. \ref{fig:Methods:OverviewCombined}a). This coarse spatial resolution is likely to persist for some time as models continue to contend with computational limits and the breakdown of parameterizations at high resolution \cite{satoh2019GCRMs, prein2017challenges, wedi2020baseline}. At the time of writing, for example, the emerging digital twin initiative at ECMWF aims to simulate the atmosphere at $\sim$1 km resolution \cite{ECMWF2025Strategy, wedi2020baseline}.

Yet resolving fine-scale atmospheric dynamics may not be necessary to recover all fine-scale weather variability. Near-surface conditions are shaped not only by turbulent atmospheric processes but also by systematic interactions between the atmospheric state and persistent surface characteristics such as topography, vegetation, and buildings. This raises the question: how much weather variability below the resolution of current models can be inferred from existing observations without explicitly resolving fine-scale atmospheric dynamics? We refer to this component as the ``statistically recoverable component of micro-weather''.

Our ability to address this question is enabled by the fact that Earth observation now resolves the land surface at meter to tens-of-meter resolution globally. Although these images do not directly encode the dynamic atmospheric state, multispectral satellite imagery, land-cover classifications, and digital elevation models capture properties that shape micro-scale atmospheric behavior \cite{drusch2012sentinel2,yang2018new,farr2007srtm}. For example, wind must maneuver around terrain, vegetation has a cooling effect, and surface materials differ in moisture retention.

We show that a substantial portion of micro-weather is statistically recoverable by combining sparse surface observations with land surface characteristics and ERA5 reanalysis. We train \textbf{Marmot}, a multimodal deep learning model, to predict observations at more than 10,000 weather stations across the contiguous United States. Then, at inference time, we generate spatially continuous near-surface micro-weather fields at 30 m resolution. We evaluate Marmot against millions of observations in entirely held-out ERA5 grid cells and find that it reduces error relative to statistical and dynamical baselines, including station interpolation, ERA5-MOS, and HRRR.
To distinguish sub-grid recovery from improvements in aggregate accuracy, we develop an evaluation framework based on observed spatial variance within coarse grid cells and contrasts between held-out station pairs as a function of their separation. Under these tests, Marmot recovers substantially more sub-grid variation than the baseline methods, with evidence of temperature recovery in some sub-kilometer station pairs.
These recovered contrasts change through time, including reversals in which location is warmer or more humid, rather than reflecting persistent differences between locations. Improvements also extend into the tails of the weather distribution, with particularly strong performance for extreme heat and humidity.

Despite being trained on sparse station measurements, the inferred fields also exhibit physically interpretable and coherent patterns at meter scales, including nighttime temperature inversions, humidity gradients due to crop irrigation, and wind channeling with terrain. Independent comparisons with land surface temperature observations from the ECOSTRESS sensor lend further support to the spatial structure of the inferred air temperature fields \cite{fisher2020ecostress}. 
Ablation experiments reveal that nearby weather stations provide a strong estimate of conditions at unobserved locations, while explicit topographic relationships and high-resolution Earth observation enable additional spatial variation to be recovered; coarse reanalysis contributes comparatively modest additional information.
Together, these results show that near-surface micro-weather contains a component that can be inferred from existing observations without explicitly modeling fine-scale atmospheric dynamics. More broadly, our work demonstrates how sparse ground observations can be densified with fine-scale environmental information to recover previously unresolved variability.

\section{The micro-weather model}

We develop Marmot, a transformer-based model that infers near-surface weather at arbitrary target locations from three complementary sources of information  (Fig. \ref{fig:Methods:OverviewCombined}b). First, a backbone network of surface weather stations from the NOAA Meteorological Assimilation Data Ingest System (MADIS) provides spatially sparse, point-based observations of 2 m temperature, 2 m dewpoint, and 10 m wind. Second, high-resolution Earth observation (EO) data characterizes fine-scale surface heterogeneity at station and target locations. Our primary model uses 30 m AlphaEarth foundation model embeddings, which summarize information from multiple EO modalities \cite{brown2025}; experiments using conventional EO inputs are described in Sections~\ref{sec:methods:data} and \ref{sec:methods:ablation}.
Third, hourly ERA5 reanalysis describes how the coarse weather field at $0.25^{\circ}\times0.25^{\circ}$ resolution varies around the target location and how ERA5 differs from observations at nearby backbone stations \cite{hersbach2020}.

Marmot first uses self-attention to integrate weather information across the backbone station network and then uses cross-attention with the target location to infer target weather conditions. The model explicitly represents spatial relationships between backbone stations and the target, including relative position, distance, and elevation difference, allowing it to learn how weather varies with geographic and topographic separation. At the target location, Marmot additionally receives ERA5 weather and high-resolution EO information describing the local environment. Additional architecture and training details are provided in Sections~\ref{sec:methods:model} and \ref{sec:methods:training}.

We evaluate Marmot on observations that are held out both spatially and temporally (Fig. \ref{fig:Methods:OverviewCombined}c). To enforce spatial separation, ERA5 grid cells across the contiguous U.S. are partitioned into disjoint backbone, training, validation, and test sets (Fig. \ref{fig:Methods:stations}). Backbone stations provide contemporaneous weather context during both training and inference, while stations in training and backbone cells provide supervision during model training. Validation cells are used for model selection, and test cells are reserved exclusively for evaluation. Importantly, no weather observations from held-out stations are seen by the model at any time; predictions at test station locations rely solely on ERA5 and EO features at the target location and observations from the backbone network.
To simultaneously enforce temporal separation, the model is trained using observations from 2020--2022 and evaluated only on 2023 observations from these spatially held-out test cells. Thus, every test observation is from both a location and a time period unseen during training.

At inference time, Marmot can be queried at arbitrary locations and timesteps by combining the backbone station observations with EO features at the target location and coarse atmospheric information from ERA5. This enables the generation of spatially continuous micro-weather fields at 30 m resolution over large geographic domains. These fields estimate the component of micro-weather that is statistically recoverable from sparse weather observations, fine-scale environmental information, and coarse atmospheric context.

\begin{figure}
\begin{center}
\includegraphics[width=1\textwidth]{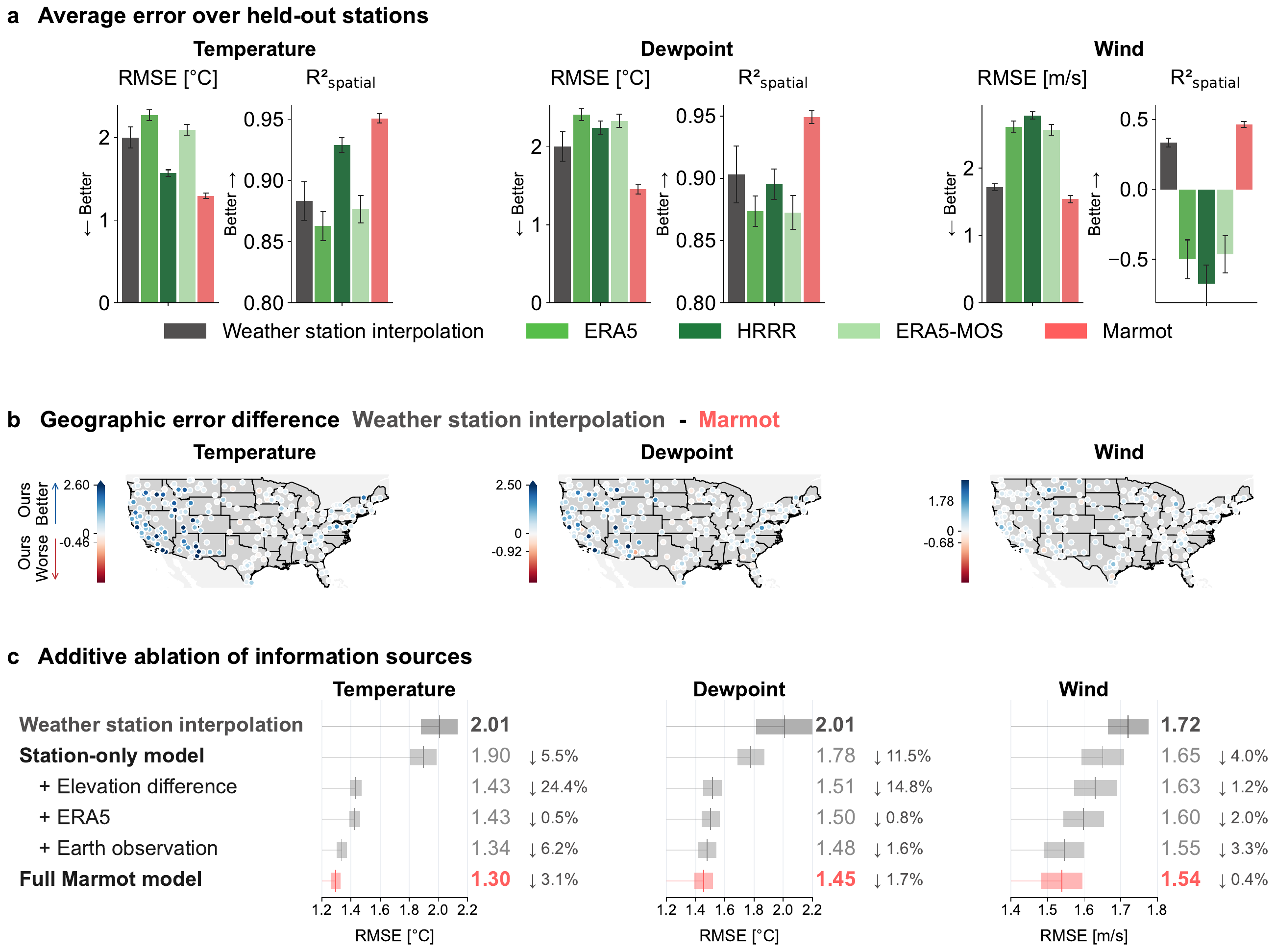}
\end{center}
\caption{
    \textbf{U.S.-wide evaluation of inferred micro-weather accuracy relative to baseline methods.}
    \textbf{(a)} Root mean square error (RMSE) and spatial coefficient of determination ($R_{\text{spatial}}^2$, de-meaned per timestep to isolate spatial variability) evaluated on 2023 observations from stations in spatially held-out test grid cells, such that all evaluation observations are held out both in space and time. Marmot is compared with station interpolation, ERA5 reanalysis, HRRR analysis, and ERA5-MOS. Error bars denote one standard deviation across bootstrap replicates, each resampling both the held-out test grid cells and the five trained model seeds.
    \textbf{(b)} Map of station-level error reduction by Marmot relative to station interpolation, averaged over 2023. Positive values indicate improvement over station interpolation.
    \textbf{(c)} Additive ablation of information sources in Marmot. Model specifications progressively add information to station interpolation; values report RMSE and the percent reduction in RMSE relative to the preceding specification.
}
\label{fig:results:overview}
\end{figure}

\section{Improved near-surface weather estimation}

We compare Marmot against four baselines chosen to represent distinct approaches to estimating local weather conditions: interpolation of station observations; globally-available ERA5 reanalysis, which provides Marmot’s coarse reanalysis input; ERA5-MOS \cite{benbouallegue2023statistical}, a statistical post-processing model trained on station observations and surface attributes; and HRRR \cite{dowell2022}, an operational regional weather analysis with 3 km grid spacing (Section~\ref{sec:methods:baselines}).

Across the contiguous U.S., Marmot consistently improves near-surface weather estimation relative to these baselines. Averaged over all held-out test stations and time steps, Marmot reduces 2 m air temperature RMSE by 18\% relative to HRRR analysis, the strongest temperature baseline, and reduces 2 m dewpoint and 10 m wind vector RMSE by 28\% and 11\%, respectively, relative to station interpolation, the strongest baseline for those variables (Fig. \ref{fig:results:overview}a). Relative to ERA5, which is provided as an input to the model, temperature RMSE is reduced by 43\%, dewpoint RMSE by 40\%, and wind vector RMSE by 41\%. Statistical post-processing improves ERA5 slightly, but ERA5-MOS remains less accurate than Marmot across all three variables.
Marmot also achieves higher national $R^2_{\text{spatial}}$ than baseline methods for all three variables (Fig. \ref{fig:results:overview}a), indicating that its improved pointwise accuracy arises not only from correction of systematic biases but also improved representation of spatial variability at fixed timesteps.

Performance gains are geographically widespread but heterogeneous (Fig. \ref{fig:results:overview}b). For temperature and dewpoint, improvements relative to station interpolation are largest across the western and eastern U.S. and smaller across much of the central U.S. For wind, error reductions are smaller and have less pronounced regional structure. The size of improvements also varies by land cover type: relative error reductions are largest for temperature and dewpoint in shrublands and for wind in forested and built-up areas, while the smallest reductions occur in cropland for temperature and dewpoint and in barren and shrubland areas for wind (Fig. \ref{fig:appendix:results:land_cover_error}). At the national scale, prediction error exhibits only a weakly positive relationship with distance to the nearest backbone stations (Fig. \ref{fig:results:error_distance}), suggesting that station proximity alone does not explain these patterns. The geographic and land-cover differences may instead reflect a combination of environmental heterogeneity, the information provided by the surrounding station network, and differences in the processes governing each weather variable.

We next examine which sources of information enable the model to estimate near-surface weather between sparse observations (Fig. \ref{fig:results:overview}c; Table \ref{tab:Appendix:Results:AblationComponents}). Station interpolation itself provides a strong baseline across all three variables. A learned station-only model, which infers conditions at held-out locations from contemporaneous observations at neighboring backbone stations, improves upon interpolation ($-7\%$ RMSE averaged across the three variables). Explicitly providing the elevation difference between neighboring stations and the target produces a large reduction in error ($-13\%$ RMSE), particularly for temperature and dewpoint. Adding ERA5 provides a small additional improvement ($-1\%$ RMSE), whereas incorporating high-resolution Earth observation through AlphaEarth embeddings produces a more substantial reduction in error ($-4\%$ RMSE). The remaining components of the full Marmot model, including temporal encodings, provide modest additional gains ($-2\%$ RMSE).

These results indicate that station observations provide substantial information about conditions at unobserved locations, while fine-scale information about topography and surface characteristics enables recovery of additional spatial variation. The small marginal improvement from ERA5 is consistent with coarse atmospheric information being largely redundant with observations from the backbone station network. Fine-scale surface information remains informative even after explicit elevation differences are provided: removing either elevation difference or AlphaEarth embeddings from the full model increases error (Table \ref{tab:Appendix:Results:AblationComponents}). Conventional Earth observation inputs, including elevation, land cover, and Sentinel-2 imagery, also improve predictions but do not reproduce the full gains obtained with AlphaEarth embeddings (Table \ref{tab:Appendix:Results:SurfaceInfo}).

\begin{figure}
\begin{center}
\includegraphics[width=1\textwidth]{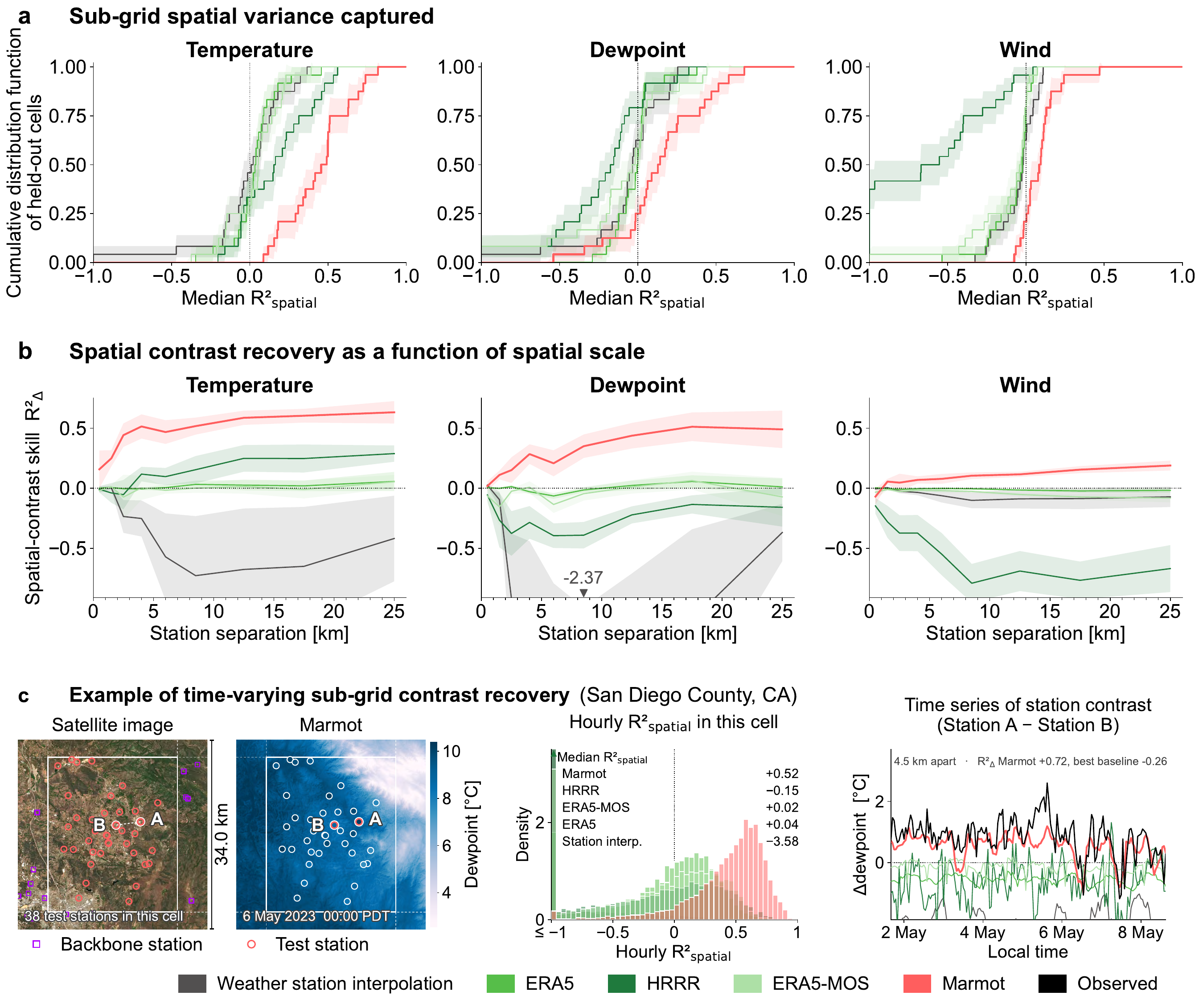}
\end{center}
\vspace{-0.5cm}
\caption{
    \textbf{Recovery of sub-grid spatial weather variability.}
    \textbf{(a)} Cumulative distribution functions (CDFs) of median $R^{2}_{\text{spatial}}$ across held-out ERA5 grid cells. For each grid cell and hourly timestep, $R^{2}_{\text{spatial}}$ measures the spatial variance captured by each method relative to observations across stations within the grid cell; values are then summarized per-cell by the median across hourly timesteps.
    \textbf{(b)} Recovery of spatial contrasts in near-surface weather as a function of spatial separation. Spatial contrast skill ($R^2_{\Delta}$) measures how well each method recovers observed differences between pairs of held-out stations and is shown as a function of the distance between stations.
    \textbf{(c)} Example of time-varying sub-grid contrast recovery in San Diego County, California. The left panels show satellite imagery and Marmot's inferred dewpoint field within a held-out ERA5 grid cell containing 38 test stations. The center panel shows the distribution of hourly \(R_{\text{spatial}}^2\) within the cell over 2023. The right panel compares the observed and predicted dewpoint contrast between two held-out stations 4.5 km apart; Marmot captures the changing contrast with \(R_\Delta^2=0.72\), compared with \(-0.26\) for the next-best baseline.
}
\label{fig:results:subgrid}
\end{figure}

\section{Recovery of sub-grid spatial weather variability}

While reductions in pointwise error and national $R^2_{\text{spatial}}$ indicate improved agreement with station observations, they do not establish that the model recovers weather at fine scales. 
To directly evaluate sub-grid spatial recovery, we measure Marmot's ability to capture observed spatial variance within entirely held-out ERA5 grid cells, in which no station observations are available to the model and the variance within each cell is unresolved by Marmot's coarse atmospheric input.

Using a de-meaned coefficient of determination $R^2_{\text{spatial}}$ that isolates spatial variability at fixed time steps (Section~\ref{sec:methods:evaluation}), we find that Marmot explains a significantly larger fraction of observed sub-grid spatial variance than baseline methods for all three weather variables (Fig. \ref{fig:results:subgrid}a). Performance is especially strong for temperature; for the median held-out ERA5 grid cell, Marmot explains 47\% of the spatial variance in temperature, compared to 16\% for HRRR, which is the next-best baseline. Dewpoint and wind variability is more challenging to recover, with Marmot achieving median $R^2_{\text{spatial}}$ of 13\% and 9\%, respectively. Nevertheless, this is an improvement upon baseline methods, which have median $R^2_{\mathrm{spatial}}$ at or below zero for both variables, indicating that their sub-grid predictions are, on average, no better than assigning all stations within a grid cell a constant value.
Notably, HRRR captures some sub-grid temperature variability, but it has the poorest performance for dewpoint and wind among the baselines.

Next, to determine how recovery varies with spatial scale, we evaluate contrasts between pairs of held-out stations as a function of their separation (Fig. \ref{fig:results:subgrid}b). Spatial contrast skill, \(R_\Delta^2\), measures recovery relative to predicting identical conditions at the two stations (Section~\ref{sec:methods:evaluation}). At separations of 1 km or greater, Marmot has the highest point estimate among the evaluated methods in every distance bin and for all three variables. In the 1--2 km bin, Marmot achieves \(R_\Delta^2\) of 25\%, 11\%, and 6\% for temperature, dewpoint, and wind, respectively. Below 1 km, the temperature point estimate remains positive (\(R_\Delta^2 = 16\% \)), but uncertainty is large because such closely spaced station pairs are rare ($n \le 13$); dewpoint and wind skill is near or below zero. Thus, contrast recovery generally declines at shorter separations and is substantially stronger for temperature than for dewpoint or wind.
Pair-level analyses reveal substantial variation in recovery across locations (Fig. \ref{fig:results:subgrid_contrast_pairs}).

We further examine sub-grid spatial recovery in example grid cells for each weather variable. In Warren County, New Jersey, Marmot captures 50\% of the observed within-cell spatial variance in temperature, compared with 15\% for HRRR and 16\% for ERA5-MOS (Fig. \ref{fig:results:subgrid_cases}). The observed temperature contrast between two stations 12 km apart varies through time: Station A is warmer than Station B during the day but cooler at night. Marmot reproduces much of this changing spatial contrast ($R^2_{\Delta} = 53\%$), whereas the baseline methods do not capture the temporal pattern. In San Diego County, California, the most densely observed test grid cell with 38 held-out stations, Marmot captures 52\% of the observed spatial variance in dewpoint, compared with 4\% for the next-best method, ERA5 interpolation (Fig. \ref{fig:results:subgrid}c). Marmot likewise recovers the changing dewpoint contrast between two stations 4.5 km apart ($R^2_{\Delta} = 72\%$, versus -26\% next-best). Wind remains more challenging: in Austin, Texas, Marmot captures little of the aggregate within-cell spatial variance (median $R^2_{\text{spatial}} = 3\%$), but recovers a substantial fraction of the time-varying wind-vector contrast between two stations ($R^2_{\Delta} = 41\%$, versus $-6\%$ next-best; Fig. \ref{fig:results:subgrid_cases}). Together, these examples show that Marmot does not simply learn persistent offsets between locations, but recovers differences that evolve dynamically through time.

\begin{figure}
\captionsetup{singlelinecheck=off}
\vspace{-1.7cm}
\centering
\includegraphics[width=0.9\textwidth]{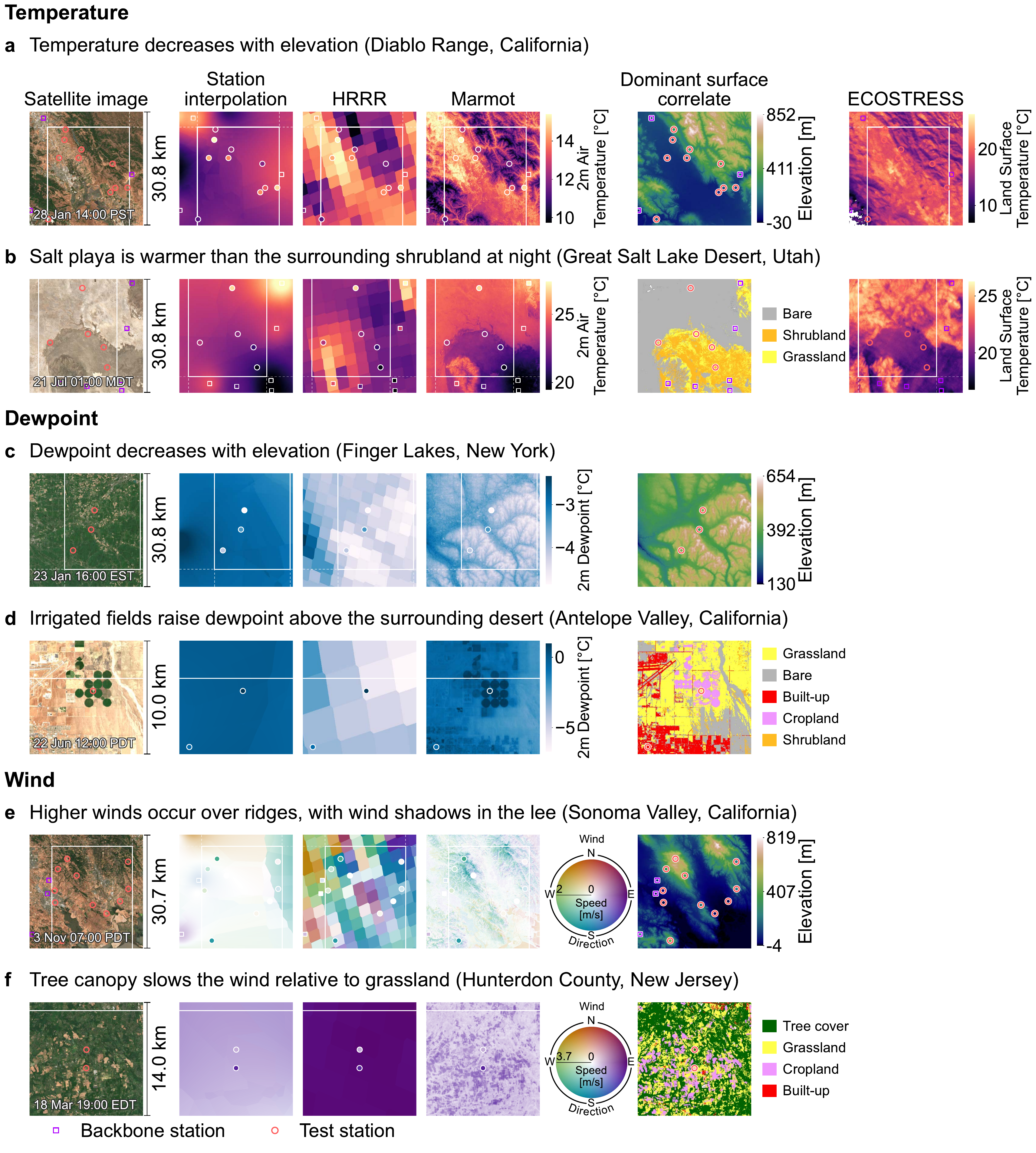}
\caption{
    \textbf{Physical structure of inferred micro-weather fields.}
    Case studies comparing inferred 30 m resolution fields with station interpolation and HRRR. Satellite imagery\protect\footnotemark\ is shown for context, and terrain elevation or land cover is shown according to the dominant surface correlate of inferred spatial variability. 
    For temperature panels, land surface temperature from ECOSTRESS is included for qualitative comparison.
    \textbf{(a)} Temperature decreases with elevation (Diablo Range, California).
    \textbf{(b)} Bare ground is warmer than the surrounding shrubland at night (Great Salt Lake Desert, Utah).
    \textbf{(c)} Dewpoint decreases with elevation (Finger Lakes, New York).
    \textbf{(d)} Irrigated cropland has higher humidity than the surrounding desert (Antelope Valley, California).
    \textbf{(e)} Higher winds occur over ridges; a southwesterly wind corresponds to windward-leeward contrasts in wind speed (Sonoma Valley, California).
    \textbf{(f)} Wind speed is higher over grass and cropland and lower over tree cover (Hunterdon County, New Jersey).
}\label{fig:Results:Inference:Phenomena}
\end{figure}

\section{Physical structure of recovered micro-weather}

Having established that Marmot recovers sub-grid spatial variability, we next examine whether the inferred fields exhibit physically coherent structure (Fig. \ref{fig:Results:Inference:Phenomena}).
For a variety of case studies, we validate the learned spatial structure against held-out stations and examine the surface variable --- either elevation or land cover --- associated most strongly with Marmot's predicted field (Table \ref{tab:Appendix:Results:SurfaceCorrelates}; Section \ref{sec:methods:evaluation}).

Marmot learns to recover the lapse rate effect. For instance, in the complex terrain of the Diablo Range surrounding Gilroy, California (Fig. \ref{fig:Results:Inference:Phenomena}a), inferred temperatures decrease with elevation over short horizontal distances. Although nearby backbone stations are located at low elevations, Marmot recovers cooler temperatures at high elevations, which is validated by held-out stations. HRRR captures similar broad spatial patterns but, at 3-km resolution, smooths over the narrow warmer valleys between peaks. Elevation explains 44\% of the spatial variance in Marmot's inferred temperature field, compared with 20\% for land cover (Table \ref{tab:Appendix:Results:SurfaceCorrelates}).

Land cover is more strongly associated with the inferred temperature structure in the Great Salt Lake Desert, Utah (Fig. \ref{fig:Results:Inference:Phenomena}b). At night, Marmot predicts warmer air temperatures over the barren salt playa and cooler temperatures over the surrounding shrubland, with temperature contrasts reaching several degrees Celsius and consistent with differences observed between held-out stations. 
An independent, contemporaneous ECOSTRESS land surface temperature (LST) observation exhibits the same fine-scale spatial pattern. Although LST and 2-m air temperature are distinct quantities, previous work has shown that they are strongly coupled at night and that nighttime LST captures spatial variation in air temperature associated with land cover \cite{good2016lst, oyler2016remotely}. 
HRRR, by contrast, predicts the opposite spatial contrast, with cooler temperatures over the playa than the shrubland.

For dewpoint, Marmot similarly recovers fine-scale gradients associated with topography and vegetation.
In the hilly terrain of the Finger Lakes, New York (Fig. \ref{fig:Results:Inference:Phenomena}c), lower dewpoint values are inferred at higher elevations. 
These gradients are smoothed in HRRR and cannot be recovered through station interpolation given the sparsity of observations.
In the arid Antelope Valley, California (Fig. \ref{fig:Results:Inference:Phenomena}d), Marmot delineates higher dewpoint over irrigated crop fields relative to the surrounding desert. 
HRRR appears to under-predict dewpoint across the broader desert region and does not resolve the more humid crop fields.

\footnotetext{All the high resolution image sources used for contextualization in the document were taken from Tiles \textcopyright\ Esri -- Source: Esri, i-cubed, USDA, USGS, AEX, GeoEye, Getmapping, Aerogrid, IGN, IGP, UPR-EGP, and the GIS User Community}

Lastly, fine-scale wind variability is more challenging to recover, but the inferred fields nevertheless exhibit coherent structure in both wind speed and direction.
In the Sonoma Valley region of California (Fig. \ref{fig:Results:Inference:Phenomena}e), predicted wind speed varies with exposure, with higher wind speeds along ridgelines and lower speeds in valleys. Under predominantly southwesterly winds, Marmot predicts higher speeds on the southwestern sides of ridges and lower speeds on the northeastern sides. Despite the lack of a nearby backbone station, it also correctly predicts a shift to northerly winds observed at a held-out station approaching San Pablo Bay.
HRRR exhibits much more varied wind directions across the region and over-predicts wind speeds.
In the varied land cover of central New Jersey (Fig. \ref{fig:Results:Inference:Phenomena}f), the inferred field shows lower wind speeds over forested areas and higher speeds over grassland and cropland. These patterns are consistent with different surface roughness and drag coefficients across land cover types. 
Again, neither station interpolation nor HRRR can recover these patterns due to sparsity and coarseness, respectively. 

We also find physically interpretable structure associated with temperature inversions, urban heat, and terrain-channeled wind across additional regions (Figs. \ref{fig:results:casestudy:sonoma_inversion} to \ref{fig:results:wind_channeling}). At the same time, Marmot does not recover micro-weather structure equally well across all conditions. Recovery is less reliable when nearby backbone observations provide limited information about conditions at the target location (Fig. \ref{fig:results:error_distance}). This occurs in some remote locations, where the nearest backbone stations are tens of kilometers away and experience very different conditions, and in nearshore settings, where backbone stations lie predominantly landward and poorly represent marine-influenced conditions (Figs. \ref{fig:results:casestudy:failureNearestStationFar} and \ref{fig:results:casestudy:failureCostalOneSided}).

\section{Performance under extreme weather}

\begin{figure}
\begin{center}
\includegraphics[width=1\textwidth]{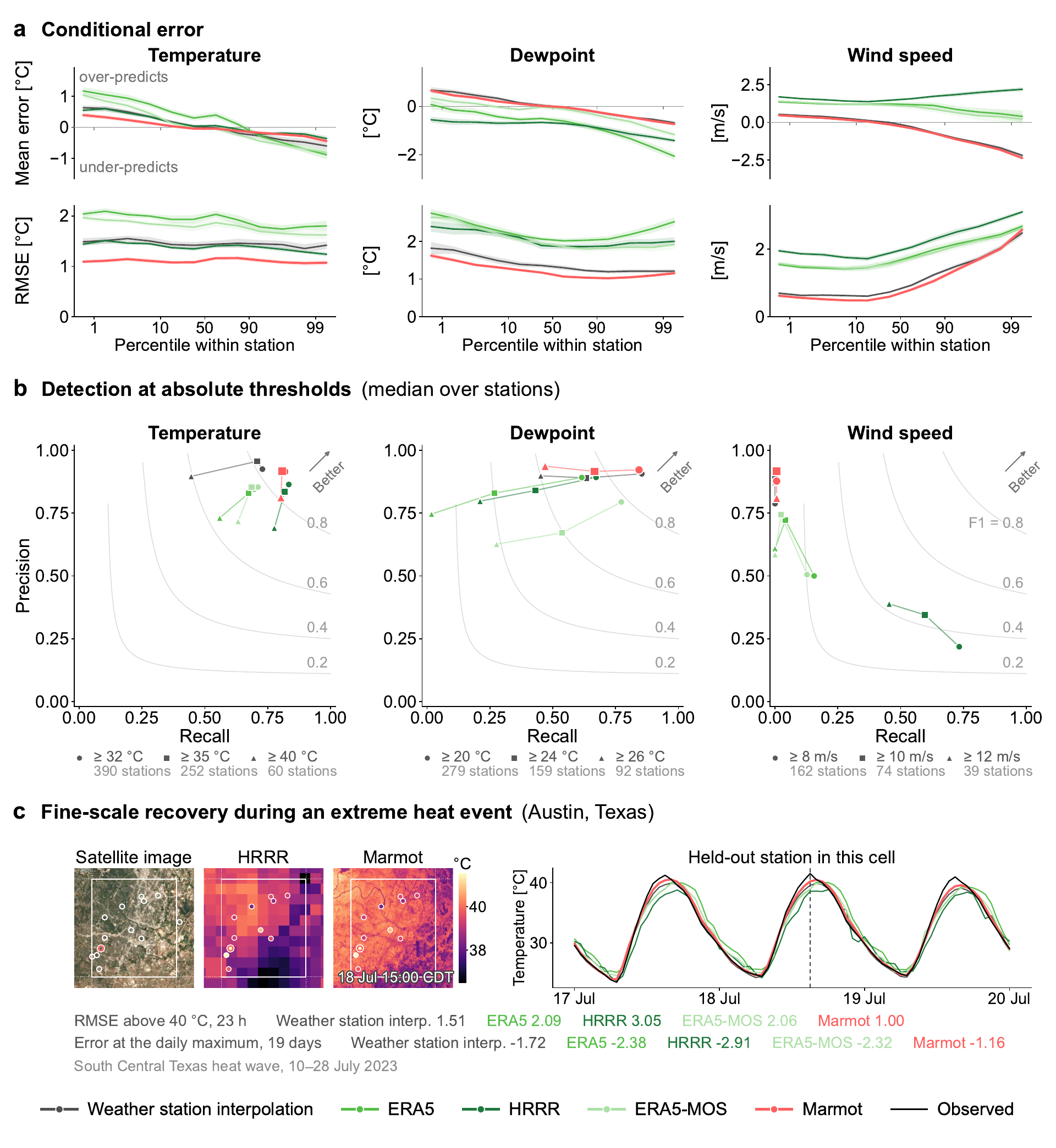}
\end{center}
\vspace{-0.5cm}
\caption{
    \textbf{Performance under extreme weather.}
    \textbf{(a)} Conditional mean error and RMSE as a function of observed percentile within each held-out station. 
    \textbf{(b)} Precision and recall for detection of temperature, dewpoint, and wind speed extremes at fixed absolute thresholds, summarized by the median across stations. Points denote different thresholds; gray curves indicate lines of constant F1 score. 
    \textbf{(c)} Fine-scale temperature recovery during the July 2023 South Central Texas heat wave. Maps compare HRRR and Marmot with held-out station observations; the time series shows observed and predicted temperature at a held-out station.
}
\label{fig:results:extrema}
\end{figure}

We next ask whether Marmot's improvements extend to extreme and rapidly changing weather, which have disproportionate impacts on human health and infrastructure despite their relative rarity. We evaluate performance across the tails of the observed weather distribution, at absolute extreme thresholds, and during rapid changes in weather (Section~\ref{sec:methods:evaluation}).

We first bin observations by their percentile within each held-out station's local distribution and compute conditional bias and RMSE within each bin. 
All three statistical methods --- station interpolation, ERA5-MOS, and Marmot --- exhibit regression toward the mean: low extremes are systematically over-predicted and high extremes under-predicted (Fig. \ref{fig:results:extrema}a). However, Marmot exhibits the least severe over- and under-prediction for temperature, and among the least severe for dewpoint and wind. Temperature estimates from ERA5 and HRRR also regress toward the mean, but their errors for dewpoint and wind are dominated by overall negative and positive biases, respectively.
Conditional RMSE is quite different across variables. For all methods, temperature error remains nearly uniform across the distribution, whereas dewpoint error increases toward the lowest percentiles and wind-speed error rises sharply toward the highest percentiles. Marmot achieves the lowest error for all three variables at nearly every percentile; the one exception is the highest wind-speed bin, where station interpolation is marginally better.

We next evaluate detection of extreme heat, humidity, and wind at fixed absolute thresholds (Fig. \ref{fig:results:extrema}b). For all methods, detection becomes more difficult at increasingly extreme thresholds, with precision and/or recall declining as events become rarer. Nevertheless, Marmot lies on the precision-recall Pareto frontier for temperature and dewpoint across all thresholds, with its largest advantages over the baselines at the most extreme thresholds. For temperatures above $40\,^\circ\mathrm{C}$, for example, Marmot achieves an F1 score of $0.81$, compared with $0.73$ for the next-best method, HRRR. A similar advantage is observed for dewpoints above $26\,^\circ\mathrm{C}$.

The South Central Texas heat wave of July 2023 illustrates this advantage  (Fig. \ref{fig:results:extrema}c). All stations within the example grid cell are held out from Marmot. During temperatures above $40\,^\circ\mathrm{C}$, Marmot achieves an RMSE of $1.00\,^\circ\mathrm{C}$, compared to station interpolation at $1.51\,^\circ\mathrm{C}$ and HRRR at $3.05\,^\circ\mathrm{C}$. 
All methods underestimate the daily temperature maximum, but Marmot has the smallest mean bias ($-1.16\,^\circ\mathrm{C}$ versus $-1.72\,^\circ\mathrm{C}$ for station interpolation and $-2.91\,^\circ\mathrm{C}$ for HRRR). The time series further shows that Marmot more closely tracks the magnitude and timing of the daily temperature peaks.

Wind is a different story. HRRR has by far the highest F1 score for detecting high wind events, with relatively high recall but low precision. Marmot, station interpolation, and ERA5-MOS instead have high precision but very low recall. In other words, Marmot rarely predicts extreme winds when they do not occur, but misses many observed high-wind events.

Finally, we evaluate rapid increases and decreases over three-hour intervals (Fig. \ref{fig:appendix:results:ramps}). Ramp detection is more challenging than detection at absolute thresholds, especially for dewpoint and wind, although the relative behavior of the methods is similar. Marmot lies on the precision-recall Pareto frontier for temperature and dewpoint ramps. Wind ramps are difficult for all methods, with HRRR achieving the highest F1 scores through a similar high-recall, low-precision tradeoff.

\section{Discussion}

Our results demonstrate that a substantial component of near-surface micro-weather can be statistically recovered without explicitly resolving fine-scale atmospheric dynamics. Across the contiguous U.S., Marmot improves estimation of temperature, dewpoint, and wind relative to both statistical and dynamical baselines, and, critically, recovers observed spatial variance within ERA5 grid cells and between nearby stations. This recovery extends to time-varying weather contrasts between locations and to extreme heat and humidity. The inferred high-resolution fields exhibit physically coherent relationships with topography and land-surface characteristics, while the ablation shows that nearby station observations provide a strong foundation for inference and that elevation and high-resolution Earth observation contribute substantial additional information.

These results suggest a useful way to conceptualize the statistically recoverable component of micro-weather. Weather stations sparsely observe the instantaneous atmospheric state, while topography and Earth observation provide spatially dense information about how locations differ from one another. This combination allows spatial structure in near-surface weather to be inferred between observations. For example, stations may reveal that bare ground is warmer than nearby shrubland under a particular atmospheric state, while continuous Earth observation allows the same relationship to be applied at unobserved locations with similar surface characteristics. Although our analyses do not directly establish the learned mechanism, the Great Salt Lake Desert example (Fig. \ref{fig:Results:Inference:Phenomena}b), among others, is consistent with this interpretation. Coarse reanalysis provides additional atmospheric context, but its modest marginal contribution in our ablations suggests large redundancy with station observations for the present inference task. We note that this differs from forecasting, where future station observations are unavailable and dynamical forecasts must instead provide information about the evolving atmospheric state.

The extent of statistical recovery varies substantially by variable, spatial scale, and region. Temperature contrasts remain more recoverable than dewpoint or wind contrasts across spatial scales (Fig. \ref{fig:results:subgrid}b), but pair-level performance varies considerably across locations (Fig. \ref{fig:results:subgrid_contrast_pairs}b). At sub-kilometer separations, aggregate temperature-contrast skill remains positive but uncertain, whereas dewpoint and wind skill approaches zero.
Rapid changes in dewpoint and wind are similarly challenging, and wind direction remains difficult even though Marmot improves over the baselines (Table \ref{tab:Appendix:Results:SR2}). Geographic gains are also heterogeneous, with smaller improvements across parts of the central U.S. and in some sparsely observed or relatively homogeneous environments. These patterns may reflect limited station information, insufficient environmental context, or variability generated by fine-scale atmospheric dynamics not represented in the model. We see these limitations in additional case studies, such as in terrain-confined valleys where Marmot only partially recovers observed wind channeling (Fig. \ref{fig:results:wind_channeling}), as well as in remote and nearshore locations, where nearby backbone stations poorly represent conditions at the target location (Figs. \ref{fig:results:casestudy:failureNearestStationFar} and \ref{fig:results:casestudy:failureCostalOneSided}).

The comparison with HRRR provides further evidence that dynamical modeling is important for some of this remaining variability. HRRR resolves atmospheric dynamics at substantially finer scales than ERA5, whereas Marmot infers spatial variation from contemporaneous observations and environmental information. Their differing performance is particularly informative for wind. Marmot estimates average wind conditions more accurately and recovers some surface-associated spatial structure, but HRRR detects substantially more extreme winds and rapid wind changes. This suggests that higher-resolution dynamical modeling contains information about rapidly evolving weather that is absent from Marmot's current inputs. 
Conversely, Marmot could complement existing NWP diagnostics by revealing where regional models fail to resolve surface-associated spatial structure.
To maximize micro-weather recovery, future work could apply a Marmot-type model to higher-resolution analyses or forecasts like HRRR.

The evaluation framework developed here separates aggregate accuracy from recovery of spatial variance within coarse grid cells. Its within-cell and pairwise metrics test whether finer-scale statistical or dynamical products recover observed sub-grid structure rather than merely reduce bias or increase nominal resolution. Validation at the shortest spatial scales nevertheless remains limited by the scarcity of closely spaced observations. Only 13 station pairs are available below 1 km, and their performance varies substantially across locations. Evidence approaching the 30 m grid spacing is therefore indirect, based on consistency with known physical relationships and, in selected temperature examples, independent ECOSTRESS land surface temperature observations. Denser networks and targeted field campaigns will be necessary to quantify recovery at these scales more comprehensively.

The ability to reconstruct near-surface micro-weather from existing observing systems could improve characterization of weather exposure in applications ranging from urban heat and wildfire behavior to wind energy and pollutant transport. More broadly, many Earth system processes are observed through a similar combination of sparse measurements of dynamic states and spatially dense observations of environmental structure. Our results show that combining information in these observing systems can reveal variability far below the resolution at which the dynamic state is directly observed or modeled.

\section{Methods}

\subsection{Data}
\label{sec:methods:data}

\paragraph{Point-based Weather Observations}

We obtained point-based weather observations from the NOAA Meteorological Assimilation Data Ingest System (MADIS), which aggregates observations from governmental, academic, private, and volunteer station networks \cite{miller2005madis}. We used the LDAD Mesonet source and retained 2 m temperature, 2 m dewpoint, and 10 m wind speed and direction.

In this work, we focus on stations over the contiguous United States (CONUS) from 2020--2023.
We calculate hourly-averaged observations and keep observations with the MADIS quality flag ``Screened'' or ``Verified''. 
We then retain only stations with at least 60\% of data of sufficient quality over the study period, leaving 11,849 stations (Fig. \ref{fig:Methods:stations}), each with a maximum of 35,064 hourly observations.

During both training and evaluation, missing values are masked out. For test stations, we applied additional variable-specific filters intended to remove records with conspicuous indications of measurement artifacts while retaining unusual but potentially valid observations. Temperature and dewpoint stations were excluded if they contained at least 24 consecutive hours with an identical value, a threshold intended to identify possible sensor freezing without excluding shorter periods of limited variability or measurement quantization. Wind stations were excluded if both wind components were exactly zero in more than 90\% of quality-controlled hours, provided the station reported at least 100 such hours. Both criteria were evaluated on the 2023 evaluation year rather than over the full 2020--2023 study period. Although zero wind is physically plausible and some anemometers report zero below a minimum operating threshold (e.g., 2 knots), records with an unusually high proportion of zeros may be unreliable for evaluation. These additional filters were not applied to training stations because moderate observation noise can be tolerated during model fitting, whereas such artifacts could bias evaluation.

\paragraph{Kilometer-scale Weather Products}

The ECMWF Reanalysis v5 (ERA5) dataset~\cite{hersbach2020} from the European Centre for Medium-Range Weather Forecasts (ECMWF) is a global gridded reanalysis product that integrates physical model outputs with observations.
ERA5 provides hourly data on a $0.25^\circ \times 0.25^\circ$ grid (about 28 km at the equator) covering the period from 1940--present.
It includes weather data both at the surface and at different pressure levels.
For this work, we curated 4 years (2020--2023) of surface variables over CONUS: 10 m wind $u$, 10 m wind $v$, 2 m temperature, and 2 m dewpoint temperature.
ERA5 serves as both an input to the Marmot model and as a baseline; it is interpolated to each backbone station and target location using bicubic interpolation.

The High-Resolution Rapid Refresh (HRRR) system~\cite{dowell2022, benjamin2016north} is an hourly updated, convection-allowing NWP system developed by the NOAA for the contiguous U.S.
HRRR assimilates observations into a regional atmospheric model and provides weather fields on an approximately 3 km grid.
We extracted the corresponding 10 m wind $u$, 10 m wind $v$, 2 m temperature, and 2 m dewpoint temperature fields for the 2023 evaluation period.
HRRR is used only as an evaluation baseline, providing a comparison with a higher-resolution regional dynamical model. It is not an input to Marmot.

\begin{figure}
\centering
\includegraphics[width=1\textwidth]{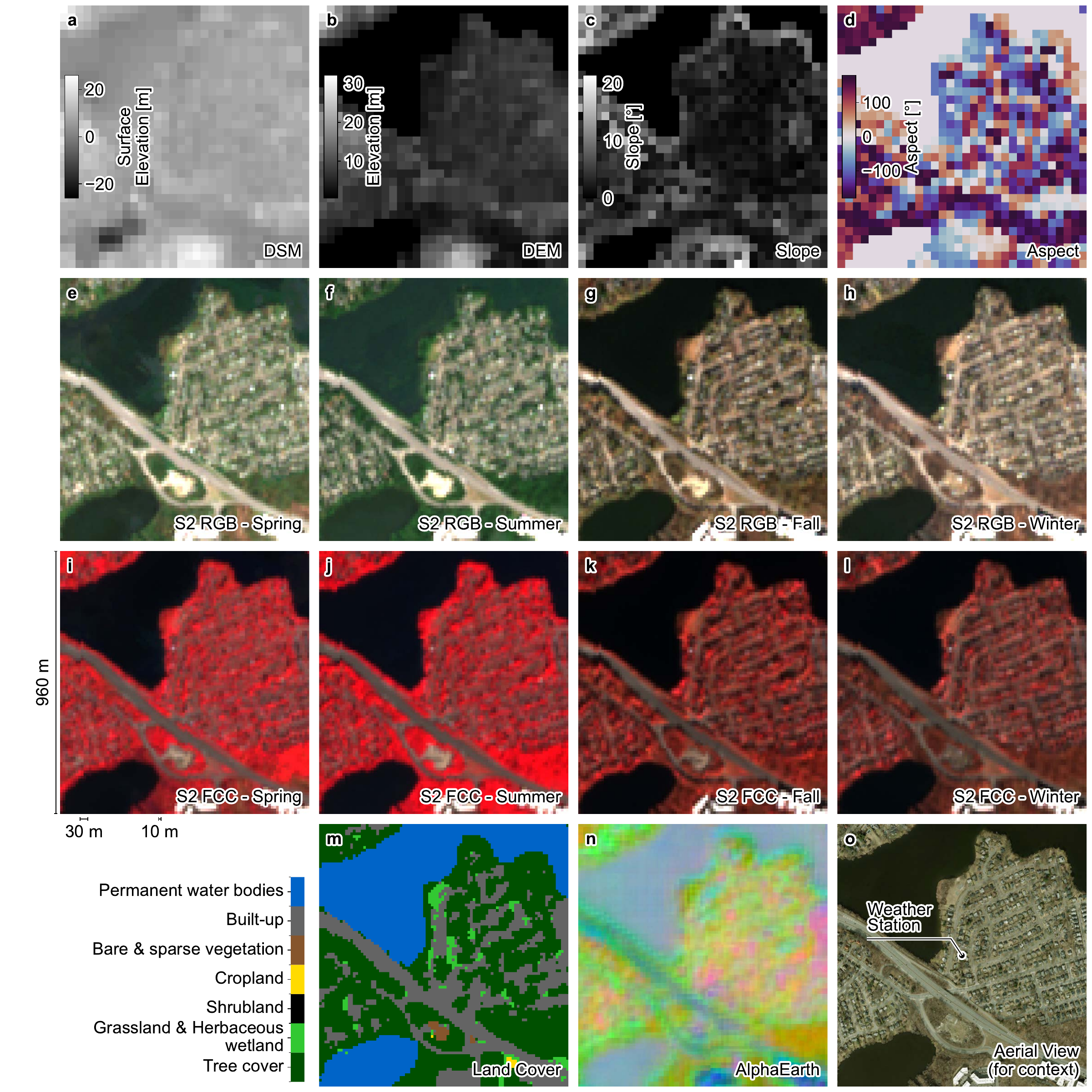}
\caption{
    \textbf{Example of Earth observation inputs for a given location.} 
    \textbf{(a--d)} DSM, DEM, and derived indices.
    \textbf{(e--h)} Sentinel-2 for spring, summer, fall, and winter seasons visualized in RGB. 
    \textbf{(i--l)} Sentinel-2 for the four seasons visualized as a False Color Composite (FCC; NIR, R, G).
    \textbf{(m)} Land cover map with seven classes selected (of which only six are present at this location).
    \textbf{(n)} Geospatial foundation model embeddings produced by AlphaEarth~\cite{brown2025}, visualized via 3 embedding values.
    \textbf{(o)} Aerial image of the same location, not used in the model, for context only.
}\label{fig:Methods:Data:SatelliteData}
\end{figure}

\paragraph{Meter-scale Satellite Imagery}

In our main model specification, we use embeddings from the AlphaEarth geospatial foundation model \cite{brown2025} to provide information about land cover, terrain, and other surface characteristics to Marmot. We also construct conventional Earth observation inputs from raw Sentinel-2 imagery, digital elevation model (DEM), digital surface model (DSM), and a land cover map, to test whether the foundation model embedding adds information beyond them. Below we describe the data used in both approaches.

The AlphaEarth foundation model ingests optical, radar, lidar, DEM, land cover, climatic, and other geospatial information into a 64-dimensional annual embedding at 10 m resolution \cite{brown2025}.
We access AlphaEarth embeddings through Google Earth Engine and downsample the embeddings to 30 m for our main model specification. We use AlphaEarth embeddings from 2019, the year preceding our study period, to ensure no temporal leakage.

Separately, we also prepare raw and derived satellite image chips for each station location, from which we construct conventional Earth observation inputs as an alternative to the AlphaEarth embedding (Fig. \ref{fig:Methods:Data:SatelliteData}).
These image chips are static in time and consist of 30 m-resolution DSM and DEM and 10 m-resolution land cover and Sentinel-2 imagery.
We construct 32$\times$32 pixel 30 m-resolution chips and 96$\times$96 pixel 10 m-resolution chips centered on each weather station.
Each chip thus covers 960$\times$960 meters surrounding each station.
The DSM data are from the JAXA ALOS DSM: Global 30m v3.2~\cite{takaku2014}, and the DEM data are from Copernicus DEM GLO-30: Global 30m Digital Elevation Model~\cite{esa2024}.
We use ESA WorldCover 10 m v200 for land cover data~\cite{zanaga2022esa} and retain 7 categories: Tree Cover, Grassland/Herbaceous wetland, Shrubland, Cropland, Bare/Sparse Vegetation, Built-up and Permanent Water Bodies.
For every location, we also create four seasonal Sentinel-2 composites, each comprised of the four 10 m resolution bands: Blue, Green, Red, and NIR. 
For each season, we select all the images that fall within the given time period, mask out clouds~\cite{pasquarella2023}, and compute the median value for each pixel. For each band and season we retain the station's own pixel together with the mean and standard deviation over the chip, giving the 48 Sentinel-2 columns of Table \ref{tab:Appendix:Results:SurfaceInfo}.

\paragraph{ECOSTRESS Land Surface Temperature (LST)}

ECOSTRESS is an instrument aboard the International Space Station that senses land surface temperature and emissivity (LST\&E) at 70 m resolution with 4 day revisit~\cite{hook2022}.
As LST and air temperature are correlated \cite{good2016lst}, we use the LST data product to qualitatively assess the micro-weather model's ability to infer sub-grid temperature variations.

However, the 2 m air temperature inferred in this work differs from ECOSTRESS LST in several important ways, and their comparison requires careful interpretation.
First, our air temperature is averaged hourly, while ECOSTRESS-sensed LST is instantaneous.
Second, the relationship between ground temperature and 2 m air temperature is highly non-linear~\cite{jin2010, good2017,mo2025,zhang2025}.
A diurnal cycle exists: at night, LST and air temperature are more closely correlated, whereas during the day, the differences can become substantial~\cite{good2017}.
Over vegetated areas, LST is generally lower than air temperature during the day due to evapotranspiration~\cite{good2017}. Bare and sparsely vegetated areas, such as urban regions, exhibit the largest differences between LST and air temperature. Third, satellite-based LST retrievals have higher errors over complex terrain than flat terrain~\cite{dong2025}. Beyond measurement errors, the coupling between LST and air temperature weakens with elevation.

For these reasons, we limit comparison of our air temperature fields with ECOSTRESS LST to a qualitative assessment within case studies to establish general spatial correlations and physical plausibility.

\subsection{Marmot Model Architecture}
\label{sec:methods:model}

Marmot uses a transformer architecture to represent spatial relationships among weather stations and transfer information to arbitrary target locations~\cite{vaswani2017}. Let
\begin{equation}
\mathbf{y} = (T, D, u, v)^\top \in \mathbf{R}^4
\end{equation}
denote the four predicted weather variables: 2 m air temperature, 2 m dewpoint, and $u,v$ components of 10 m wind. Each prediction is conditioned on same-timestamp observations of $\mathbf{y}$ from the 64 nearest backbone stations, spatial relationships between those stations and the target, high-resolution surface information at the station and target locations, the spatial pattern in ERA5, and the time of prediction. Marmot predicts the four components of $\mathbf{y}$ jointly.

The architecture consists of three stages. First, Marmot produces a representation of the neighboring weather-station snapshot. Second, it uses AlphaEarth surface embeddings and station-to-target geometry to determine how informative each station is for the target and combines the stations into a target-specific summary. Third, it combines this summary with ERA5 and temporal information to predict $\mathbf{y}$. An auxiliary connection also allows an attention-weighted average of the station-level differences between observations and ERA5 to contribute directly to the corresponding outputs. In total, the Marmot model contains 2.33 million trainable parameters.

\paragraph{Representing Neighboring Stations}
For each target, Marmot selects the 64 nearest backbone stations by Haversine distance. Each station $i \in \{1, \dots, 64\}$ is represented by a 77-dimensional token containing the four normalized weather observations $\mathbf{y}_i \in \mathbf{R}^4$; five spatial features; a 64-dimensional AlphaEarth embedding at the station location; and four ERA5 innovation terms. The spatial features are latitude and longitude offsets, Haversine distance, logarithm of Haversine distance, and elevation difference. Defining $\mathbf{e}_i \in \mathbb{R}^4$ as the corresponding ERA5 values at station $i$, its innovation vector is
\begin{equation}
\boldsymbol{\delta}_i = \mathbf{y}_i - \mathbf{e}_i.
\end{equation}
Observations that are unavailable or fail quality control are masked out. When a backbone station is itself used as a training target, that station is excluded from its own neighbor set.

Marmot first processes the stations jointly using self-attention. The 77 station features are projected to a 128-dimensional representation and passed through four transformer blocks, each with four attention heads and a feed-forward layer. This stage allows the representation of each station to incorporate information from the surrounding station network. For example, it can represent whether an observation is part of a spatially coherent weather pattern or differs from the other nearby observations. A residual projection adds the resulting contextual information to the station representations.

\paragraph{Querying at the Target Location}
Marmot next assigns each neighboring station $i$ a weight $a_i$ according to how informative it appears for the target using cross-attention~\cite{lin2022,chen2021,vaswani2017}. The target query is generated from the AlphaEarth embedding at the target location. Each neighboring station supplies a key constructed from its AlphaEarth embedding and its location relative to the target, and a value $\mathbf{v}_i \in \mathbf{R}^{256}$ constructed from its updated station representation. The scaled inner product of the target query and each station key determines its attention score. The resulting weights are normalized across the valid neighboring stations so that
\(\sum_{i=1}^{64} a_i = 1\).
Marmot uses these weights to construct a neighboring-station summary
\( \mathbf{h} = \sum_{i=1}^{64} a_i \mathbf{v}_i\),
with more relevant stations contributing more strongly. One set of station weights is used to construct this shared summary for all four components of $\mathbf{y}$. Subsequent prediction layers can nevertheless use the information in $\mathbf{h}$ differently for each output variable.

\paragraph{Adding ERA5 Spatial Information}
Marmot next combines $\mathbf{h}$ with information about how ERA5 varies around the target. For each weather variable, the model receives a $5 \times 5$ ERA5 neighborhood centered on the target. Let $\mathbf{E}_0 \in \mathbf{R}^4$ denote the ERA5 values at the center grid node and $\mathbf{E}_j \in \mathbb{R}^4$ the values at each of the other 24 grid cells. Marmot forms the differences
\begin{equation}
\mathbf{r} = \left[ \mathbf{E}_1 - \mathbf{E}_0, \dots, \mathbf{E}_{24} - \mathbf{E}_0 \right] \in \mathbf{R}^{96}.
\end{equation}
Thus, $\mathbf{r}$ describes the spatial pattern of the coarse ERA5 weather field around the target rather than its absolute level.

The 96 ERA5 features in $\mathbf{r}$ are concatenated with the 256-dimensional station summary $\mathbf{h}$ and projected back to a 256-dimensional representation $\mathbf{z}$.
Marmot passes $\mathbf{z}$ through three residual gated feed-forward blocks and then maps the result to the four weather outputs. These prediction layers learn flexible relationships among the contemporaneous station observations, station and target environments, and the coarse spatial pattern in ERA5.

\paragraph{Representing Time Dependence}
The relationship between a target and its neighboring stations can vary by time of day and season. Marmot represents local solar hour and day of year using sine and cosine features, denoted collectively by $\boldsymbol{\tau}$. It combines $\boldsymbol{\tau}$ with the AlphaEarth embedding at the target, $\mathbf{s}_0$, to calculate a separate scale and offset for each component of $\mathbf{z}$:
\begin{equation}
\widetilde{\mathbf{z}} = \boldsymbol{\gamma}(\boldsymbol{\tau}, \mathbf{s}_0) \odot \mathbf{z} + \boldsymbol{\beta}(\boldsymbol{\tau}, \mathbf{s}_0),
\end{equation}
where $\odot$ denotes elementwise multiplication. This adjustment is applied once, before the three feed-forward blocks. The same station and ERA5 information can therefore be interpreted differently depending on the target environment, local time, and season. For example, relationships between forested and bare locations may differ between day and night or across seasons.

\paragraph{Producing the Weather Predictions}
Let $F(\widetilde{\mathbf{z}}) \in \mathbf{R}^4$ denote the output of the three feed-forward blocks and final prediction layer. An auxiliary connection uses the same station weights to calculate an average innovation,
\begin{equation}
\overline{\boldsymbol{\delta}} = \sum_{i=1}^{64} a_i \boldsymbol{\delta}_i,
\end{equation}
and applies four learned, variable-specific scaling factors $\mathbf{g}(\mathbf{h}) \in \mathbf{R}^4$. The final prediction in the normalized output space is
\begin{equation}
\widehat{\mathbf{y}}_{\mathrm{norm}} = F(\widetilde{\mathbf{z}}) + \mathbf{g}(\mathbf{h}) \odot \overline{\boldsymbol{\delta}}.
\end{equation}
The four components are then transformed back to physical units to obtain $\widehat{\mathbf{y}} = (\widehat{T}, \widehat{D}, \widehat{u}, \widehat{v})^\top$.

\subsection{Marmot Training and Inference}
\label{sec:methods:training}

\begin{figure}
\centering
\includegraphics[width=1\textwidth]{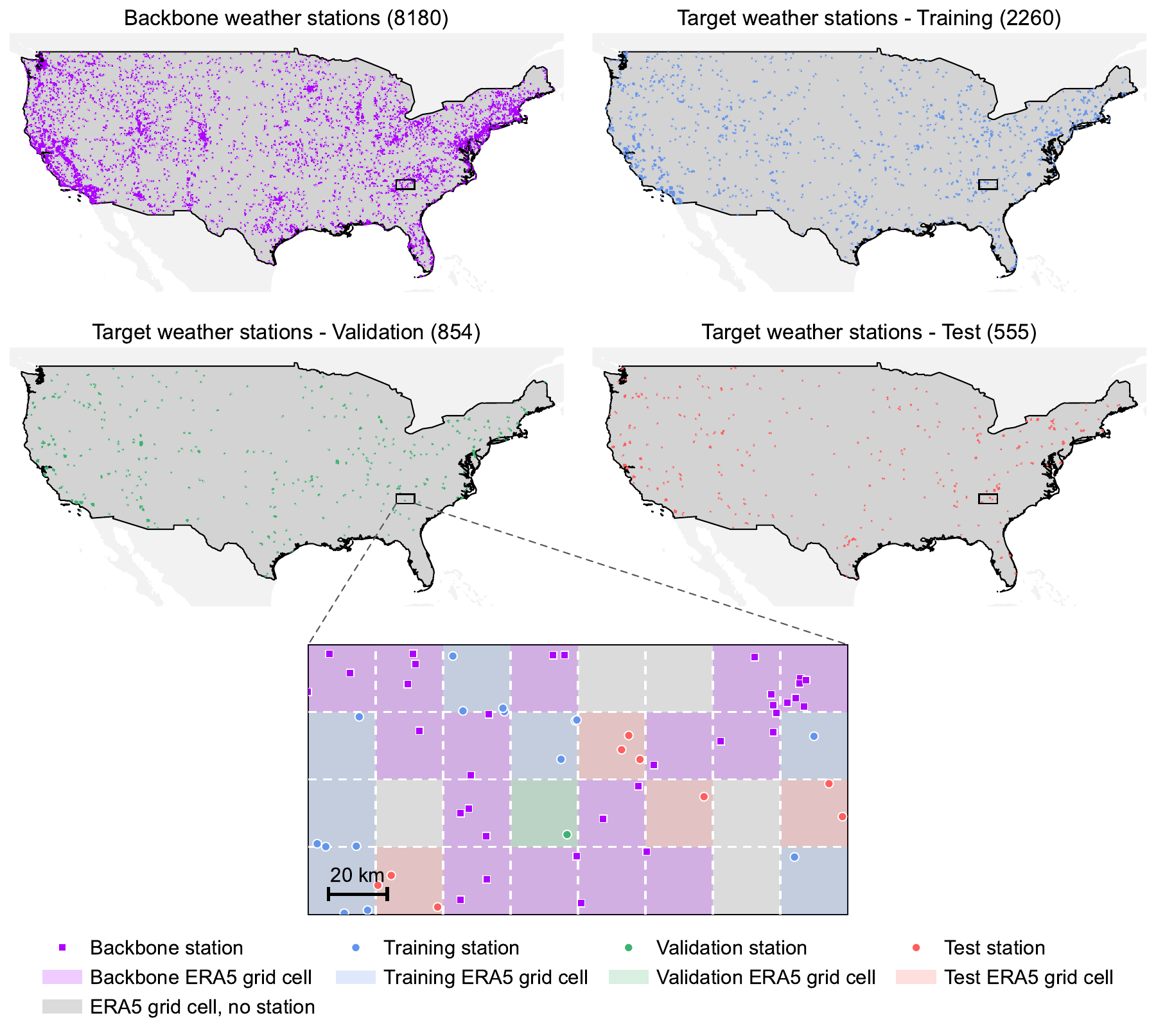}
\caption{
    \textbf{Spatial partitioning of contiguous U.S. ERA5 cells and weather stations for training and evaluation.}
    ERA5 grid cells across the contiguous U.S. are split into four disjoint sets: backbone, training, validation, and test. Surface weather stations from NOAA MADIS inherit the assignment of the ERA5 grid cell in which they are located. Only test set stations are used for evaluation.
}\label{fig:Methods:stations}
\end{figure}

We divided the 11,849 weather stations into 4 groups (Fig. \ref{fig:Methods:stations}): backbone, training, validation, and test sets.
Stations were partitioned by whole $0.25^\circ$ ERA5 grid cell, so that every station in a cell belongs to the same group and no cell is shared between groups. The backbone set (8,180 stations, corresponding to 245,562,977 valid station-time step observations) supplies information on local weather to nearby target locations for as many time steps as possible. The training set contains 2,260 stations (67,915,653 valid observations), and the validation and test sets contain 854 stations (25,651,801 valid observations) and 555 stations (16,632,860 valid observations), respectively.

The model weights were trained by tasking the model with predicting temperature, dewpoint, and the $u,v$ components of wind at training-set and backbone stations (10,440 stations in total) for hourly time steps between 2020--2022. Models were trained for 120,000 steps with a batch size of 128 target station-hours (64 per GPU across two GPUs), using the AdamW optimizer~\cite{loshchilov2019decoupled} with a learning rate of $5\times10^{\text{-4}}$ modified with cosine annealing and a weight decay of 10$^{\text{-2}}$. The model checkpoint used for test set evaluation and inference was selected based on the best validation set performance.

At inference time, Marmot can be queried at any location and timestamp for which an AlphaEarth embedding is available. Each output pixel is treated as an independent target and receives the same inputs as a target station during training. Spatially continuous fields at 30 m resolution are assembled from these per-pixel predictions, with the maximum region size constrained by available memory.

\subsection{Baseline Models}
\label{sec:methods:baselines}

In meteorology, the term ``downscaling'' is broad and encompasses methods with different prediction targets, spatial and temporal scales, and evaluation objectives. Because our task is to predict weather at point-level locations and hourly timesteps, we selected baselines that provide temporally aligned estimates that can be evaluated directly against held-out station observations. These baselines span four sources of local weather information: nearby station observations, the global reanalysis used as our model input, a higher-resolution regional NWP product, and statistical postprocessing learned from station observations. Specifically, we compared against backbone station interpolation, ERA5, HRRR, and ERA5-Model Output Statistics (ERA5-MOS).

\paragraph{Station Interpolation Baseline} This baseline measures how well local conditions can be estimated from nearby station observations using standard spatial interpolation. We interpolated each weather variable from the backbone stations to every test station location by inverse-distance weighting ($1/d^2$ where $d$ is Haversine distance) using the $k=12$ nearest backbone stations, with $k$ selected on the validation set and weights renormalized over whichever of those stations report at that hour. Where none of the twelve report, the estimate escalates to the nearest reporting stations among the 64 nearest backbone stations, and then to the whole backbone network, so that the baseline is defined on the same station-hours as the other methods; this affects fewer than 0.004\% of values that are evaluated. 
We also evaluated radial-basis-function interpolation, but inverse-distance weighting performed better on the validation set and is therefore reported as the station-interpolation baseline.

\paragraph{ERA5 Baseline} ERA5 is a leading global atmospheric reanalysis and provides the coarse weather fields used as inputs to our model; this baseline therefore measures both performance relative to a widely used global weather product and the improvement provided by our model over its own meteorological inputs. We used bilinear interpolation of the ERA5 grid to obtain weather values at test station locations.

\paragraph{HRRR Baseline} Because HRRR is a regional weather model that simulates atmospheric dynamics at substantially higher spatial resolution than ERA5, comparing our model against HRRR places Marmot's improvements in the context of what can be achieved through higher-resolution physical modeling. We assigned each test station the HRRR value from the grid cell containing it at the corresponding hour. Because HRRR stores wind components relative to its Lambert conformal grid, we rotated them to true east and north before evaluation.

\paragraph{ERA5-MOS} This baseline tests whether station-level accuracy can be improved through conventional statistical correction of ERA5 using observations and hand-crafted surface attributes. Model output statistics (MOS) is a statistical postprocessing approach that relates NWP output to observations to correct systematic model errors~\cite{glahn1972mos}. We implemented a MOS baseline that learned the systematic difference between ERA5 and station observations~\cite{benbouallegue2023statistical}. Separate models were trained for 2 m temperature, 2 m dew-point temperature, and the two 10 m wind components using the same training, validation, and test partitions as our model. The target was the hourly residual \(e=x_{\mathrm{ERA5}}-x_{\mathrm{obs}}\), and the MOS prediction was obtained as \(\hat{x}=x_{\mathrm{ERA5}}-\hat{e}\). Training targets were restricted to hours whose MADIS quality flag marks an observed value. Predictors comprised station and ERA5-grid surface attributes --- including location, orographic variability, land-sea mask, vegetation characteristics, station elevation, and the elevation difference between the station and ERA5 grid cell --- together with annual and local-time indicators. We evaluated second-order ridge regression, random forests, and multilayer perceptrons as the regression model. Their performance was similar, consistent with previous findings~\cite{benbouallegue2023statistical}, and we report the specification with the best validation performance, a random forest.

\subsection{Evaluation Protocol}
\label{sec:methods:evaluation}

We evaluate model and baseline performance at held-out test stations using metrics designed to measure overall predictive accuracy, recovery of spatial variability, recovery of local contrasts, and performance under extreme or rapidly changing weather. Unless otherwise stated, metrics are computed using hourly observations from the held-out set. All methods are evaluated on the same available station-time observations.

\paragraph{Aggregate Error Metrics} 
Temperature and dewpoint are evaluated using scalar root mean square error (RMSE) and mean absolute error (MAE). We report RMSE in the main text and MAE in additional tables. Wind is evaluated using the Euclidean distance between observed and predicted horizontal wind vectors, summarized by its root mean square error (vector RMSE) and mean absolute error (vector MAE). These metrics jointly account for errors in wind speed and direction. Scalar errors in wind speed and direction are reported separately.

\paragraph{National Spatial Variance Explained}
Spatial $R^2$ is computed to quantify how well the model captures spatial structure at a particular snapshot in time. At each time step $t$, we remove the across-station mean from both observations and predictions so that
\begin{equation*}
    \tilde{y}_{i,t} = y_{i,t} - \bar{y}_t, \quad \tilde{m}_{i,t} = m_{i,t} - \bar{m}_t,
\end{equation*}
where $\bar{y}_t, \bar{m}_t$ denote spatial means across stations at time $t$. Spatial $R^2$ is then defined as
\begin{equation*}
R^2_{\text{spatial}}(t)
\;=\;
1 - \frac{\sum_{i} (\tilde{y}_{i,t} - \tilde{m}_{i,t})^2}{\sum_{i} \tilde{y}_{i,t}^{\,2}}.
\end{equation*}
For wind, the multivariate $R_{\text{spatial}}^2$ is calculated from the predicted and observed wind vectors using squared Euclidean distances in the numerator and denominator.

Because the observations and predictions are centered separately, this statistic measures agreement in their spatial deviations rather than differences in their spatial means. A value of one indicates perfect recovery of the observed spatial pattern, a value of zero indicates performance equivalent to predicting no spatial variation, and a negative value indicates poorer performance than that reference.
When reporting a single $R^2_{\text{spatial}}$ metric for a model, we average $R^2_{\text{spatial}}(t)$ across all test time steps.

\paragraph{Sub-grid Spatial Variance} 

To evaluate spatial variability below the resolution of Marmot's ERA5 input, we compute $R^2_{\mathrm{spatial}}$ separately within each held-out ERA5 grid cell and hour.

Let $\mathcal{I}_{g,t}$ denote the set of test stations with valid observations in ERA5 grid cell $g$ at time $t$. For a scalar variable, the within-cell observed and predicted anomalies are
\begin{equation*}
\widetilde{y}_{i,g,t}
=
y_{i,t}-\overline{y}_{g,t},
\qquad
\widetilde{m}_{i,g,t}
=
m_{i,t}-\overline{m}_{g,t},
\end{equation*}
where $\overline{y}_{g,t}$ and $\overline{m}_{g,t}$ are the respective means over stations in $\mathcal{I}_{g,t}$. Within-cell spatial variance explained is
\begin{equation*}
R^2_{\mathrm{spatial}}(g,t)
=
1-
\frac{
\sum_{i\in\mathcal{I}_{g,t}}
\left(
\widetilde{y}_{i,g,t}
-
\widetilde{m}_{i,g,t}
\right)^2
}{
\sum_{i\in\mathcal{I}_{g,t}}
\widetilde{y}_{i,g,t}^{2}
}.
\end{equation*}
For wind, the scalar differences are replaced with squared Euclidean distances between the centered horizontal wind vectors.

We compute this statistic for cell--time combinations in which at least five test stations report simultaneously and all five methods yield finite predictions, and we exclude combinations whose observed spatial variance is zero, for which $R^2_{\mathrm{spatial}}$ is undefined. Cells are eligible if they contain at least five test stations. Performance for each grid cell is summarized by the median of $R^2_{\mathrm{spatial}}(g,t)$ over its eligible hours. Figure~\ref{fig:results:subgrid}a reports the distribution of these cell-level medians across the 24 held-out ERA5 cells eligible for each variable, which contain a median of 6 test stations per cell (maximum 38) and a median of 7,991 eligible hours per cell for temperature and dewpoint and 8,148 for wind (minimum 227).

\paragraph{Sub-grid Spatial Contrast}
We also evaluate whether each method recovers differences in weather at the same timestamp between nearby locations. For a pair of test stations $(i,j)$ observed at time $t$, we define the observed and predicted spatial contrasts as
\begin{equation*}
\Delta y_{ij,t}
=
y_{i,t}-y_{j,t},
\qquad
\Delta m_{ij,t}
=
m_{i,t}-m_{j,t}.
\end{equation*}

For a set $\mathcal{P}$ of valid station-pair-time observations, spatial-contrast recovery is quantified using
\begin{equation*}
R^2_{\Delta}
=
1-
\frac{
\sum_{(i,j,t)\in\mathcal{P}}
\left(
\Delta y_{ij,t}
-
\Delta m_{ij,t}
\right)^2
}{
\sum_{(i,j,t)\in\mathcal{P}}
\left(
\Delta y_{ij,t}
\right)^2
},
\end{equation*}
For wind, the contrasts are vector-valued, and the numerator and denominator are computed using squared Euclidean norms. A value of one indicates perfect recovery, zero indicates performance equivalent to predicting identical conditions at both locations, and negative values indicate performance worse than that reference.

For the national analysis, we form pairs of held-out stations within the same ERA5 grid cell, group them according to their geographic separation, and compute $R^2_{\Delta}$ separately within each distance bin. The distance bins are 0--1, 1--2, 2--3, 3--5, 5--7, 7--10, 10--15, 15--20 and 20--30 km. No pairs are excluded on distance grounds: the widest separation between two test stations sharing an ERA5 cell is 28.5 km, so the bins span the full range. Each station-pair-hour observation receives equal weight.

To characterize heterogeneity across station pairs, we additionally calculate \(R_\Delta^2\) separately for each pair using all valid observations for that pair in 2023. For each variable, the observed contrast magnitude of pair \((i,j)\) is summarized by its root-mean-square contrast,
$\operatorname{RMS}(\Delta y_{ij}) = \sqrt{\frac{1}{n_{ij}}\sum_t\left\lVert\Delta y_{ij,t}\right\rVert^2}$.
Figure \ref{fig:results:subgrid_contrast_pairs} shows pair-specific scores as functions of observed RMS contrast and station separation. 

\paragraph{Surface Correlates of Inferred Fields}

To quantify the association between inferred micro-weather structure and surface characteristics, we fit ordinary least-squares regressions separately for each case study in Fig.~\ref{fig:Results:Inference:Phenomena}. The response variable is Marmot's predicted weather value at each 30 m pixel in the inferred field. We fit three specifications using (i) terrain, represented by elevation and slope magnitude derived from the DEM, (ii) land cover as indicator variables for each class covering at least 1\% of the region, and (iii) terrain and land cover together. We report the coefficient of determination, $R^2$, which measures the fraction of spatial variance in the inferred field associated with each set of surface variables (Table~\ref{tab:Appendix:Results:SurfaceCorrelates}).

\paragraph{Extreme Weather Evaluation}

We evaluate performance under extreme weather in three ways: conditional errors across the observed weather distribution, detection at fixed absolute thresholds, and detection of rapid weather changes.

First, to examine performance across the tails of the distribution, we assign each observation its percentile within the empirical distribution of its station. For wind, percentiles are computed using wind speed. Observations are grouped into twelve percentile bins with edges at $0$, $1$, $2$, $5$, $10$, $25$, $50$, $75$, $90$, $95$, $98$, $99$, and $100$. Within each bin we compute RMSE and mean error at each station, then report the median across stations; a station contributes to a bin only if it has at least 20 observations in it, and stations with fewer than 2{,}000 valid hours over the test year are excluded entirely. Mean error indicates systematic over- or underprediction, with positive values denoting overprediction and negative values underprediction.

Second, we evaluate detection of extreme heat, humidity, and wind at fixed absolute thresholds. Temperature thresholds are $32$, $35$, and $40^\circ\mathrm{C}$; dewpoint thresholds are $20$, $24$, and $26^\circ\mathrm{C}$; and wind-speed thresholds are $8$, $10$, and $12\text{m s}^{-1}$. An observed or predicted event occurs when the corresponding value meets or exceeds the threshold. We calculate precision and recall separately at each station and report the median across stations. Precision measures how often predicted extreme events occur, whereas recall measures how often observed extreme events are detected. Recall is only calculated on stations with at least 10 observed events, and precision only at those of these stations where the method predicts at least one event, since precision is otherwise undefined. The number of stations eligible for recall at each threshold is reported in Figure~\ref{fig:results:extrema}b.

Finally, we evaluate rapid increases and decreases over 3-hour intervals. Ramps are defined as the difference between values 3 hours apart. Positive and negative ramp events are identified when the observed or predicted change exceeds the threshold in either direction. We evaluate thresholds of $\pm 5^{\circ}\mathrm{C}$ and $\pm 8^{\circ}\mathrm{C}$ for temperature, $\pm 5^{\circ}\mathrm{C}$ and $\pm 8^{\circ}\mathrm{C}$ for dewpoint, and $\pm 3\text{ m s}^{-1}$ and $\pm 5\text{ m s}^{-1}$for wind speed. Ramp detection is evaluated using station-level precision, recall, and F1 score and summarized across stations in the same manner as detection at absolute thresholds.

\paragraph{Uncertainty Estimation}
We train five models with identical hyperparameters and different random seeds. All aggregate metrics reported for Marmot, and correspondingly for ERA5-MOS, are computed separately for each of the five models and then averaged. The spatially continuous 30 m fields, and the paired-station contrast statistics reported for individual case studies, are generated from a single trained model. Uncertainty in reported performance reflects variation in both the held-out test locations and the training of the learned methods. We use a bootstrap over the held-out ERA5 grid cells and the five independently trained models of Marmot and of ERA5-MOS. Each replicate resamples $0.25$\textdegree{} ERA5 cell blocks rather than individual stations, because stations within a cell share weather; within that same draw it also resamples, with replacement, the five training seeds of ERA5-MOS and Marmot, the two methods that are trained, and averages over them. The other three methods are deterministic and carry no seed term. One cell draw is shared across methods, so differences between models are computed within a replicate. We report error bars for one standard deviation across 2,000 bootstrap replicates.

\subsection{Ablation Studies}
\label{sec:methods:ablation}

\paragraph{Information Source Ablation}
To understand the contribution of different information sources to Marmot's ability to recover micro-weather, we conducted an additive ablation and a leave-one-out ablation. 

In the additive ablation, which is shown in Figure \ref{fig:results:overview}c, we constructed the model progressively from a station-only configuration. The initial learned model received the contemporaneous observations and locations of the 64 nearest stations, along with pre-computed distances between the stations and target. We then sequentially added station-target elevation difference, ERA5 features (comprised of ``innovation'' --- the difference between each neighboring station observation and the corresponding ERA5 value --- and a \(5\times5\) neighborhood of ERA5 cells around the target), AlphaEarth embeddings, and a temporal encoding and neighbor dropout. Each configuration inherited all components in preceding configurations, allowing the incremental contribution of each addition to be evaluated.

In the leave-one-out ablation, we removed components from the full model one at a time. All remaining components were held at their full-model settings.

We note that the additive ablation depends on the order in which components are introduced; for example, because station-target elevation differences and AlphaEarth embeddings contain overlapping information about topography, adding elevation differences first may reduce the subsequent improvement from AlphaEarth. The leave-one-out ablation is instead symmetric across components, but performance losses may be small when other components provide overlapping information.

\paragraph{Surface Information Ablation}
To identify which sources of surface information contribute most to model performance and whether the AlphaEarth embedding provides information beyond conventional surface products, we replaced the AlphaEarth embedding with features derived from terrain (the Copernicus GLO-30 digital elevation model and the ALOS AW3D30 digital surface model, with slope and aspect derived from the DEM), ESA WorldCover land cover, and seasonal Sentinel-2 reflectance composites, individually and in combination.

\subsection{Sensitivity Analysis}
\label{sec:methods:sensitivity}

We sought to understand Marmot's sensitivity to four major hyperparameters: the number of neighboring stations, the spatial resolution of the AlphaEarth embeddings, the spatial extent of the ERA5 input, and the neighbor-dropout rate. All sensitivity analyses were conducted on the validation set. Unless otherwise specified, each analysis varied one setting while holding the remaining architecture, inputs, and training procedure at the reference configuration: 64 neighboring stations, 30-m AlphaEarth embeddings, a \(5\times5\) ERA5 neighborhood, and 25\% neighbor dropout. 

We varied the number of neighboring stations among 16, 32, 64, and 128. We compared AlphaEarth embeddings extracted at 10-, 30-, and 100-m spatial resolution. We varied the ERA5 neighborhood from the target cell alone (\(1\times1\)) to \(3\times3\), \(5\times5\), \(7\times7\), and \(9\times9\) cells. To assess sensitivity to training-time neighbor dropout, we randomly withheld 0, 10, 25, 50, 60, or 75\% of neighboring stations during training; all available neighbors were retained at inference. 

Finally, we evaluated sensitivity to data partitioning by comparing our primary ERA5 grid-cell split against a random station-level split (Table \ref{tab:Appendix:Results:Partitions}). Because our primary evaluation holds out entire ERA5 grid cells, it measures spatial generalization to regions where no backbone stations are available. The random station split, by contrast, allows neighboring stations within the same grid cell to remain in the training set. Comparing the two quantifies the performance impact of spatial autocorrelation between nearby stations during evaluation.

\clearpage 

%
\bibliography{science_template} 

@techreport{ECMWF2025Strategy,
  title        = {ECMWF Strategy 2025--2034},
  author       = {{European Centre for Medium-Range Weather Forecasts}},
  institution  = {European Centre for Medium-Range Weather Forecasts (ECMWF)},
  year         = {2024},
  month        = dec,
  doi          = {10.21957/05dd9b657e},
  url          = {https://www.ecmwf.int/en/about/strategy},
  address      = {Reading, UK},
}

@article{satoh2019GCRMs,
	author = {Satoh, Masaki and Stevens, Bjorn and Judt, Falko and Khairoutdinov, Marat and Lin, Shian-Jiann and Putman, William M. and D{\"u}ben, Peter},
	date = {2019/09/01},
	doi = {10.1007/s40641-019-00131-0},
	id = {Satoh2019},
	isbn = {2198-6061},
	journal = {Current Climate Change Reports},
	number = {3},
	pages = {172--184},
	title = {Global Cloud-Resolving Models},
	url = {https://doi.org/10.1007/s40641-019-00131-0},
	volume = {5},
	year = {2019}}

@article {prein2017challenges,
      author = "Andreas F. Prein and Roy Rasmussen and Graeme Stephens",
      title = "Challenges and Advances in Convection-Permitting Climate Modeling",
      journal = "Bulletin of the American Meteorological Society",
      year = "2017",
      publisher = "American Meteorological Society",
      address = "Boston MA, USA",
      volume = "98",
      number = "5",
      doi = "10.1175/BAMS-D-16-0263.1",
      pages=      "1027 - 1030",
      url = "https://journals.ametsoc.org/view/journals/bams/98/5/bams-d-16-0263.1.xml"
}

@article{li2023divergent,
title = {Divergent urbanization-induced impacts on global surface urban heat island trends since 1980s},
journal = {Remote Sensing of Environment},
volume = {295},
pages = {113650},
year = {2023},
issn = {0034-4257},
doi = {10.1016/j.rse.2023.113650},
url = {https://www.sciencedirect.com/science/article/pii/S0034425723002018},
author = {Long Li and Wenfeng Zhan and Leiqiu Hu and TC Chakraborty and Zhihua Wang and Peng Fu and Dazhong Wang and Weilin Liao and Fan Huang and Huyan Fu and Jiufeng Li and Zihan Liu and Huilin Du and Shasha Wang}
}

@article{wang2025dual,
	author = {Wang, Shasha and Zhan, Wenfeng and Zhou, Bingbing and Tong, Shilu and Chakraborty, TC and Wang, Zhihua and Huang, Kangning and Du, Huilin and Middel, Ariane and Li, Jiufeng and Liu, Zihan and Li, Long and Huang, Fan and Li, Manchun},
	date = {2025/05/01},
	doi = {10.1038/s41558-025-02303-3},
	id = {Wang2025},
	isbn = {1758-6798},
	journal = {Nature Climate Change},
	number = {5},
	pages = {497--504},
	title = {Dual impact of global urban overheating on mortality},
	url = {https://doi.org/10.1038/s41558-025-02303-3},
	volume = {15},
	year = {2025}}

@report{giebel2011wind,
  title       = {The State-of-the-Art in Short-Term Prediction of Wind Power: A Literature Overview},
  author      = {Giebel, Gregor and Brownsword, Richard and Kariniotakis, George and Denhard, Michael and Draxl, Caroline},
  year        = {2011},
  edition     = {2},
  institution = {ANEMOS.plus},
  doi         = {10.11581/DTU:00000017}
}

@Article{eghdami2025sensitivity,
AUTHOR = {Eghdami, Masih and {Jiménez y Muñoz}, Pedro A. and DeCastro, Amy},
TITLE = {Sensitivity to the Representation of Wind for Wildfire Rate of Spread: Case Studies with the Community Fire Behavior Model},
JOURNAL = {Fire},
VOLUME = {8},
YEAR = {2025},
NUMBER = {4},
ARTICLE-NUMBER = {135},
URL = {https://www.mdpi.com/2571-6255/8/4/135},
ISSN = {2571-6255},
DOI = {10.3390/fire8040135}
}

@article{cruz2019fire,
	author = {Cruz, Miguel G. and Alexander, Martin E.},
	journal = {Annals of Forest Science},
	number = {2},
	pages = {44},
	title = {The 10{\%} wind speed rule of thumb for estimating a wildfire's forward rate of spread in forests and shrublands},
	volume = {76},
	year = {2019}}

@book{baklanov2009air,
  title     = {Meteorological and Air Quality Models for Urban Areas},
  editor    = {Baklanov, Alexander and Grimmond, Sue and Mahura, Alexander and Athanassiadou, Maria},
  year      = {2009},
  publisher = {Springer},
  address   = {Berlin, Heidelberg},
  edition   = {1},
  isbn      = {978-3-642-00298-4},
  doi       = {10.1007/978-3-642-00298-4}
}

@article{farr2007srtm,
author = {Farr, Tom G. and Rosen, Paul A. and Caro, Edward and Crippen, Robert and Duren, Riley and Hensley, Scott and Kobrick, Michael and Paller, Mimi and Rodriguez, Ernesto and Roth, Ladislav and Seal, David and Shaffer, Scott and Shimada, Joanne and Umland, Jeffrey and Werner, Marian and Oskin, Michael and Burbank, Douglas and Alsdorf, Douglas},
title = {The Shuttle Radar Topography Mission},
journal = {Reviews of Geophysics},
volume = {45},
number = {2},
pages = {},
doi = {10.1029/2005RG000183},
url = {https://agupubs.onlinelibrary.wiley.com/doi/abs/10.1029/2005RG000183},
eprint = {https://agupubs.onlinelibrary.wiley.com/doi/pdf/10.1029/2005RG000183},
year = {2007}
}

@article{wedi2020baseline,
author = {Wedi, Nils P. and Polichtchouk, Inna and Dueben, Peter and Anantharaj, Valentine G. and Bauer, Peter and Boussetta, Souhail and Browne, Philip and Deconinck, Willem and Gaudin, Wayne and Hadade, Ioan and Hatfield, Sam and Iffrig, Olivier and Lopez, Philippe and Maciel, Pedro and Mueller, Andreas and Saarinen, Sami and Sandu, Irina and Quintino, Tiago and Vitart, Frederic},
title = {A Baseline for Global Weather and Climate Simulations at 1 km Resolution},
journal = {Journal of Advances in Modeling Earth Systems},
volume = {12},
number = {11},
pages = {e2020MS002192},
doi = {10.1029/2020MS002192},
url = {https://agupubs.onlinelibrary.wiley.com/doi/abs/10.1029/2020MS002192},
year = {2020}
}

@article{fisher2020ecostress,
author = {Fisher, Joshua B. and Lee, Brian and Purdy, Adam J. and Halverson, Gregory H. and Dohlen, Matthew B. and Cawse-Nicholson, Kerry and Wang, Audrey and Anderson, Ray G. and Aragon, Bruno and Arain, M. Altaf and Baldocchi, Dennis D. and Baker, John M. and Barral, Hélène and Bernacchi, Carl J. and Bernhofer, Christian and Biraud, Sébastien C. and Bohrer, Gil and Brunsell, Nathaniel and Cappelaere, Bernard and Castro-Contreras, Saulo and Chun, Junghwa and Conrad, Bryan J. and Cremonese, Edoardo and Demarty, Jérôme and Desai, Ankur R. and De Ligne, Anne and Foltýnová, Lenka and Goulden, Michael L. and Griffis, Timothy J. and Grünwald, Thomas and Johnson, Mark S. and Kang, Minseok and Kelbe, Dave and Kowalska, Natalia and Lim, Jong-Hwan and Maïnassara, Ibrahim and McCabe, Matthew F. and Missik, Justine E.C. and Mohanty, Binayak P. and Moore, Caitlin E. and Morillas, Laura and Morrison, Ross and Munger, J. William and Posse, Gabriela and Richardson, Andrew D. and Russell, Eric S. and Ryu, Youngryel and Sanchez-Azofeifa, Arturo and Schmidt, Marius and Schwartz, Efrat and Sharp, Iain and Šigut, Ladislav and Tang, Yao and Hulley, Glynn and Anderson, Martha and Hain, Christopher and French, Andrew and Wood, Eric and Hook, Simon},
title = {ECOSTRESS: NASA's Next Generation Mission to Measure Evapotranspiration From the International Space Station},
journal = {Water Resources Research},
volume = {56},
number = {4},
pages = {e2019WR026058},
doi = {10.1029/2019WR026058},
url = {https://agupubs.onlinelibrary.wiley.com/doi/abs/10.1029/2019WR026058},
year = {2020}
}

@article{drusch2012sentinel2,
title = {Sentinel-2: ESA's Optical High-Resolution Mission for GMES Operational Services},
journal = {Remote Sensing of Environment},
volume = {120},
pages = {25-36},
year = {2012},
note = {The Sentinel Missions - New Opportunities for Science},
issn = {0034-4257},
doi = {10.1016/j.rse.2011.11.026},
url = {https://www.sciencedirect.com/science/article/pii/S0034425712000636},
author = {M. Drusch and U. {Del Bello} and S. Carlier and O. Colin and V. Fernandez and F. Gascon and B. Hoersch and C. Isola and P. Laberinti and P. Martimort and A. Meygret and F. Spoto and O. Sy and F. Marchese and P. Bargellini}
}

@article{yang2018new,
title = {A new generation of the United States National Land Cover Database: Requirements, research priorities, design, and implementation strategies},
journal = {ISPRS Journal of Photogrammetry and Remote Sensing},
volume = {146},
pages = {108-123},
year = {2018},
issn = {0924-2716},
doi = {10.1016/j.isprsjprs.2018.09.006},
url = {https://www.sciencedirect.com/science/article/pii/S092427161830251X},
author = {Limin Yang and Suming Jin and Patrick Danielson and Collin Homer and Leila Gass and Stacie M. Bender and Adam Case and Catherine Costello and Jon Dewitz and Joyce Fry and Michelle Funk and Brian Granneman and Greg C. Liknes and Matthew Rigge and George Xian},
}

@misc{alet2025,
  title = {Skillful Joint Probabilistic Weather Forecasting from Marginals},
  author = {Alet, Ferran and Price, Ilan and {El-Kadi}, Andrew and Masters, Dominic and Markou, Stratis and Andersson, Tom R. and Stott, Jacklynn and Lam, Remi and Willson, Matthew and {Sanchez-Gonzalez}, Alvaro and Battaglia, Peter},
  year = 2025,
  month = jun,
  number = {arXiv:2506.10772},
  eprint = {2506.10772},
  primaryclass = {cs},
  publisher = {arXiv},
  doi = {10.48550/arXiv.2506.10772},
  archiveprefix = {arXiv}
}

@inproceedings{kurth2023fourcastnet,
author = {Kurth, Thorsten and Subramanian, Shashank and Harrington, Peter and Pathak, Jaideep and Mardani, Morteza and Hall, David and Miele, Andrea and Kashinath, Karthik and Anandkumar, Anima},
title = {FourCastNet: Accelerating Global High-Resolution Weather Forecasting Using Adaptive Fourier Neural Operators},
year = {2023},
isbn = {9798400701900},
publisher = {Association for Computing Machinery},
address = {New York, NY, USA},
url = {https://doi.org/10.1145/3592979.3593412},
doi = {10.1145/3592979.3593412},
booktitle = {Proceedings of the Platform for Advanced Scientific Computing Conference},
articleno = {13},
numpages = {11},
location = {Davos, Switzerland},
series = {PASC '23}
}

@InProceedings{nguyen2023climax,
  title = 	 {{C}lima{X}: A foundation model for weather and climate},
  author =       {Nguyen, Tung and Brandstetter, Johannes and Kapoor, Ashish and Gupta, Jayesh K and Grover, Aditya},
  booktitle = 	 {Proceedings of the 40th International Conference on Machine Learning},
  pages = 	 {25904--25938},
  year = 	 {2023},
  editor = 	 {Krause, Andreas and Brunskill, Emma and Cho, Kyunghyun and Engelhardt, Barbara and Sabato, Sivan and Scarlett, Jonathan},
  volume = 	 {202},
  series = 	 {Proceedings of Machine Learning Research},
  month = 	 {23--29 Jul},
  publisher =    {PMLR},
  url = 	 {https://proceedings.mlr.press/v202/nguyen23a.html}
}

@article{bi2023accurate,
	author = {Bi, Kaifeng and Xie, Lingxi and Zhang, Hengheng and Chen, Xin and Gu, Xiaotao and Tian, Qi},
	journal = {Nature},
	number = {7970},
	pages = {533--538},
	title = {Accurate medium-range global weather forecasting with 3D neural networks},
	volume = {619},
	year = {2023}}

@article{bodnar2024,
  title = {A Foundation Model for the {{Earth}} System},
  author = {Bodnar, Cristian and Bruinsma, Wessel P. and Lucic, Ana and Stanley, Megan and Allen, Anna and Brandstetter, Johannes and Garvan, Patrick and Riechert, Maik and Weyn, Jonathan A. and Dong, Haiyu and Gupta, Jayesh K. and Thambiratnam, Kit and Archibald, Alexander T. and Wu, Chun-Chieh and Heider, Elizabeth and Welling, Max and Turner, Richard E. and Perdikaris, Paris},
  year = 2025,
  month = may,
  journal = {Nature},
  volume = {641},
  number = {8065},
  pages = {1180--1187},
  issn = {0028-0836, 1476-4687},
  doi = {10.1038/s41586-025-09005-y}
}

@misc{brown2025,
  title = {{{AlphaEarth Foundations}}: {{An}} Embedding Field Model for Accurate and Efficient Global Mapping from Sparse Label Data},
  author = {Brown, Christopher F. and Kazmierski, Michal R. and Pasquarella, Valerie J. and Rucklidge, William J. and Samsikova, Masha and Zhang, Chenhui and Shelhamer, Evan and Lahera, Estefania and Wiles, Olivia and Ilyushchenko, Simon and Gorelick, Noel and Zhang, Lihui Lydia and Alj, Sophia and Schechter, Emily and Askay, Sean and Guinan, Oliver and Moore, Rebecca and Boukouvalas, Alexis and Kohli, Pushmeet},
  year = 2025,
  month = sep,
  number = {arXiv:2507.22291},
  eprint = {2507.22291},
  primaryclass = {cs},
  publisher = {arXiv},
  doi = {10.48550/arXiv.2507.22291},
  archiveprefix = {arXiv}
}

@article{chandler2020,
  title = {Multisite, Multivariate Weather Generation Based on Generalised Linear Models},
  author = {Chandler, Richard E.},
  year = 2020,
  month = dec,
  journal = {Environmental Modelling \& Software},
  volume = {134},
  pages = {104867},
  issn = {1364-8152},
  doi = {10.1016/j.envsoft.2020.104867}
}

@inproceedings{chen2021,
  title = {{{CrossViT}}: {{Cross-Attention Multi-Scale Vision Transformer}} for {{Image Classification}}},
  booktitle = {2021 {{IEEE}}/{{CVF International Conference}} on {{Computer Vision}} ({{ICCV}})},
  author = {Chen, Chun-Fu Richard and Fan, Quanfu and Panda, Rameswar},
  year = 2021,
  month = oct,
  pages = {347--356},
  publisher = {IEEE},
  address = {Montreal, QC, Canada},
  doi = {10.1109/ICCV48922.2021.00041},
  copyright = {https://doi.org/10.15223/policy-029},
  isbn = {978-1-6654-2812-5}
}

@misc{dong2025,
  title = {Quantitative {{Evaluation}} of {{Complex Terrain Effects}} on {{Land Surface Temperature}}: {{A Data-driven Coupling Model}} of {{Dynamic}} and {{Static Features}}},
  author = {Dong, Ben and Song, Rongcai and Luo, Lirui and Jiang, Xiushi and Chen, Haiwen and Li, Bo and Wang, Yingchun},
  year = 2025,
  month = aug,
  doi = {10.2139/ssrn.5384696}
}

@article{dowell2022,
  title = {The {{High-Resolution Rapid Refresh}} ({{HRRR}}): {{An Hourly Updating Convection-Allowing Forecast Model}}. {{Part I}}: {{Motivation}} and {{System Description}}},
  author = {Dowell, David C. and Alexander, Curtis R. and James, Eric P. and Weygandt, Stephen S. and Benjamin, Stanley G. and Manikin, Geoffrey S. and Blake, Benjamin T. and Brown, John M. and Olson, Joseph B. and Hu, Ming and Smirnova, Tatiana G. and Ladwig, Terra and Kenyon, Jaymes S. and Ahmadov, Ravan and Turner, David D. and Duda, Jeffrey D. and Alcott, Trevor I.},
  year = 2022,
  month = aug,
  journal = {Weather and Forecasting},
  volume = {37},
  number = {8},
  pages = {1371--1395},
  issn = {0882-8156, 1520-0434},
  doi = {10.1175/WAF-D-21-0151.1},
  copyright = {http://www.ametsoc.org/PUBSReuseLicenses}
}

@misc{esa2024,
  title = {Copernicus {{Global Digital Elevation Model}}},
  author = {{European Space Agency}},
  year = 2024,
  doi = {10.5069/G9028PQB}
}

@article{giorgi1991,
  title = {Approaches to the Simulation of Regional Climate Change: {{A}} Review},
  author = {Giorgi, Filippo and Mearns, Linda O.},
  year = 1991,
  month = may,
  journal = {Reviews of Geophysics},
  volume = {29},
  number = {2},
  pages = {191--216},
  issn = {8755-1209, 1944-9208},
  doi = {10.1029/90RG02636},
  copyright = {http://onlinelibrary.wiley.com/termsAndConditions\#vor}
}

@article{good2017,
  title = {A Spatiotemporal Analysis of the Relationship between Near-Surface Air Temperature and Satellite Land Surface Temperatures Using 17 Years of Data from the {{ATSR}} Series},
  author = {Good, Elizabeth J. and Ghent, Darren J. and Bulgin, Claire E. and Remedios, John J.},
  year = 2017,
  journal = {Journal of Geophysical Research: Atmospheres},
  volume = {122},
  number = {17},
  pages = {9185--9210},
  issn = {2169-8996},
  doi = {10.1002/2017JD026880}
}

@article {oyler2016remotely,
      author = "Jared W. Oyler and Solomon Z. Dobrowski and Zachary A. Holden and Steven W. Running",
      title = "Remotely Sensed Land Skin Temperature as a Spatial Predictor of Air Temperature across the Conterminous United States",
      journal = "Journal of Applied Meteorology and Climatology",
      year = "2016",
      publisher = "American Meteorological Society",
      address = "Boston MA, USA",
      volume = "55",
      number = "7",
      doi = "10.1175/JAMC-D-15-0276.1",
      pages=      "1441 - 1457",
      url = "https://journals.ametsoc.org/view/journals/apme/55/7/jamc-d-15-0276.1.xml"
}

@article{hersbach2020,
  title = {The {{ERA5}} Global Reanalysis},
  author = {Hersbach, Hans and Bell, Bill and Berrisford, Paul and Hirahara, Shoji and Hor{\'a}nyi, Andr{\'a}s and {Mu{\~n}oz-Sabater}, Joaqu{\'i}n and Nicolas, Julien and Peubey, Carole and Radu, Raluca and Schepers, Dinand and Simmons, Adrian and Soci, Cornel and Abdalla, Saleh and Abellan, Xavier and Balsamo, Gianpaolo and Bechtold, Peter and Biavati, Gionata and Bidlot, Jean and Bonavita, Massimo and De Chiara, Giovanna and Dahlgren, Per and Dee, Dick and Diamantakis, Michail and Dragani, Rossana and Flemming, Johannes and Forbes, Richard and Fuentes, Manuel and Geer, Alan and Haimberger, Leo and Healy, Sean and Hogan, Robin J. and H{\'o}lm, El{\'i}as and Janiskov{\'a}, Marta and Keeley, Sarah and Laloyaux, Patrick and Lopez, Philippe and Lupu, Cristina and Radnoti, Gabor and {de Rosnay}, Patricia and Rozum, Iryna and Vamborg, Freja and Villaume, Sebastien and Th{\'e}paut, Jean-No{\"e}l},
  year = 2020,
  journal = {Quarterly Journal of the Royal Meteorological Society},
  volume = {146},
  number = {730},
  pages = {1999--2049},
  issn = {1477-870X},
  doi = {10.1002/qj.3803},
  copyright = {\copyright{} 2020 The Authors. Quarterly Journal of the Royal Meteorological Society published by John Wiley \& Sons Ltd on behalf of the Royal Meteorological Society.}
}

@misc{hook2022,
  title = {{{ECOSTRESS}} Swath Land Surface Temperature and Emissivity Instantaneous {{L2}} Global 70 m V002},
  author = {Hook, Simon and Hulley, Glynn},
  year = 2022,
  publisher = {NASA EOSDIS Land Processes Distributed Active Archive Center (DAAC) data set}
}

@article{jijon2021,
  title = {Augmenting the Spatial Resolution of Climate-Change Temperature Projections for City Planners and Local Decision Makers},
  author = {Jij{\'o}n, Juan Diego and Gaudry, Karl-Heinz and Constante, Jessica and Valencia, C{\'e}sar},
  year = 2021,
  month = apr,
  journal = {Environmental Research Letters},
  volume = {16},
  number = {5},
  pages = {054028},
  publisher = {IOP Publishing},
  issn = {1748-9326},
  doi = {10.1088/1748-9326/abf7f2}
}

@article{jin2010,
  title = {Land Surface Skin Temperature Climatology: Benefitting from the Strengths of Satellite Observations},
  author = {Jin, Menglin and Dickinson, Robert E},
  year = 2010,
  month = oct,
  journal = {Environmental Research Letters},
  volume = {5},
  number = {4},
  pages = {044004},
  issn = {1748-9326},
  doi = {10.1088/1748-9326/5/4/044004}
}

@article{lam2023,
  title = {Learning Skillful Medium-Range Global Weather Forecasting},
  author = {Lam, Remi and {Sanchez-Gonzalez}, Alvaro and Willson, Matthew and Wirnsberger, Peter and Fortunato, Meire and Alet, Ferran and Ravuri, Suman and Ewalds, Timo and {Eaton-Rosen}, Zach and Hu, Weihua and Merose, Alexander and Hoyer, Stephan and Holland, George and Vinyals, Oriol and Stott, Jacklynn and Pritzel, Alexander and Mohamed, Shakir and Battaglia, Peter},
  year = 2023,
  month = dec,
  journal = {Science},
  volume = {382},
  number = {6677},
  pages = {1416--1421},
  issn = {0036-8075, 1095-9203},
  doi = {10.1126/science.adi2336}
}

@inproceedings{lin2022,
  title = {{{CAT}}: {{Cross Attention}} in {{Vision Transformer}}},
  booktitle = {2022 {{IEEE International Conference}} on {{Multimedia}} and {{Expo}} ({{ICME}})},
  author = {Lin, Hezheng and Cheng, Xing and Wu, Xiangyu and Shen, Dong},
  year = 2022,
  month = jul,
  pages = {1--6},
  publisher = {IEEE},
  address = {Taipei, Taiwan},
  doi = {10.1109/ICME52920.2022.9859720},
  copyright = {https://doi.org/10.15223/policy-029},
  isbn = {978-1-6654-8563-0}
}

@article{mo2025,
  title = {Understanding Temperature Variations in Mountainous Regions: {{The}} Relationship between Satellite-Derived Land Surface Temperature and in Situ near-Surface Air Temperature},
  author = {Mo, Yaping and Pepin, Nick and Lovell, Harold},
  year = 2025,
  month = mar,
  journal = {Remote Sensing of Environment},
  volume = {318},
  pages = {114574},
  issn = {00344257},
  doi = {10.1016/j.rse.2024.114574}
}

@inproceedings{pasquarella2023,
  title = {Comprehensive Quality Assessment of Optical Satellite Imagery Using Weakly Supervised Video Learning},
  booktitle = {2023 {{IEEE}}/{{CVF Conference}} on {{Computer Vision}} and {{Pattern Recognition Workshops}} ({{CVPRW}})},
  author = {Pasquarella, Valerie J. and Brown, Christopher F. and Czerwinski, Wanda and Rucklidge, William J.},
  year = 2023,
  month = jun,
  pages = {2125--2135},
  publisher = {IEEE},
  address = {Vancouver, BC, Canada},
  doi = {10.1109/CVPRW59228.2023.00206},
  copyright = {https://doi.org/10.15223/policy-029},
  isbn = {979-8-3503-0249-3}
}

@article{price2025,
  title = {Probabilistic Weather Forecasting with Machine Learning},
  author = {Price, Ilan and {Sanchez-Gonzalez}, Alvaro and Alet, Ferran and Andersson, Tom R. and {El-Kadi}, Andrew and Masters, Dominic and Ewalds, Timo and Stott, Jacklynn and Mohamed, Shakir and Battaglia, Peter and Lam, Remi and Willson, Matthew},
  year = 2025,
  month = jan,
  journal = {Nature},
  volume = {637},
  number = {8044},
  pages = {84--90},
  issn = {0028-0836, 1476-4687},
  doi = {10.1038/s41586-024-08252-9}
}

@article{rummukainen2010,
  title = {State-of-the-art with Regional Climate Models},
  author = {Rummukainen, Markku},
  year = 2010,
  month = jan,
  journal = {WIREs Climate Change},
  volume = {1},
  number = {1},
  pages = {82--96},
  issn = {1757-7780, 1757-7799},
  doi = {10.1002/wcc.8},
  copyright = {http://onlinelibrary.wiley.com/termsAndConditions\#vor}
}

@article{schar2020,
  title = {Kilometer-{{Scale Climate Models}}: {{Prospects}} and {{Challenges}}},
  author = {Sch{\"a}r, Christoph and Fuhrer, Oliver and Arteaga, Andrea and Ban, Nikolina and Charpilloz, Christophe and Di Girolamo, Salvatore and Hentgen, Laureline and Hoefler, Torsten and Lapillonne, Xavier and Leutwyler, David and Osterried, Katherine and Panosetti, Davide and R{\"u}dis{\"u}hli, Stefan and Schlemmer, Linda and Schulthess, Thomas C. and Sprenger, Michael and Ubbiali, Stefano and Wernli, Heini},
  year = 2020,
  month = may,
  journal = {Bulletin of the American Meteorological Society},
  volume = {101},
  number = {5},
  pages = {E567-E587},
  issn = {0003-0007, 1520-0477},
  doi = {10.1175/BAMS-D-18-0167.1}
}

@misc{schmude2024,
  title = {Prithvi {{WxC}}: {{Foundation Model}} for {{Weather}} and {{Climate}}},
  author = {Schmude, Johannes and Roy, Sujit and Trojak, Will and Jakubik, Johannes and Civitarese, Daniel Salles and Singh, Shraddha and Kuehnert, Julian and Ankur, Kumar and Gupta, Aman and Phillips, Christopher E. and Kienzler, Romeo and Szwarcman, Daniela and Gaur, Vishal and Shinde, Rajat and Lal, Rohit and Da Silva, Arlindo and Diaz, Jorge Luis Guevara and Jones, Anne and Pfreundschuh, Simon and Lin, Amy and Sheshadri, Aditi and Nair, Udaysankar and Anantharaj, Valentine and Hamann, Hendrik and Watson, Campbell and Maskey, Manil and Lee, Tsengdar J. and Moreno, Juan Bernabe and Ramachandran, Rahul},
  year = 2024,
  month = sep,
  number = {arXiv:2409.13598},
  eprint = {2409.13598},
  primaryclass = {physics},
  publisher = {arXiv},
  archiveprefix = {arXiv},
  doi = {10.48550/arXiv.2409.13598},
}

@article{takaku2014,
  title = {Generation of {{High Resolution Global DSM}} from {{ALOS PRISM}}},
  author = {Takaku, J. and Tadono, T. and Tsutsui, K.},
  year = 2014,
  month = apr,
  journal = {The International Archives of the Photogrammetry, Remote Sensing and Spatial Information Sciences},
  volume = {XL-4},
  pages = {243--248},
  issn = {2194-9034},
  doi = {10.5194/isprsarchives-XL-4-243-2014},
  copyright = {https://creativecommons.org/licenses/by/3.0/}
}

@inproceedings{vaswani2017,
  title = {Attention Is {{All}} You {{Need}}},
  booktitle = {Advances in Neural Information Processing Systems},
  author = {Vaswani, Ashish and Shazeer, Noam and Parmar, Niki and Uszkoreit, Jakob and Jones, Llion and Gomez, Aidan N and Kaiser, {\L}ukasz and Polosukhin, Illia},
  year = 2017,
  month = jun,
  volume = {30},
  address = {Long Beach, CA, USA}
}

@article{zanaga2022esa,
  title = {{{ESA WorldCover}} 10 m 2021 V200},
  author = {Zanaga, Daniele and Van De Kerchove, Ruben and Daems, Dirk and De Keersmaecker, Wanda and Brockmann, Carsten and Kirches, Grit and Wevers, Jan and Cartus, Oliver and Santoro, Maurizio and Fritz, Steffen and others},
  year = 2022,
  publisher = {Zenodo}
}

@article{zhang2025,
  title = {Global Evaluation of High-Resolution {{ECOSTRESS}} Land Surface Temperature and Emissivity Products: {{Collection}} 1 versus {{Collection}} 2},
  author = {Zhang, Huanyu and Mahmood, Amber N. and Hu, Tian and Mallick, Kanishka and Didry, Yoanne and Hitzelberger, Patrik and Szantoi, Zoltan and {P{\'e}rez-Planells}, Llu{\'i}s and G{\"o}ttsche, Frank M. and Hulley, Glynn C. and Hook, Simon J.},
  year = 2025,
  month = aug,
  journal = {Remote Sensing of Environment},
  volume = {326},
  pages = {114799},
  issn = {00344257},
  doi = {10.1016/j.rse.2025.114799}
}

@article{good2016lst,
author = {Good, Elizabeth Jane},
title = {An in situ-based analysis of the relationship between land surface “skin” and screen-level air temperatures},
journal = {Journal of Geophysical Research: Atmospheres},
volume = {121},
number = {15},
pages = {8801-8819},
doi = {10.1002/2016JD025318},
url = {https://agupubs.onlinelibrary.wiley.com/doi/abs/10.1002/2016JD025318},
year = {2016}
}

@article {benbouallegue2023statistical,
      author = "Zied Ben Bouallègue and Fenwick Cooper and Matthew Chantry and Peter Düben and Peter Bechtold and Irina Sandu",
      title = "Statistical Modeling of 2-m Temperature and 10-m Wind Speed Forecast Errors",
      journal = "Monthly Weather Review",
      year = "2023",
      publisher = "American Meteorological Society",
      address = "Boston MA, USA",
      volume = "151",
      number = "4",
      doi = "10.1175/MWR-D-22-0107.1",
      pages=      "897 - 911",
      url = "https://journals.ametsoc.org/view/journals/mwre/151/4/MWR-D-22-0107.1.xml"
}

@article {glahn1972mos,
      author = "Harry R.  Glahn and Dale A.  Lowry",
      title = "The Use of Model Output Statistics (MOS) in Objective Weather Forecasting",
      journal = "Journal of Applied Meteorology and Climatology",
      year = "1972",
      publisher = "American Meteorological Society",
      address = "Boston MA, USA",
      volume = "11",
      number = "8",
      doi = "10.1175/1520-0450(1972)011<1203:TUOMOS>2.0.CO;2",
      pages=      "1203 - 1211",
      url = "https://journals.ametsoc.org/view/journals/apme/11/8/1520-0450_1972_011_1203_tuomos_2_0_co_2.xml"
}

@article { benjamin2016north,
      author = "Stanley G. Benjamin and Stephen S. Weygandt and John M. Brown and Ming Hu and Curtis R. Alexander and Tatiana G. Smirnova and Joseph B. Olson and Eric P. James and David C. Dowell and Georg A. Grell and Haidao Lin and Steven E. Peckham and Tracy Lorraine Smith and William R. Moninger and Jaymes S. Kenyon and Geoffrey S. Manikin",
      title = "A North American Hourly Assimilation and Model Forecast Cycle: The Rapid Refresh",
      journal = "Monthly Weather Review",
      year = "2016",
      publisher = "American Meteorological Society",
      address = "Boston MA, USA",
      volume = "144",
      number = "4",
      doi = "10.1175/MWR-D-15-0242.1",
      pages=      "1669 - 1694",
      url = "https://journals.ametsoc.org/view/journals/mwre/144/4/mwr-d-15-0242.1.xml"
}

@inproceedings{loshchilov2019decoupled,
  title     = {Decoupled Weight Decay Regularization},
  author    = {Loshchilov, Ilya and Hutter, Frank},
  booktitle = {International Conference on Learning Representations},
  year      = {2019},
  url       = {https://openreview.net/forum?id=Bkg6RiCqY7}
}

@inproceedings{miller2005madis,
  author    = {Miller, Patricia A. and Barth, Michael F. and Benjamin, Leon A.},
  title     = {An Update on {MADIS} Observation Ingest, Integration,
               Quality Control, and Distribution Capabilities},
  booktitle = {Proceedings of the 21st International Conference on
               Interactive Information Processing Systems for Meteorology,
               Oceanography, and Hydrology},
  year      = {2005},
  address   = {San Diego, California},
  publisher = {American Meteorological Society},
  note      = {Paper J7.12},
  url       = {https://ams.confex.com/ams/pdfpapers/86703.pdf}
}
\bibliographystyle{sciencemag}

%
%
%
%
%
%


\newpage

\section*{Acknowledgments}
The authors express their gratitude to Daniel Salles Civitarese, Johannes Jakubik, and Bj\"{o}rn L\"{u}tjens for inspiring discussions and to Chris Hill for his support and guidance.

\paragraph*{Funding:}
J.G. was supported by an MIT-IBM Watson AI Lab award sponsored by Shell Information Technology. Q.Y. and R.H. were partially supported by the National Aeronautics and Space Administration (NASA) under award No. 80NSSC26K0189. 

\paragraph*{Author contributions:}
Conceptualization: J.G., Q.Y., R.H., S.W. Data curation: J.G., Q.Y., R.H. Formal analysis: J.G., Q.Y., R.H., S.W. Funding acquisition: J.V., D.H., C.W., S.W. Investigation: J.G., Q.Y., R.H. Methodology: J.G., Q.Y., R.H., E.S., A.C., S.W. Software: J.G., Q.Y., R.H., Y.Z. Supervision: S.W. Visualization: J.G., R.H., S.W.
Writing -- original draft: J.G., S.W. Writing -- review and editing: J.G., Q.Y., R.H., S.W.

\paragraph*{Competing interests:}
There are no competing interests to declare.

\paragraph*{Data and materials availability:}
Processed data and code will be made publicly available upon publication.
ERA5 data are publicly available from the Copernicus Climate Data Store
(\path{DOI: 10.24381/cds.adbb2d47}), and MADIS observations are available from
NOAA (\url{https://madis.ncep.noaa.gov/}). Surface data were accessed
through the Google Earth Engine Data Catalog using collections
\path{ESA/WorldCover/v200},
\path{COPERNICUS/DEM/GLO30},
\path{JAXA/ALOS/AW3D30/V3_2},
\path{COPERNICUS/S2_SR_HARMONIZED}, and
\path{GOOGLE/SATELLITE_EMBEDDING/V1/ANNUAL}.
ECOSTRESS L2T Land Surface Temperature and Emissivity (LSTE) Version 2 data were obtained through the NASA Earthdata portal. HRRR data were obtained from NOAA's HRRR archive in the Registry of Open Data on AWS. We used the zero-hour analysis fields.

\newpage


\renewcommand{\thefigure}{S\arabic{figure}}
\renewcommand{\thetable}{S\arabic{table}}
\renewcommand{\theequation}{S\arabic{equation}}
\renewcommand{\thepage}{S\arabic{page}}
\setcounter{figure}{0}
\setcounter{table}{0}
\setcounter{equation}{0}
\setcounter{page}{1} 







\newpage
\section{Supplementary Materials}

\subsection{Station Pair-level Variation in Spatial Contrast Recovery}
\label{sec:supp:pairs}

\begin{figure}[bp]
\begin{center}
\includegraphics[width=1\textwidth]{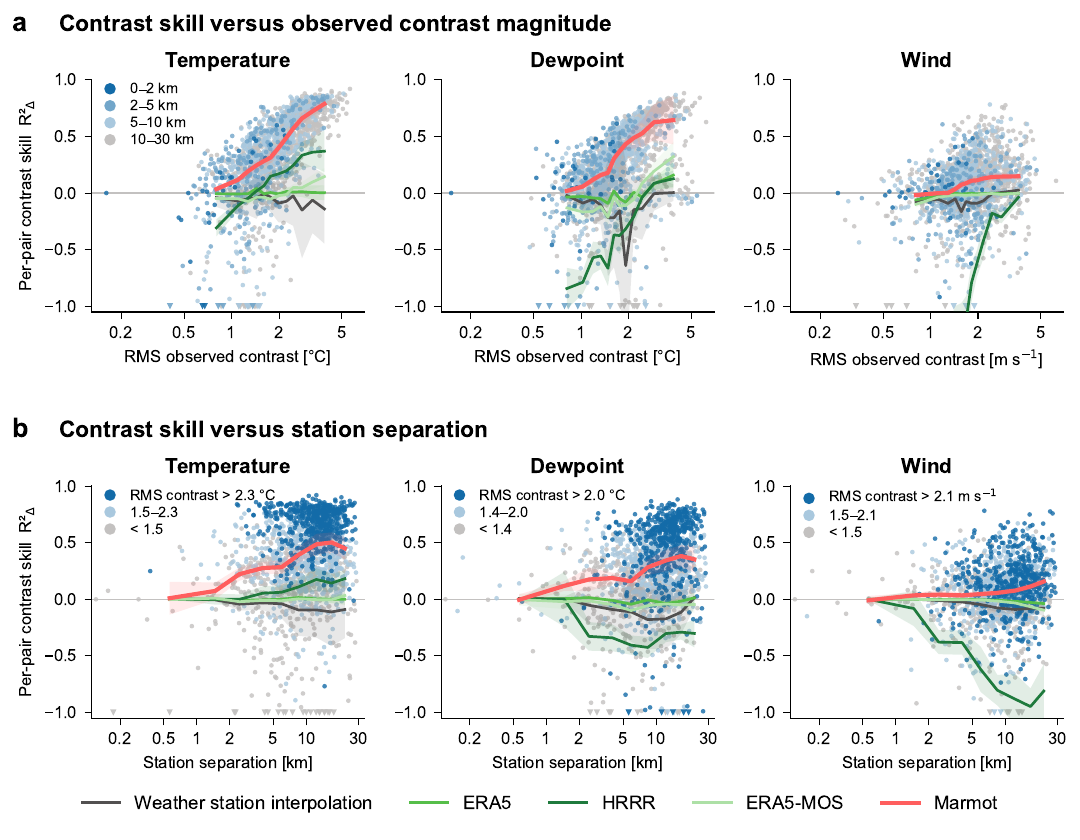}
\end{center}
\caption{
    \textbf{Station pair-level contrast skill as a function of observed contrast magnitude and station separation.}
    Contrast skill is calculated separately for each pair of held-out stations using valid observations for that pair in 2023. Each point therefore represents one station pair. \textbf{(a)} Scatter plot of pair-specific Marmot skill as a function of the RMS observed contrast for temperature, dewpoint, and wind. Point color denotes station separation. Curves show running median pair-level skill for Marmot and the baseline methods. \textbf{(b)} Scatter plot of pair-specific Marmot skill as a function of station separation. Point color denotes the observed RMS contrast. Downward triangles at the lower axis boundary indicate scores below the displayed range.
    }
\label{fig:results:subgrid_contrast_pairs}
\end{figure}

Figure \ref{fig:results:subgrid}b measures aggregate contrast recovery by pooling station-pair-hour observations within each separation bin. Because pairs with larger observed contrasts contribute more to this pooled statistic, we additionally calculate contrast skill separately for each station pair (Fig. \ref{fig:results:subgrid_contrast_pairs}). Pair-specific scores vary substantially at all separations. Median temperature skill generally declines toward shorter separations, while dewpoint and wind approach zero more rapidly. Scores are also generally higher for pairs with larger observed contrasts, consistent in part with the greater relative penalty imposed by \(R_\Delta^2\) when the true contrast is small (Section~\ref{sec:methods:evaluation}). At sub-kilometer separations, the limited station sample includes both substantial temperature recovery and negative scores associated with incorrect inferred contrasts. Thus, the positive aggregate temperature estimate below 1 km reflects recovery of some observed spatial contrasts rather than uniform skill.

\subsection{Temporal Recovery at Held-out Stations}
\label{sec:supp:temporal}

\begin{figure}[tbp]
\begin{center}
\includegraphics[width=1\textwidth]{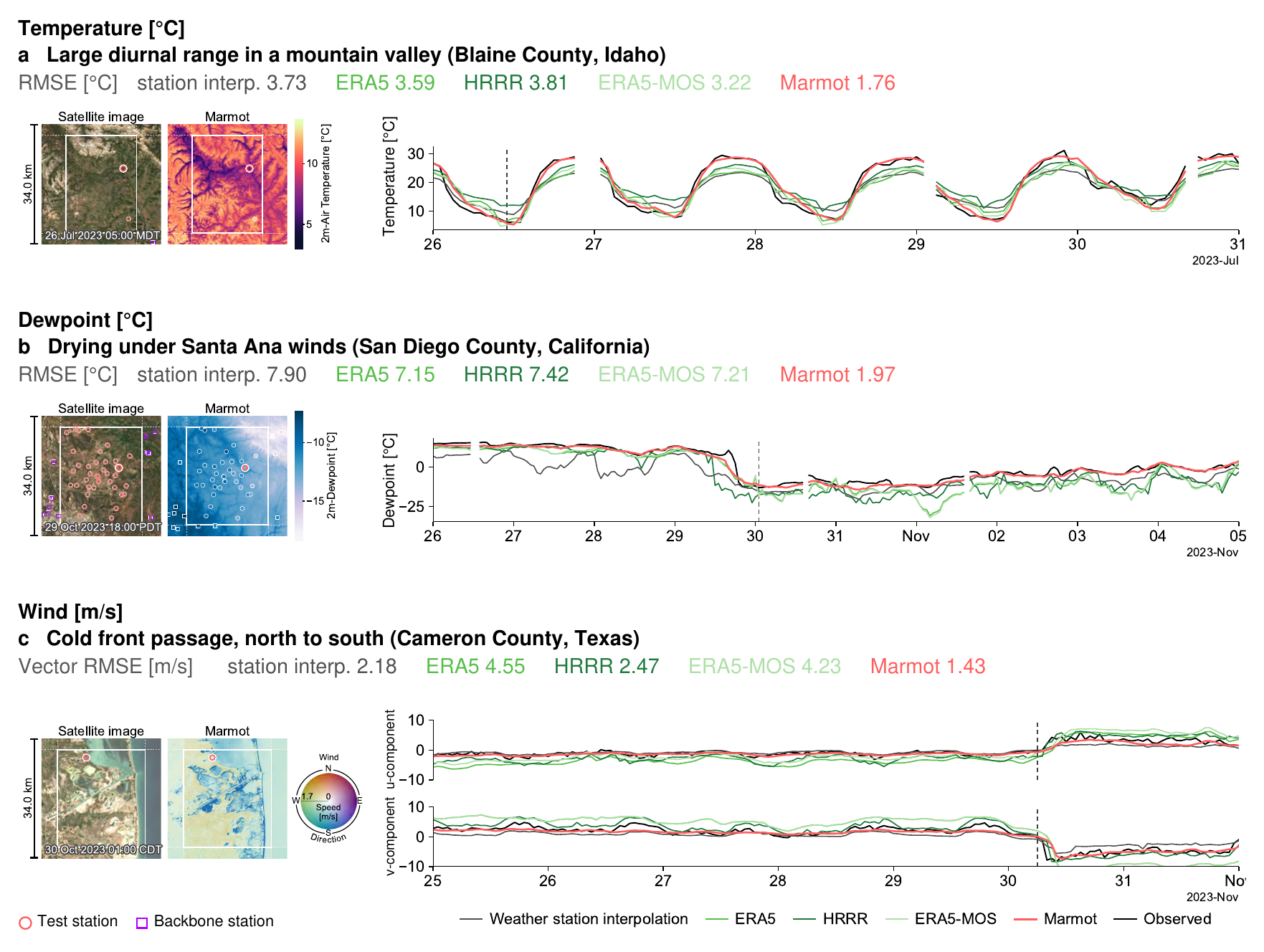}
\end{center}
\caption{
    \textbf{Temporal behavior of inferred micro-weather at example held-out stations.} Hourly time series compare weather station interpolation, ERA5 reanalysis, HRRR analysis, ERA5-MOS and Marmot against station observations during 2023.
     \textbf{(a)} Station located on a valley floor between mountain ranges (Blaine County, Idaho).
     \textbf{(b)} Station located in inland terrain during a Santa Ana wind event (San Diego County, California).
     \textbf{(c)} Station located on the coastal plain during a cold-front passage (Cameron County, Texas).
}
\label{fig:results:timeseries:phenomena}
\end{figure}

To complement the spatial analyses in the main text, we examine the temporal behavior of Marmot's predictions at held-out test stations. We first revisit the station pairs highlighted in the sub-grid analysis and characterize their observed and predicted contrasts over diurnal and seasonal cycles (Fig. \ref{fig:results:subgrid_case_cycles}). We then examine individual weather events at held-out stations where Marmot improves upon the baselines (Fig. \ref{fig:results:timeseries:phenomena}); failure cases are examined separately in Section~\ref{sec:supp:failures}.

The station-pair contrasts in Warren County, San Diego, and Austin (Fig. \ref{fig:results:subgrid_cases}) exhibit systematic temporal variation rather than persistent offsets. In Warren County, the observed contrast reverses sign over the diurnal cycle. Marmot reproduces this reversal and much of its magnitude, whereas baselines predict weak contrasts. Marmot also recovers the seasonal trend in temperature contrast most closely among the methods. In San Diego County, Marmot captures the dewpoint diurnal evolution better than baselines, whereas the seasonal magnitude is overall best reproduced by HRRR. Lastly, the Austin wind-speed contrast is partially recovered by Marmot, but HRRR recovers the seasonal contrast better.

Selected weather events provide additional evidence of temporal recovery (Fig. \ref{fig:results:timeseries:phenomena}). 
In a valley in central Idaho, Marmot captures a large diurnal temperature range that is underestimated by the baselines, reducing RMSE to $1.76^\circ$C compared to $3.22^\circ$C for the next-best method. During a Santa Ana event in San Diego, observed dewpoint falls rapidly by more than $20^\circ$C; Marmot captures both the pre-event dewpoint and the subsequent drying, reducing RMSE to $1.97^\circ$C versus $7.15^\circ$C for the next-best baseline. Finally, during a cold-front passage along the Texas Gulf Coast, Marmot follows the frontal wind shift with the lowest vector error (1.43 m s\(^{-1}\) versus 2.18 m s\(^{-1}\) next-best), although its timing is delayed relative to HRRR.

\subsection{Additional Physical Structure of Recovered Micro-weather}
\label{sec:supp:cases}

The main text presents examples of inferred micro-weather associated with topography and land cover. Here, we examine additional cases in which these relationships vary with atmospheric conditions, including temperature inversions, urban heat, and terrain-channeled wind.

\begin{figure}[tbp]
\begin{center}
\includegraphics[width=1\textwidth]{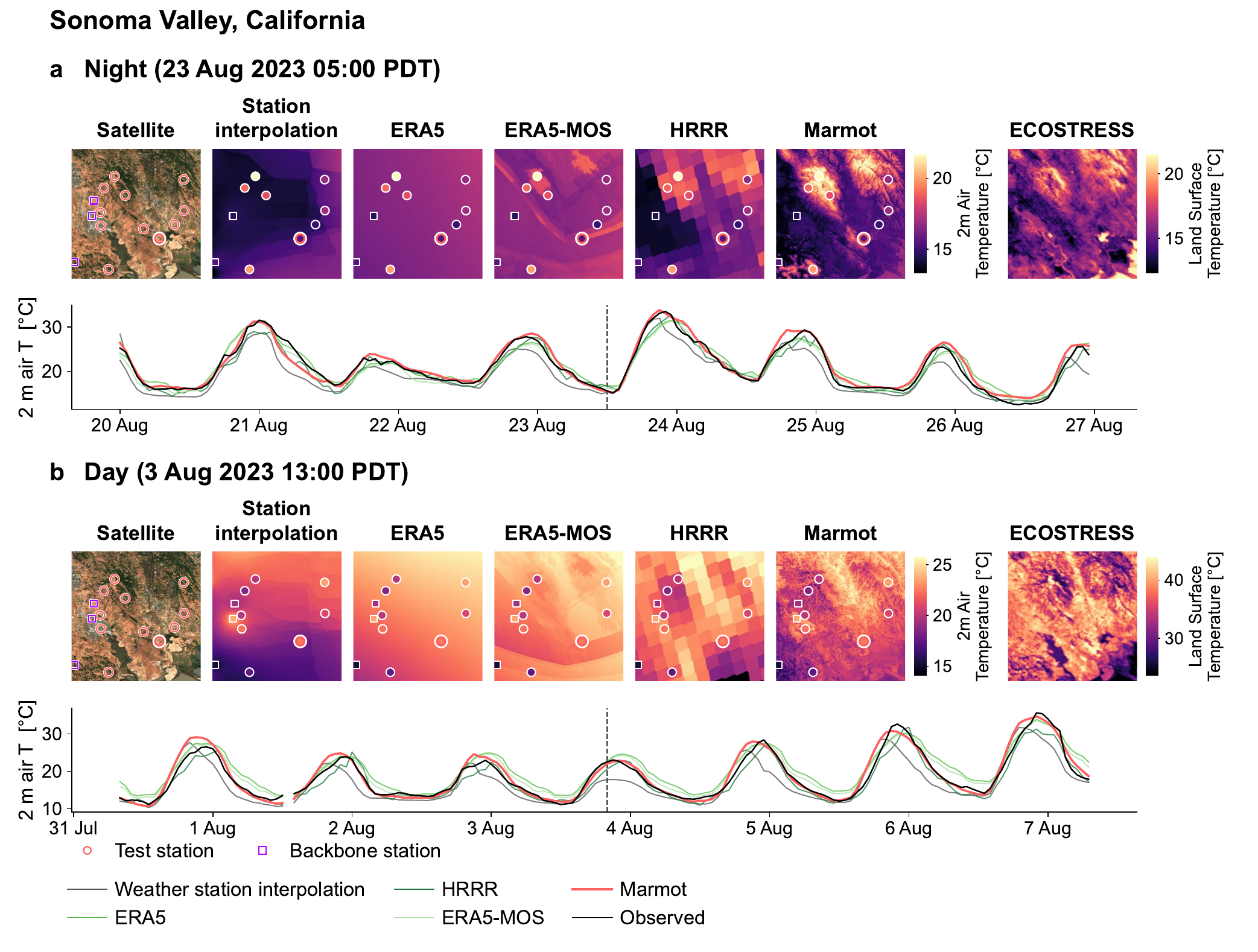}
\end{center}
\caption{
    \textbf{Example of temperature inversion in Sonoma Valley, California.}
    \textbf{(a)} Nighttime temperatures observed at weather stations show a temperature inversion on August 23, 2023, in which higher elevations exhibit higher temperatures than lower elevations. \textbf{(b)} In the daytime on August 3, 2023, no temperature inversion is observed. Temperature fields generated by Marmot recover inversions at night and the lack thereof in the day.  Satellite imagery, temperature fields predicted by baseline methods, and ECOSTRESS land surface temperature are also shown for comparison. Time series of temperature compare Marmot and baseline method predictions against observed values.
}
\label{fig:results:casestudy:sonoma_inversion}
\end{figure}

In Sonoma Valley, California, the relationship between temperature and elevation varies with atmospheric conditions (Fig. \ref{fig:results:casestudy:sonoma_inversion}). Across 4,380 nighttime hours in 2023, 57\% exhibit a temperature inversion --- defined here as a positive relationship between temperature and elevation --- across 17 stations spanning 8--723 m elevation, compared with 24\% during daytime. The occurrence of nighttime inversions is strongly related to wind speed: 79\% of nighttime hours with winds below 1 m s\(^{-1}\) are inverted, compared with only 1\% above 3 m s\(^{-1}\). During a calm night on 23 August, observations show a positive lapse rate of \(+0.54^\circ\)C per 100 m, with warmer temperatures at higher elevations; Marmot reproduces this inversion and its spatial structure. During a daytime example under stronger winds, observations instead show the conventional decrease in temperature with elevation, which Marmot also recovers. Contemporaneous ECOSTRESS observations exhibit similar spatial structure in both examples (Fig. \ref{fig:results:casestudy:sonoma_inversion}). Together, these cases show that Marmot does not impose a fixed relationship between temperature and elevation, but can recover changes in the sign and magnitude of this relationship under different atmospheric conditions.

\begin{figure}[tbp]
\begin{center}
\includegraphics[width=1\textwidth]{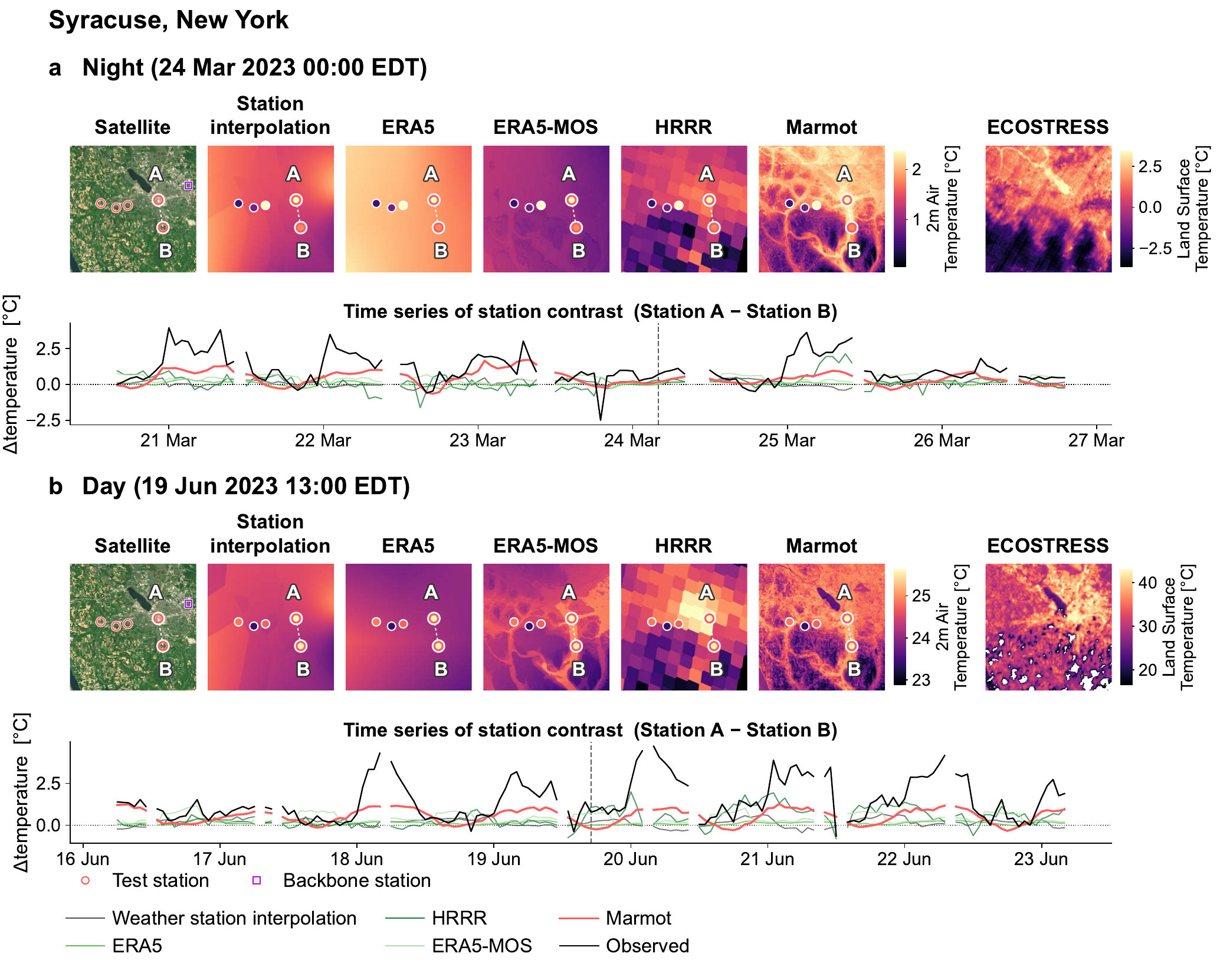}
\end{center}
\caption{
    \textbf{Example of urban heat in Syracuse, New York.}
    \textbf{(a)} Nighttime and \textbf{(b)} daytime temperature contrast between Station A, a held-out station surrounded by built area, and Station B, a held-out station 6.8 km away at the same elevation surrounded by tree cover and grassland. Station A is the warmer of the two in 91\% of the hours drawn in (a) and 95\% in (b). Every method understates the contrast; Marmot recovers the largest share of it, 39\% at night and 36\% by day, against 32\% and 30\% for the best baseline.
}
\label{fig:results:casestudy:syracuse_urban}
\end{figure}

A second example illustrates recovery of urban temperature contrasts (Fig. \ref{fig:results:casestudy:syracuse_urban}). In Syracuse, New York, a held-out station in a 53\%-built environment is on average 1.49 °C warmer at night than a held-out station 6.8 km away in an 11\%-built environment at the same elevation, a contrast that falls to 0.74 °C by day. Marmot partially recovers this spatial contrast and more closely estimates temperature at the urban station than the baseline methods, although all methods underestimate its magnitude. The inferred field delineates fine-scale temperature structure associated with the urban surface that is absent from station interpolation and the coarser atmospheric products, with qualitatively similar structure visible in contemporaneous ECOSTRESS land surface temperature.

\begin{figure}[tbp]
\begin{center}
\includegraphics[width=0.8\textwidth]{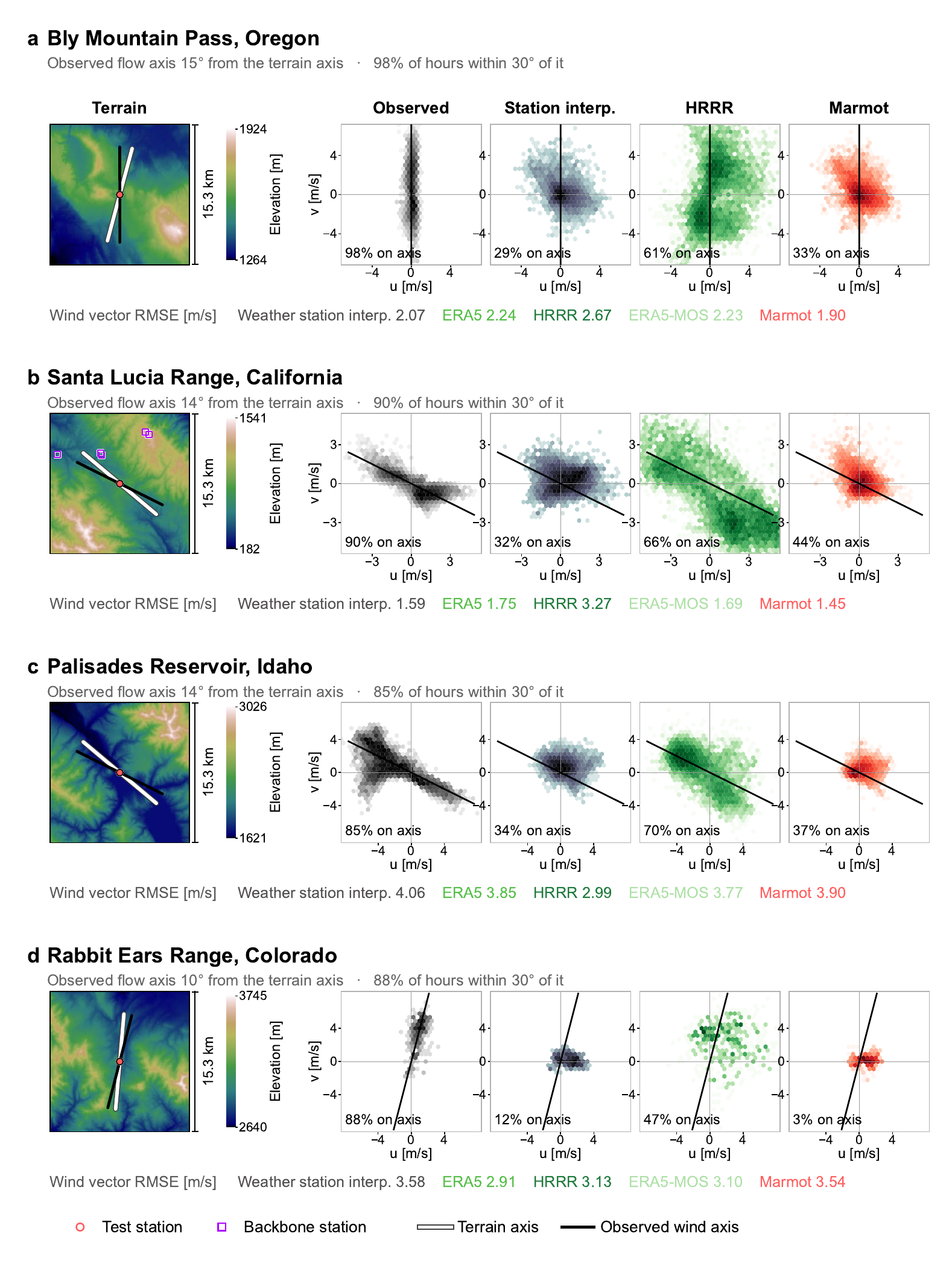}
\end{center}
\caption{
    \textbf{Examples of wind channeling through valleys.} In \textbf{(a)} Bly Mountain Pass, Oregon and \textbf{(b)} Santa Lucia Range, California, Marmot's wind vector RMSE is lower than baseline methods'. In \textbf{(c)} Palisades Reservoir, Idaho and \textbf{(d)} Rabbit Ears Range, Colorado, one or more baseline methods outperform Marmot.
}
\label{fig:results:wind_channeling}
\end{figure}

We additionally test whether Marmot recovers terrain channeling of near-surface wind (Fig. \ref{fig:results:wind_channeling}). We first identify valley stations among the 504 held-out stations with sufficient wind observations, using the DEM to determine whether a station is bounded by higher terrain and, if so, the orientation and degree of topographic confinement of the valley. The observed average wind direction increasingly aligns with the valley orientation as confinement increases, with the median angular difference decreasing from \(38^\circ\) in weakly confined valleys (channel walls are $<50$ m height; $n=228$) to \(16^\circ\) in strongly confined valleys ($>150$ m height; $n=29$). We therefore focus on the most strongly confined stations where observed winds clearly follow the terrain. In 9 out of the 15 most highly confined stations, Marmot achieves the lowest wind-vector RMSE. However, even in those cases, it only partially recovers the strong directional confinement seen in the observations. Wind channeling through narrow valleys therefore exposes a remaining limitation shared by all methods. Figure \ref{fig:results:wind_channeling} shows two of Marmot's relative successes and two cases in which another method performs better.

\subsection{Failure Modes}
\label{sec:supp:failures}

Marmot does not recover fine-scale weather equally well across all locations and atmospheric conditions. We examine cases in which Marmot is outperformed by baseline methods to better understand its failure modes.

\begin{figure}[tbp]
\begin{center}
\includegraphics[width=1\textwidth]{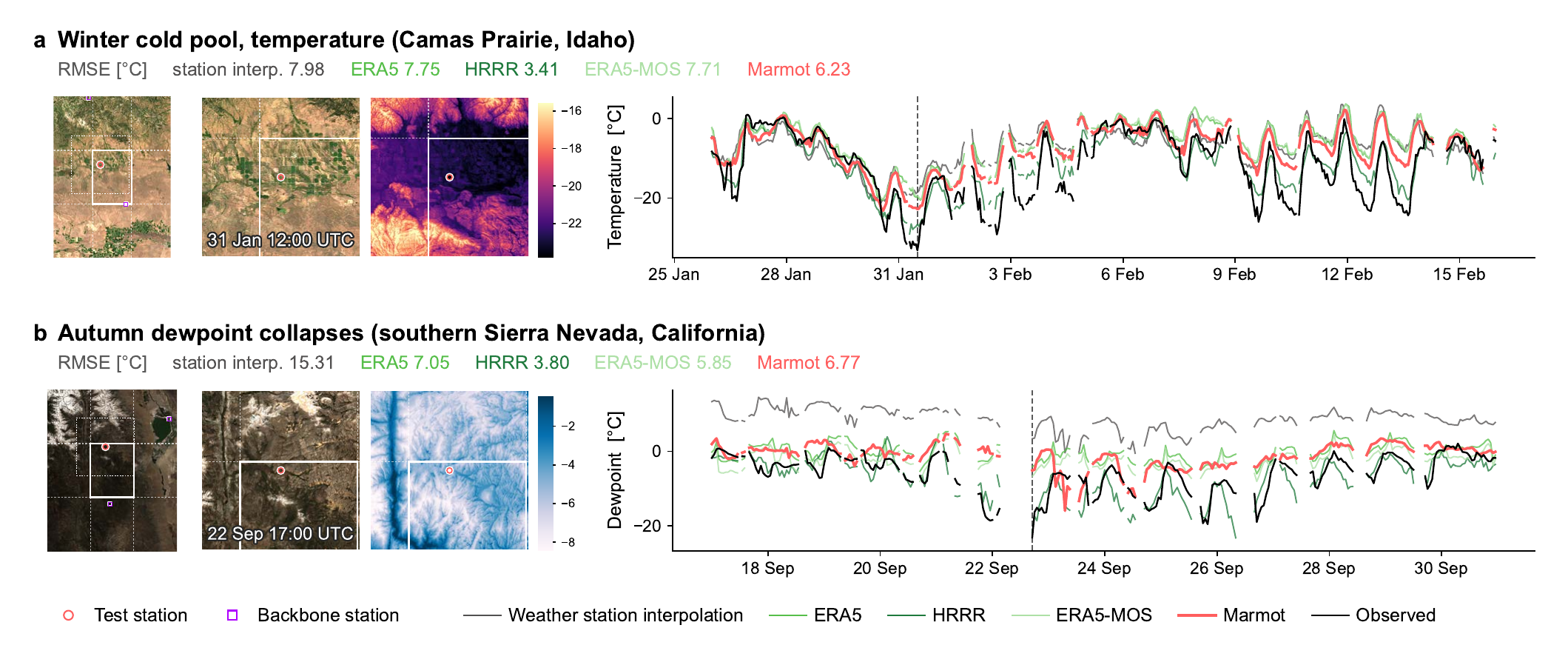}
\end{center}
\caption{
    \textbf{Examples of Marmot failure in remote mountain valleys.} \textbf{(a)} Temperature during a winter cold-pool event in Camas Prairie, Idaho. \textbf{(b)} Dewpoint during repeated drying events in the southern Sierra Nevada, California. For each example, maps show the broader station context, satellite imagery, and Marmot prediction at the indicated time. Time series compare the held-out station observation with Marmot and baseline methods. RMSE is reported above each panel. In both examples, the held-out station is distant from the backbone network and experiences large declines that Marmot does not fully recover. HRRR has the lowest error.
}
\label{fig:results:casestudy:failureNearestStationFar}
\end{figure}

One recurring failure occurs at remote locations that are far from backbone stations and experience large departures from surrounding conditions. In the Camas Prairie, Idaho, the nearest backbone station is 24 km from the held-out station. During a winter cold-pool event, Marmot substantially underestimates the magnitude of local cooling, with an RMSE of \(6.23^\circ\)C compared with \(3.41^\circ\)C for HRRR (Fig. \ref{fig:results:casestudy:failureNearestStationFar}); dewpoint at the same station exhibits a similar pattern. At a high-elevation site in the southern Sierra Nevada, California (2{,}743 m), Marmot likewise misses repeated declines in dewpoint that are captured more accurately by HRRR (Marmot RMSE \(6.77^\circ\)C versus HRRR \(3.80^\circ\)C). These examples could reflect limited information from distant backbone observations, limited training stations in comparable settings, difficulty in recovering locally generated atmospheric processes, or combinations of these.

\begin{figure}[tbp]
\begin{center}
\includegraphics[width=1\textwidth]{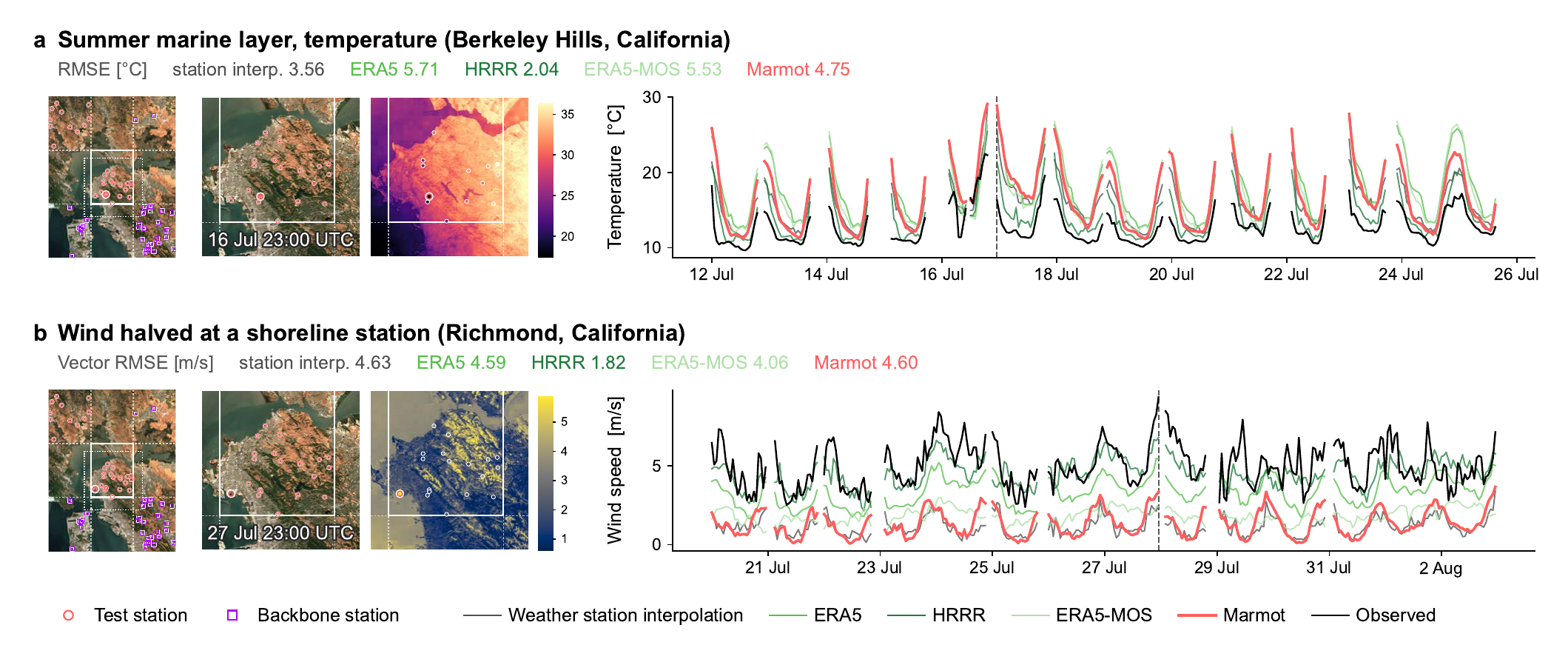}
\end{center}
\caption{
    \textbf{Examples of Marmot failure in East Bay, California.} 
    \textbf{(a)} Temperature at a held-out station in the Berkeley Hills, California, during summer marine-layer conditions. 
    \textbf{(b)} Wind speed at a held-out station on the Richmond shoreline, California. For each example, maps show the broader station context, satellite imagery, and Marmot prediction at the indicated time. Time series compare the held-out station observation with Marmot and baseline methods. RMSE is reported above each panel. In both cases, nearby backbone stations lie landward of the target, and Marmot does not recover the cooler temperature or stronger winds observed at the nearshore stations. HRRR has the lowest error.
}
\label{fig:results:casestudy:failureCostalOneSided}
\end{figure}

A second failure mode occurs at some nearshore locations in the East Bay of California, where weather transitions sharply between marine and inland conditions and neither nearby backbone stations nor ERA5 estimate the weather well (Fig. \ref{fig:results:casestudy:failureCostalOneSided}). At a station in the Berkeley Hills, California, where the neighboring backbone stations are landward, Marmot overestimates temperature with an RMSE of \(4.75^\circ\)C, while HRRR achieves an RMSE of \(2.04^\circ\)C. At a nearby station on the Richmond shoreline, Marmot also underestimates wind speed, while HRRR is more accurate (vector RMSE \(4.60\) versus  \(1.82\) m s\(^{-1}\)). We hypothesize that the asymmetric placement of stations relative to the bay (since this work does not include measurements over bodies of water), steep gradients in weather in the East Bay, and/or the poor baseline performance of nearby stations and ERA5 may make these conditions difficult to infer.

\subsection{Information Source Ablation Results}
\label{sec:supp:ablation}

The additive ablation in the main text evaluates how performance changes as information sources are progressively introduced to a station-only model (Fig. \ref{fig:results:overview}c; Section~\ref{sec:methods:ablation}). We additionally perform subtractive ablations from the full Marmot model, removing individual information sources while holding the remaining model components fixed (Table \ref{tab:Appendix:Results:AblationComponents}). The results are broadly consistent: station observations provide the foundation for prediction, while elevation, Earth observation, ERA5, and temporal information each contribute additional information, with ERA5 and the temporal encoding providing the smallest marginal improvements. The relative contributions of elevation and Earth observation depend on the other information available to the model. Adding explicit elevation differences to the station-only model produces a large improvement, whereas in the full model, removing AlphaEarth embeddings has a slightly larger effect than removing elevation differences. This difference likely reflects overlapping information about topography contained in the two representations.

\subsection{Sensitivity Analysis Results}
\label{sec:supp:sensitivity}

Marmot is generally robust to the hyperparameter choices examined (Table \ref{tab:Appendix:Results:Sensitivity}; Section \ref{sec:methods:sensitivity}). Performance improves as the number of neighboring stations increases, with diminishing returns or slightly worse performance beyond 64 neighbors. Performance is similarly stable across neighbor-dropout rates, especially between 10\% and 50\%, with no single setting consistently optimal across temperature, dewpoint, and wind. Varying the ERA5 spatial context from a single grid cell to a \(9\times9\) neighborhood also produces only small differences, with stencil sizes between $3\times 3$ and $7\times 7$ performing slightly better overall. The best AlphaEarth resolution --- 10, 30, or 100 m --- depends on the weather variable and metric, but differences between the three resolutions are within 1\%. These results indicate that performance is not strongly dependent on the particular hyperparameter configuration used for the primary Marmot specification.

Beyond model hyperparameters, we also examine sensitivity to the spatial partitioning used for evaluation. As expected, Marmot performs better under the random station split, where training and backbone stations can occur in the same ERA5 grid cells as held-out stations: RMSE decreases by 6\% for temperature, 6\% for dewpoint, and 2\% for wind relative to the by-cell split (Table \ref{tab:Appendix:Results:Partitions}). The random split provides a complementary estimate of performance when nearby station information is not deliberately excluded at test locations, whereas our primary grid-cell split tests the setting of spatial extrapolation beyond the backbone network.

\clearpage

\begin{figure}[bp]
\begin{center}
\includegraphics[width=1\textwidth]{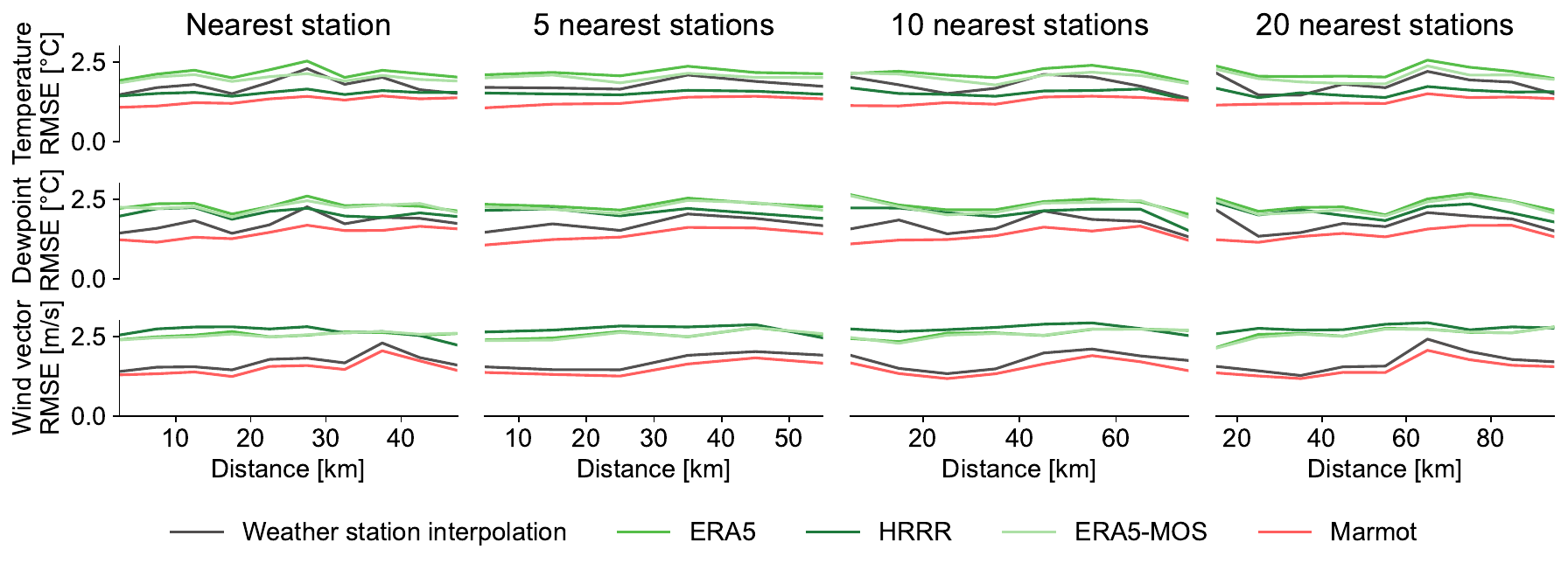}
\end{center}
\caption{
    \textbf{Prediction error as a function of distance to the nearest $k$ backbone stations,} for $k=1, 5, 10$, and $20$. For $k>1$ the distance is the mean over the $k$ nearest backbone stations. Across the 555 test stations, these distances span 2.1--71.6\,km ($k=1$; median 15.7\,km), 4.0--101.3\,km ($k=5$; median 23.2\,km), 6.5--158.9\,km ($k=10$; median 28.7\,km), and 10.0--228.2\,km ($k=20$; median 39.6\,km). Distance bins are 5\,km wide from 0--50\,km for $k=1$; 10\,km wide from 0--60\,km for $k=5$; 10\,km wide from 0--80\,km for $k=10$; and 10\,km wide from 10--100\,km for $k=20$.
}
\label{fig:results:error_distance}
\end{figure}
\begin{figure}[htbp]
\begin{center}
\includegraphics[width=1\textwidth]{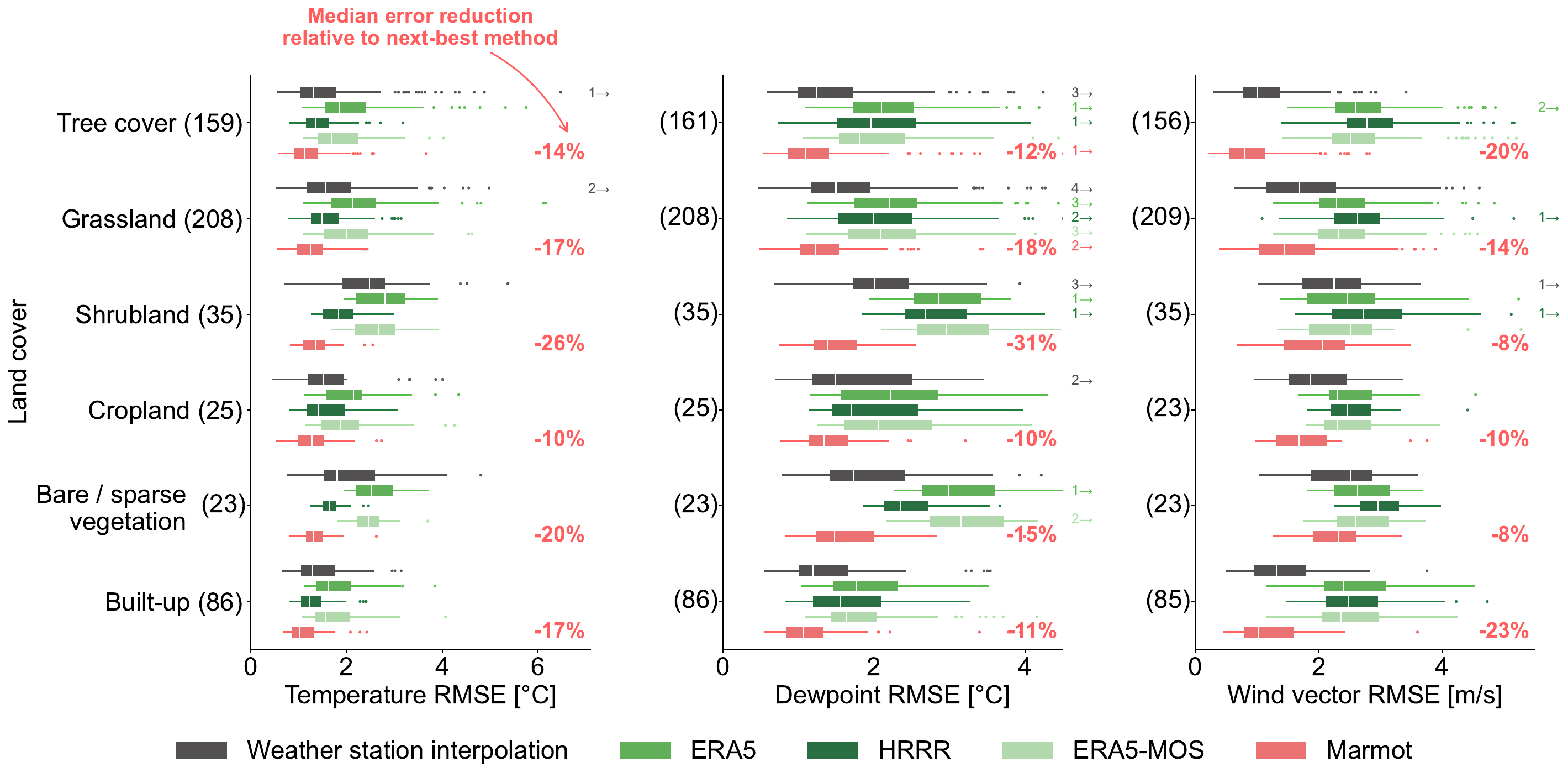}
\end{center}
\caption{
    \textbf{Distribution of prediction error by land cover type.} Land cover labels are derived from the ESA WorldCover 10 m v200, and we retain the 6 land cover categories with sufficient stations to summarize. Percentage labels show the median error reduction by Marmot relative to the next-best baseline method for each land cover type and weather variable.
}
\label{fig:appendix:results:land_cover_error}
\end{figure}
\begin{figure}[tbp]
\begin{center}
\includegraphics[width=1\textwidth]{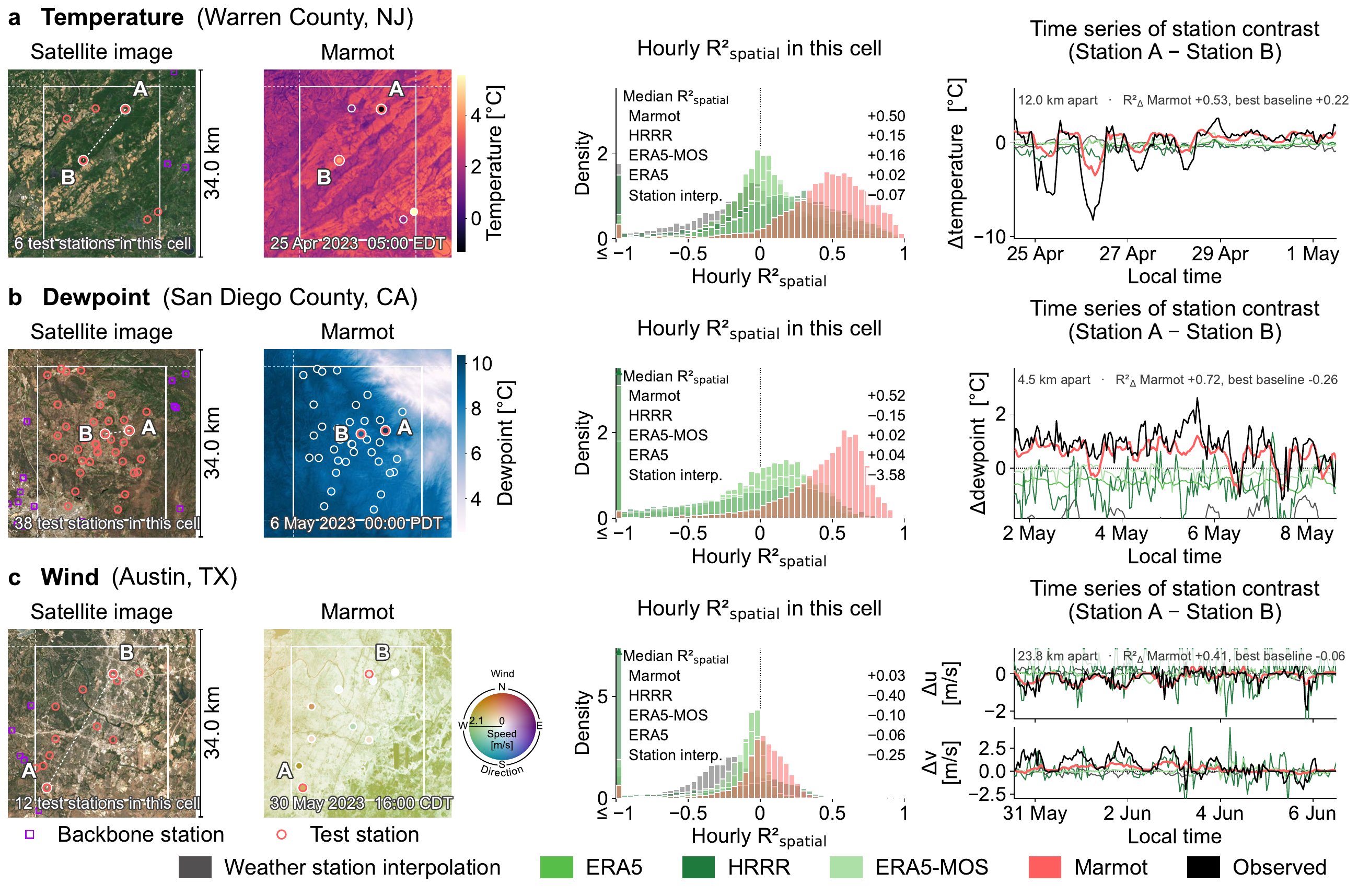}
\end{center}
\caption{
    \textbf{Case studies of sub-grid spatial variance recovery.}
    Examples are shown for \textbf{(a)} air temperature in Warren County, New Jersey, \textbf{(b)} dewpoint in San Diego County, California, and \textbf{(c)} wind in Austin, Texas. For each case study, panels show satellite imagery and the inferred micro-weather field within a single ERA5 grid cell, the distribution of hourly $R^{2}_{\text{spatial}}$ within the grid cell across 2023, and a time series of the observed and predicted weather contrast between two selected held-out stations (A and B). 
    Station contrasts are calculated as the difference in weather conditions between stations A and B and evaluate each method's ability to reproduce time-varying sub-grid spatial differences.
}
\label{fig:results:subgrid_cases}
\end{figure}
\begin{figure}[tbp]
\begin{center}
\includegraphics[width=1\textwidth]{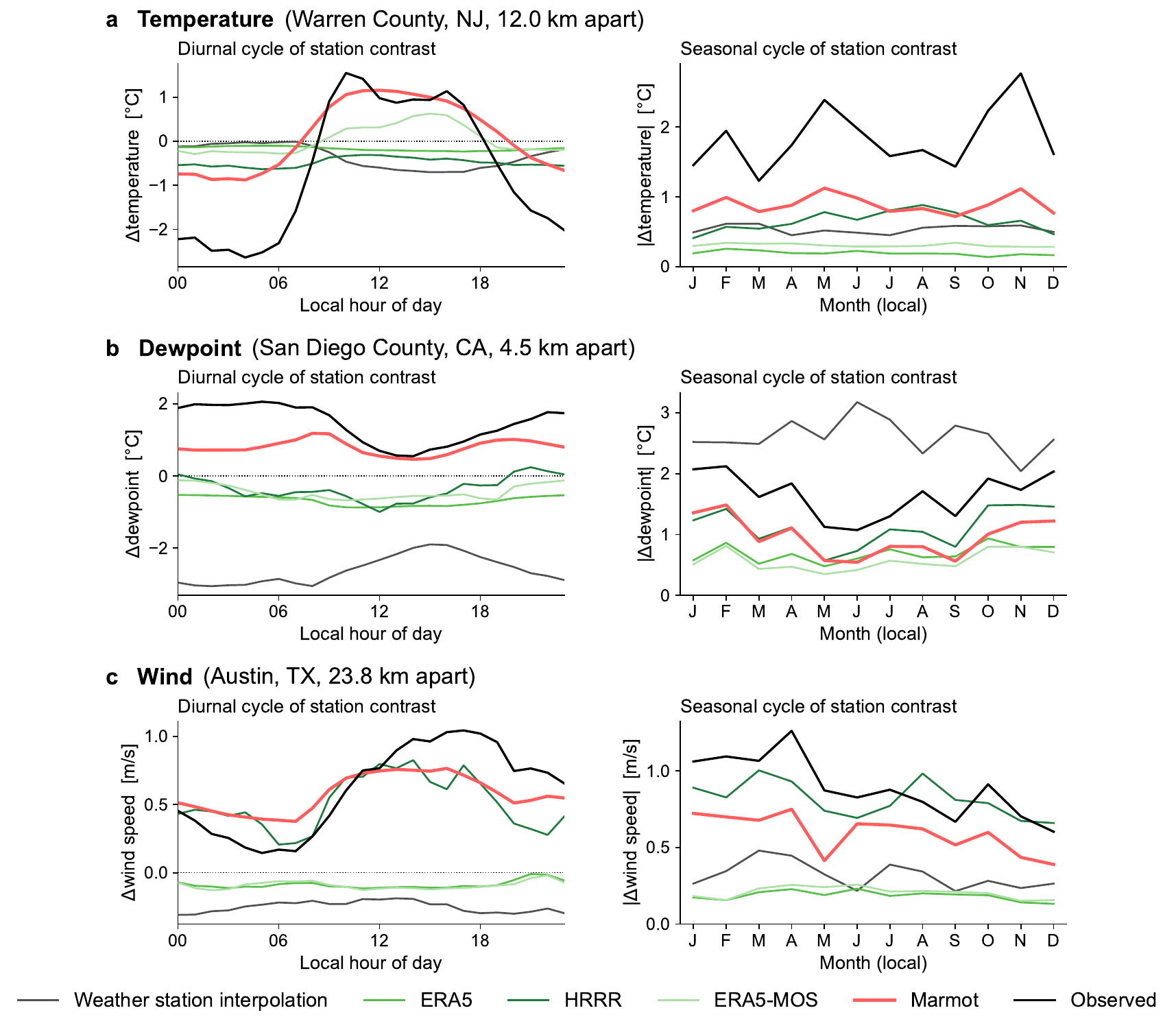}
\end{center}
\caption{
    \textbf{Diurnal and seasonal patterns in local weather contrasts.}
    Observed and modeled contrasts are aggregated by local hour of day (left) and local month (right) for the three station pairs shown in Figure~\ref{fig:results:subgrid_cases}: \textbf{(a)} temperature in Warren County, New Jersey; \textbf{(b)} dewpoint in San Diego County, California; and \textbf{(c)} wind speed in Austin, Texas. Diurnal panels show signed station contrasts, whereas seasonal panels show absolute contrasts.
}
\label{fig:results:subgrid_case_cycles}
\end{figure}

\begin{figure}[htbp]
\begin{center}
\includegraphics[width=1\textwidth]{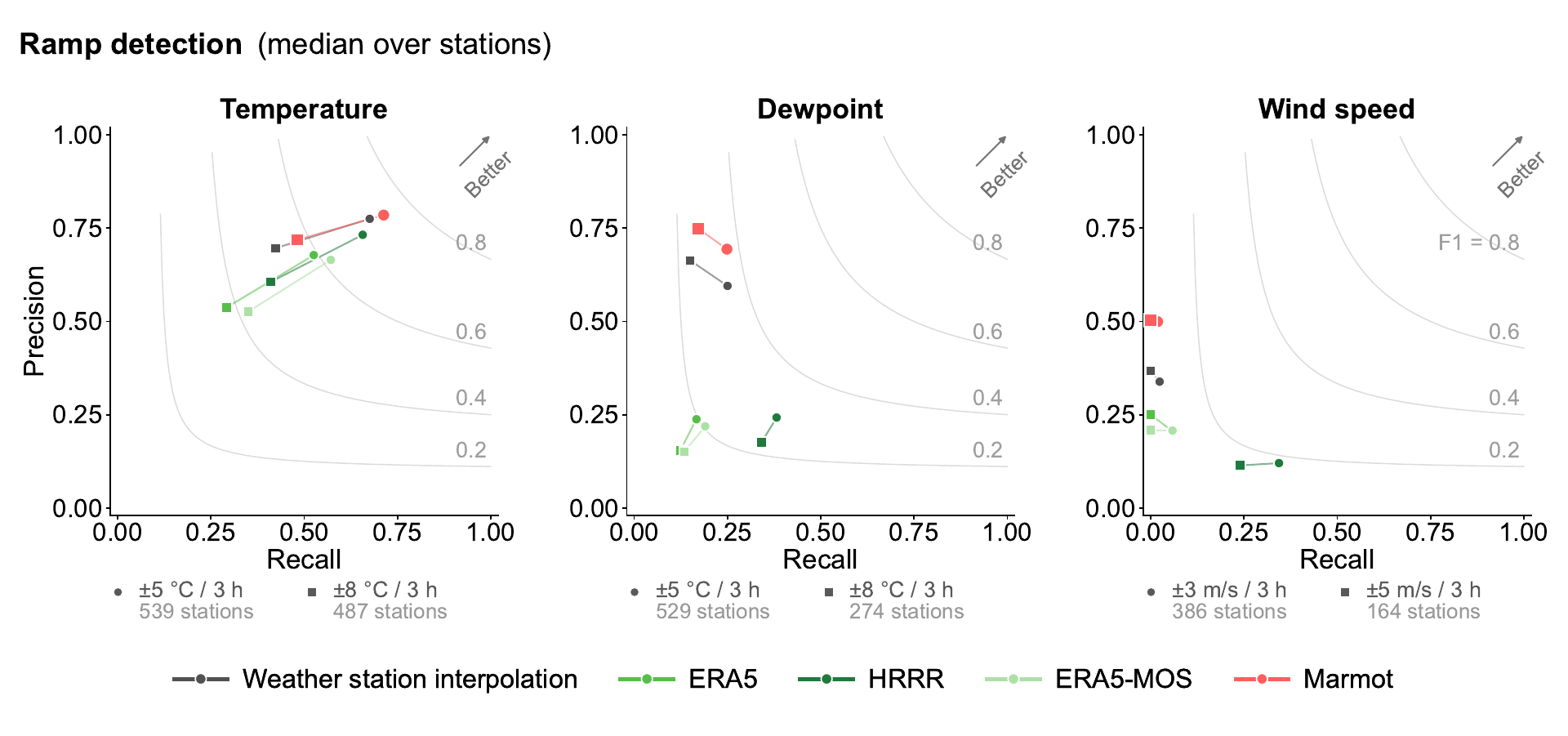}
\end{center}
\caption{
    \textbf{Detection of ramps (changes) in weather.}
    Precision and recall for detection of changes over 3 hour intervals. Points denote different magnitude thresholds; gray curves indicate lines of constant F1 score.
}
\label{fig:appendix:results:ramps}
\end{figure}


\clearpage

\begin{table} 
\centering
\caption{\textbf{Model errors and spatial R$^2$ at held-out test stations.} All methods are scored on the 555 test stations in 2023, a year absent from training, so the evaluation is held out in space and in time. \textbf{(a)} RMSE and MAE of temperature, dewpoint, and the wind vector. \textbf{(b)} Wind error broken out into speed and direction. \textbf{(c)} Spatial variance explained is calculated for each time step over all test stations and then averaged over all time steps. We report one standard deviation over the 2{,}000 replicates of a nested bootstrap. Bold and underline mark the best and second-best point estimate in each column; overlapping intervals do not resolve an ordering.}

\label{tab:Appendix:Results:SR2}
\setlength{\tabcolsep}{2pt}
\setlength{\aboverulesep}{0.25ex}
\setlength{\belowrulesep}{0.35ex}
\footnotesize
\providecommand{\uncsize}{\fontsize{6}{7}\selectfont}
\providecommand{\pmcell}[2]{}
\providecommand{\bpmcell}[2]{}
\providecommand{\upmcell}[2]{}
\renewcommand{\pmcell}[2]{$#1$\,{\uncsize$\pm#2$}}
\renewcommand{\bpmcell}[2]{$\mathbf{#1}$\,{\uncsize$\mathbf{\pm#2}$}}
\renewcommand{\upmcell}[2]{$\underline{#1}$\,{\uncsize$\pm#2$}}
\begin{tabular}{>{\raggedright\arraybackslash}m{3.0cm}cccccc}
\toprule
\multicolumn{7}{l}{\textbf{(a)} Error at held-out test stations}\\
\midrule
&\multicolumn{2}{c}{Temperature [\textdegree C]}&\multicolumn{2}{c}{Dewpoint [\textdegree C]}&\multicolumn{2}{c}{Wind vector [m/s]}\\
Model & RMSE & MAE & RMSE & MAE & RMSE & MAE \\
\midrule
Station interpolation & \pmcell{2.006}{0.126} & \pmcell{1.378}{0.084} & \upmcell{2.008}{0.192} & \upmcell{1.295}{0.097} & \upmcell{1.720}{0.055} & \upmcell{1.263}{0.047} \\
ERA5 & \pmcell{2.277}{0.066} & \pmcell{1.680}{0.048} & \pmcell{2.419}{0.077} & \pmcell{1.749}{0.051} & \pmcell{2.613}{0.085} & \pmcell{2.232}{0.083} \\
HRRR & \upmcell{1.574}{0.040} & \upmcell{1.165}{0.029} & \pmcell{2.248}{0.088} & \pmcell{1.565}{0.057} & \pmcell{2.781}{0.056} & \pmcell{2.326}{0.052} \\
ERA5-MOS & \pmcell{2.093}{0.065} & \pmcell{1.535}{0.047} & \pmcell{2.329}{0.085} & \pmcell{1.655}{0.055} & \pmcell{2.570}{0.081} & \pmcell{2.195}{0.079} \\
Marmot & \bpmcell{1.295}{0.034} & \bpmcell{0.919}{0.022} & \bpmcell{1.455}{0.063} & \bpmcell{0.985}{0.032} & \bpmcell{1.539}{0.056} & \bpmcell{1.104}{0.046} \\
\midrule
\multicolumn{7}{l}{\textbf{(b)} Wind error at held-out test stations, broken out into speed and direction}\\
\midrule
&\multicolumn{3}{c}{Wind speed [m/s]}&\multicolumn{3}{c}{Wind direction [\textdegree]}\\
Model & \multicolumn{1}{c}{RMSE} & \multicolumn{2}{c}{MAE} & \multicolumn{1}{c}{RMSE} & \multicolumn{2}{c}{MAE} \\
\midrule
Station interpolation & \multicolumn{1}{c}{\upmcell{1.326}{0.041}} & \multicolumn{2}{c}{\upmcell{0.898}{0.030}} & \multicolumn{1}{c}{\pmcell{67.38}{1.40}} & \multicolumn{2}{c}{\pmcell{49.54}{1.36}} \\
ERA5 & \multicolumn{1}{c}{\pmcell{2.161}{0.095}} & \multicolumn{2}{c}{\pmcell{1.717}{0.093}} & \multicolumn{1}{c}{\pmcell{66.39}{1.34}} & \multicolumn{2}{c}{\pmcell{48.68}{1.28}} \\
HRRR & \multicolumn{1}{c}{\pmcell{2.319}{0.058}} & \multicolumn{2}{c}{\pmcell{1.837}{0.059}} & \multicolumn{1}{c}{\upmcell{63.46}{1.50}} & \multicolumn{2}{c}{\upmcell{45.53}{1.36}} \\
ERA5-MOS & \multicolumn{1}{c}{\pmcell{2.115}{0.090}} & \multicolumn{2}{c}{\pmcell{1.673}{0.087}} & \multicolumn{1}{c}{\pmcell{67.53}{1.29}} & \multicolumn{2}{c}{\pmcell{49.93}{1.25}} \\
Marmot & \multicolumn{1}{c}{\bpmcell{1.182}{0.046}} & \multicolumn{2}{c}{\bpmcell{0.780}{0.033}} & \multicolumn{1}{c}{\bpmcell{62.71}{1.24}} & \multicolumn{2}{c}{\bpmcell{45.22}{1.16}} \\
\midrule
\multicolumn{7}{l}{\textbf{(c)} Spatial $R^2$ at held-out test stations}\\
\midrule
Model &\multicolumn{2}{c}{Temperature}&\multicolumn{2}{c}{Dewpoint}&\multicolumn{2}{c}{Wind}\\
\midrule
Station interpolation & \multicolumn{2}{c}{\pmcell{0.883}{0.016}} & \multicolumn{2}{c}{\upmcell{0.903}{0.023}} & \multicolumn{2}{c}{\upmcell{0.334}{0.031}} \\
ERA5 & \multicolumn{2}{c}{\pmcell{0.863}{0.012}} & \multicolumn{2}{c}{\pmcell{0.874}{0.012}} & \multicolumn{2}{c}{\pmcell{-0.501}{0.140}} \\
HRRR & \multicolumn{2}{c}{\upmcell{0.929}{0.006}} & \multicolumn{2}{c}{\pmcell{0.895}{0.012}} & \multicolumn{2}{c}{\pmcell{-0.675}{0.135}} \\
ERA5-MOS & \multicolumn{2}{c}{\pmcell{0.877}{0.011}} & \multicolumn{2}{c}{\pmcell{0.873}{0.013}} & \multicolumn{2}{c}{\pmcell{-0.465}{0.134}} \\
Marmot & \multicolumn{2}{c}{\bpmcell{0.951}{0.004}} & \multicolumn{2}{c}{\bpmcell{0.949}{0.005}} & \multicolumn{2}{c}{\bpmcell{0.464}{0.021}} \\
\bottomrule
\end{tabular}
\end{table}
\begin{table}[htbp]
\centering
\caption{\textbf{Model and baseline errors under two station partitions.} The by-ERA5-cell split withholds whole grid cells, so held-out stations have no training or backbone station in their own cell; it measures spatial extrapolation and is the partition used throughout the main text. The random split withholds individual stations, so neighbors in the same cell may remain available. Because the two partitions are evaluated in different station-hours ($n$ given per group), rows may be compared within a group but not across groups. We report one standard deviation over the 2{,}000 replicates of a nested bootstrap. Bold marks the best point estimate in each column.}
\label{tab:Appendix:Results:Partitions}
\setlength{\tabcolsep}{2pt}
\setlength{\aboverulesep}{0.25ex}
\setlength{\belowrulesep}{0.35ex}
\footnotesize
\providecommand{\pmcell}[2]{}
\providecommand{\bpmcell}[2]{}
\renewcommand{\pmcell}[2]{$#1$\,{\scriptsize$\pm#2$}}
\renewcommand{\bpmcell}[2]{$\mathbf{#1}$\,{\scriptsize$\mathbf{\pm#2}$}}
\begin{tabular}{>{\raggedright\arraybackslash}m{3.4cm}*{6}{>{\centering\arraybackslash}m{\dimexpr(\textwidth-3.4cm-28pt)/6\relax}}}
\toprule
&\multicolumn{2}{c}{Temperature [\textdegree C]}&\multicolumn{2}{c}{Dewpoint [\textdegree C]}&\multicolumn{2}{c}{Wind vector [m/s]}\\
\cmidrule(lr){2-3}\cmidrule(lr){4-5}\cmidrule(lr){6-7}
Model & RMSE & MAE & RMSE & MAE & RMSE & MAE \\
\midrule
\multicolumn{7}{@{}l@{}}{\makebox[0pt][l]{\textit{By-ERA5-cell split} \;\footnotesize(used in paper; whole cells withheld together, no training station in target cell; $n=3{,}885{,}810$)}}\\
\midrule
\textbf{Marmot} & \bpmcell{1.295}{0.034} & \bpmcell{0.919}{0.022} & \bpmcell{1.455}{0.063} & \bpmcell{0.985}{0.032} & \bpmcell{1.539}{0.056} & \bpmcell{1.104}{0.046} \\
Station interpolation & \pmcell{2.006}{0.126} & \pmcell{1.378}{0.084} & \pmcell{2.008}{0.192} & \pmcell{1.295}{0.097} & \pmcell{1.720}{0.055} & \pmcell{1.263}{0.047} \\
HRRR & \pmcell{1.574}{0.040} & \pmcell{1.165}{0.029} & \pmcell{2.248}{0.088} & \pmcell{1.565}{0.057} & \pmcell{2.781}{0.056} & \pmcell{2.326}{0.052} \\
ERA5 & \pmcell{2.277}{0.066} & \pmcell{1.680}{0.048} & \pmcell{2.419}{0.077} & \pmcell{1.749}{0.051} & \pmcell{2.613}{0.085} & \pmcell{2.232}{0.083} \\
ERA5-MOS & \pmcell{2.093}{0.065} & \pmcell{1.535}{0.047} & \pmcell{2.329}{0.085} & \pmcell{1.655}{0.055} & \pmcell{2.570}{0.081} & \pmcell{2.195}{0.079} \\
\addlinespace[2pt]\midrule
\multicolumn{7}{@{}l@{}}{\makebox[0pt][l]{\textit{Random split} \;\footnotesize(stations drawn at random across the US; $n=3{,}928{,}815$)}}\\
\midrule
\textbf{Marmot} & \bpmcell{1.213}{0.027} & \bpmcell{0.847}{0.016} & \bpmcell{1.372}{0.033} & \bpmcell{0.950}{0.022} & \bpmcell{1.509}{0.044} & \bpmcell{1.066}{0.030} \\
Station interpolation & \pmcell{1.717}{0.063} & \pmcell{1.158}{0.035} & \pmcell{1.671}{0.045} & \pmcell{1.150}{0.030} & \pmcell{1.744}{0.058} & \pmcell{1.243}{0.035} \\
HRRR & \pmcell{1.652}{0.033} & \pmcell{1.200}{0.022} & \pmcell{2.367}{0.045} & \pmcell{1.658}{0.031} & \pmcell{2.889}{0.043} & \pmcell{2.390}{0.034} \\
ERA5 & \pmcell{2.438}{0.065} & \pmcell{1.765}{0.043} & \pmcell{2.581}{0.051} & \pmcell{1.868}{0.034} & \pmcell{2.684}{0.046} & \pmcell{2.273}{0.035} \\
ERA5-MOS & \pmcell{2.118}{0.040} & \pmcell{1.540}{0.028} & \pmcell{2.353}{0.040} & \pmcell{1.693}{0.027} & \pmcell{2.638}{0.046} & \pmcell{2.227}{0.035} \\
\bottomrule
\end{tabular}
\end{table}

\begin{table}[htbp]
\centering
\caption{\textbf{Component ablations of the micro-weather model.} \textbf{(a)} Additive ladder: the base row is a pure station-interpolation model; each row below adds one information source and inherits everything above it. The sources enter in the order: station elevation difference, two ERA5 features (innovation, then patch), Earth observation features (AlphaEarth), a temporal encoding, and a regularizer (neighbor dropout). \textbf{(b)} Leave-one-out: one component removed from the Marmot (full model), everything else kept, in that same order. Bold marks Marmot (full model), which is the last row of (a) and the first of (b). Both panels use the column header printed above (a). We report one standard deviation over the 2{,}000 replicates of a nested bootstrap.}
\label{tab:Appendix:Results:AblationComponents}
\setlength{\tabcolsep}{2pt}
\setlength{\aboverulesep}{0.25ex}
\setlength{\belowrulesep}{0.35ex}
\footnotesize
\providecommand{\pmcell}[2]{}
\providecommand{\bpmcell}[2]{}
\renewcommand{\pmcell}[2]{$#1$\,{\scriptsize$\pm#2$}}
\renewcommand{\bpmcell}[2]{$\mathbf{#1}$\,{\scriptsize$\mathbf{\pm#2}$}}
\begin{tabular}{>{\raggedright\arraybackslash}m{3.4cm}*{6}{>{\centering\arraybackslash}m{\dimexpr(\textwidth-3.4cm-28pt)/6\relax}}}
\toprule
&\multicolumn{2}{c}{Temperature [\textdegree C]}&\multicolumn{2}{c}{Dewpoint [\textdegree C]}&\multicolumn{2}{c}{Wind vector [m/s]}\\
\cmidrule(lr){2-3}\cmidrule(lr){4-5}\cmidrule(lr){6-7}
Model & RMSE & MAE & RMSE & MAE & RMSE & MAE \\
\midrule
\multicolumn{7}{@{}l@{}}{\makebox[0pt][l]{\textbf{(a)} Additive ladder: each row adds one information source on top of everything above}}\\
\midrule
Station interpolation & $2.006$ & $1.378$ & $2.008$ & $1.295$ & $1.720$ & $1.263$ \\
Station-only model & \pmcell{1.896}{0.091} & \pmcell{1.320}{0.058} & \pmcell{1.778}{0.092} & \pmcell{1.216}{0.057} & \pmcell{1.651}{0.058} & \pmcell{1.204}{0.048} \\
\addlinespace[2pt]
$+$ elevation difference & \pmcell{1.433}{0.040} & \pmcell{1.007}{0.025} & \pmcell{1.515}{0.064} & \pmcell{1.033}{0.033} & \pmcell{1.631}{0.058} & \pmcell{1.185}{0.047} \\
\addlinespace[2pt]
$+$ ERA5 innovation & \pmcell{1.428}{0.038} & \pmcell{1.002}{0.024} & \pmcell{1.507}{0.063} & \pmcell{1.028}{0.032} & \pmcell{1.603}{0.056} & \pmcell{1.175}{0.045} \\
$+$ ERA5 patch & \pmcell{1.426}{0.037} & \pmcell{1.002}{0.024} & \pmcell{1.503}{0.063} & \pmcell{1.027}{0.033} & \pmcell{1.598}{0.055} & \pmcell{1.177}{0.045} \\
\addlinespace[2pt]
$+$ AlphaEarth & \pmcell{1.337}{0.034} & \pmcell{0.946}{0.023} & \pmcell{1.479}{0.064} & \pmcell{1.000}{0.033} & \pmcell{1.545}{0.055} & \pmcell{1.110}{0.046} \\
\addlinespace[2pt]
$+$ temporal encoding & \pmcell{1.308}{0.034} & \pmcell{0.926}{0.022} & \pmcell{1.464}{0.062} & \pmcell{0.995}{0.031} & \pmcell{1.541}{0.055} & \pmcell{1.106}{0.046} \\
\addlinespace[2pt]
\textbf{$+$ neighbor dropout (Marmot)} & \bpmcell{1.295}{0.034} & \bpmcell{0.919}{0.022} & \bpmcell{1.455}{0.063} & \bpmcell{0.985}{0.032} & \bpmcell{1.539}{0.056} & \bpmcell{1.104}{0.046} \\
\midrule
\multicolumn{7}{@{}l@{}}{\makebox[0pt][l]{\textbf{(b)} Leave-one-out: one component removed from the Marmot (full model)}}\\
\midrule
\textbf{Marmot} & \bpmcell{1.295}{0.034} & \bpmcell{0.919}{0.022} & \bpmcell{1.455}{0.063} & \bpmcell{0.985}{0.032} & \bpmcell{1.539}{0.056} & \bpmcell{1.104}{0.046} \\
\addlinespace[2pt]
$-$ station observations & \pmcell{1.682}{0.036} & \pmcell{1.244}{0.026} & \pmcell{1.904}{0.066} & \pmcell{1.350}{0.043} & \pmcell{1.626}{0.064} & \pmcell{1.172}{0.052} \\
\addlinespace[2pt]
$-$ elevation difference & \pmcell{1.396}{0.034} & \pmcell{1.012}{0.025} & \pmcell{1.511}{0.063} & \pmcell{1.030}{0.031} & \pmcell{1.549}{0.057} & \pmcell{1.115}{0.047} \\
\addlinespace[2pt]
$-$ ERA5 innovation & \pmcell{1.312}{0.035} & \pmcell{0.931}{0.023} & \pmcell{1.468}{0.064} & \pmcell{0.992}{0.032} & \pmcell{1.552}{0.056} & \pmcell{1.110}{0.046} \\
$-$ ERA5 patch & \pmcell{1.295}{0.034} & \pmcell{0.916}{0.022} & \pmcell{1.461}{0.063} & \pmcell{0.990}{0.031} & \pmcell{1.540}{0.055} & \pmcell{1.102}{0.046} \\
\addlinespace[2pt]
$-$ AlphaEarth & \pmcell{1.410}{0.037} & \pmcell{0.993}{0.024} & \pmcell{1.492}{0.064} & \pmcell{1.013}{0.032} & \pmcell{1.593}{0.055} & \pmcell{1.172}{0.045} \\
\addlinespace[2pt]
$-$ temporal encoding & \pmcell{1.318}{0.034} & \pmcell{0.934}{0.022} & \pmcell{1.468}{0.064} & \pmcell{0.994}{0.032} & \pmcell{1.546}{0.056} & \pmcell{1.108}{0.047} \\
\addlinespace[2pt]
$-$ neighbor dropout & \pmcell{1.308}{0.034} & \pmcell{0.926}{0.022} & \pmcell{1.464}{0.062} & \pmcell{0.995}{0.031} & \pmcell{1.541}{0.055} & \pmcell{1.106}{0.046} \\
\bottomrule
\end{tabular}
\end{table}
\begin{table}[htbp]
\centering
\caption{\textbf{Ablation of Earth observation information sources.} The first row begins with the model with no Earth observation information and the last row, marked in bold, is the Marmot model used throughout the paper. The rows are grouped into single hand-built sources, combinations of them, and the learned embedding. ``Cols'' is the number of feature columns each input contributes. We report one standard deviation over the 2,000 replicates of a nested bootstrap.}
\label{tab:Appendix:Results:SurfaceInfo}
\setlength{\tabcolsep}{2pt}
\setlength{\aboverulesep}{0.25ex}
\setlength{\belowrulesep}{0.35ex}
\footnotesize
\providecommand{\uncsize}{\fontsize{6}{7}\selectfont}
\providecommand{\pmcell}[2]{}
\providecommand{\bpmcell}[2]{}
\renewcommand{\pmcell}[2]{$#1$\,{\uncsize$\pm#2$}}
\renewcommand{\bpmcell}[2]{$\mathbf{#1}$\,{\uncsize$\mathbf{\pm#2}$}}
\begin{tabular}{>{\raggedright\arraybackslash}m{3.4cm}>{\centering\arraybackslash}m{0.7cm}*{6}{>{\centering\arraybackslash}m{\dimexpr(\textwidth-3.2cm-0.7cm-32pt)/6\relax}}}
\toprule
&&\multicolumn{2}{c}{Temperature [\textdegree C]}&\multicolumn{2}{c}{Dewpoint [\textdegree C]}&\multicolumn{2}{c}{Wind vector [m/s]}\\
\cmidrule(lr){3-4}\cmidrule(lr){5-6}\cmidrule(lr){7-8}
Surface information & Cols & RMSE & MAE & RMSE & MAE & RMSE & MAE \\
\midrule
No surface information & 0 & \pmcell{1.410}{0.037} & \pmcell{0.993}{0.024} & \pmcell{1.492}{0.064} & \pmcell{1.013}{0.032} & \pmcell{1.593}{0.055} & \pmcell{1.172}{0.045} \\
\midrule
\multicolumn{8}{l}{\textit{Single source}}\\
\quad Terrain & 5 & \pmcell{1.400}{0.037} & \pmcell{0.990}{0.025} & \pmcell{1.492}{0.064} & \pmcell{1.014}{0.033} & \pmcell{1.588}{0.056} & \pmcell{1.161}{0.045} \\
\quad Land cover & 7 & \pmcell{1.390}{0.038} & \pmcell{0.984}{0.025} & \pmcell{1.490}{0.063} & \pmcell{1.011}{0.033} & \pmcell{1.581}{0.057} & \pmcell{1.147}{0.047} \\
\quad Sentinel-2 & 48 & \pmcell{1.368}{0.035} & \pmcell{0.965}{0.023} & \pmcell{1.482}{0.063} & \pmcell{1.012}{0.032} & \pmcell{1.576}{0.054} & \pmcell{1.140}{0.046} \\
\midrule
\multicolumn{8}{l}{\textit{Combined hand-built sources}}\\
\quad Land cover + terrain & 12 & \pmcell{1.390}{0.039} & \pmcell{0.981}{0.025} & \pmcell{1.486}{0.063} & \pmcell{1.008}{0.032} & \pmcell{1.578}{0.057} & \pmcell{1.143}{0.046} \\
\quad All three & 60 & \pmcell{1.371}{0.036} & \pmcell{0.971}{0.024} & \pmcell{1.480}{0.063} & \pmcell{1.007}{0.032} & \pmcell{1.580}{0.056} & \pmcell{1.143}{0.046} \\
\midrule
\multicolumn{8}{l}{\textit{Learned embedding}}\\
\quad \textbf{AlphaEarth 30\,m (Marmot)} & \textbf{64} & \bpmcell{1.295}{0.034} & \bpmcell{0.919}{0.022} & \bpmcell{1.455}{0.063} & \bpmcell{0.985}{0.032} & \bpmcell{1.539}{0.056} & \bpmcell{1.104}{0.046} \\
\bottomrule
\end{tabular}
\end{table}
\begin{table}[htbp]
\centering
\caption{\textbf{Sensitivity to model design choices.} Each row is the Marmot (full model) with one setting changed; the row marked ``Marmot'' is the configuration used throughout the paper and repeats identically in every panel. \textbf{(a)} Number of backbone neighbors $k$ attended to. \textbf{(b)} Fraction of neighbors randomly dropped during training. \textbf{(c)} Size of the ERA5 stencil centered on the target. \textbf{(d)} Resolution of the AlphaEarth embedding. Bold marks the best point estimate in each column within a panel. We report one standard deviation over 2{,}000 replicates of a nested bootstrap, scored on the 854 validation stations over 2020--2022 rather than the test partition.}
\label{tab:Appendix:Results:Sensitivity}
\setlength{\tabcolsep}{2pt}
\setlength{\aboverulesep}{0.25ex}
\setlength{\belowrulesep}{0.35ex}
\footnotesize
\providecommand{\pmcell}[2]{}
\providecommand{\bpmcell}[2]{}
\renewcommand{\pmcell}[2]{$#1$\,{\scriptsize$\pm#2$}}
\renewcommand{\bpmcell}[2]{$\mathbf{#1}$\,{\scriptsize$\mathbf{\pm#2}$}}
\begin{tabular}{>{\raggedright\arraybackslash}m{3.4cm}*{6}{>{\centering\arraybackslash}m{\dimexpr(\textwidth-3.4cm-28pt)/6\relax}}}
\toprule
\multicolumn{7}{@{}l@{}}{\makebox[0pt][l]{\textbf{(a)} Neighbor count $k$, with and without the embedding}}\\
\midrule
&\multicolumn{2}{c}{Temperature [\textdegree C]}&\multicolumn{2}{c}{Dewpoint [\textdegree C]}&\multicolumn{2}{c}{Wind vector [m/s]}\\
\cmidrule(lr){2-3}\cmidrule(lr){4-5}\cmidrule(lr){6-7}
Setting & RMSE & MAE & RMSE & MAE & RMSE & MAE \\
\midrule
\quad $k=16$ & \pmcell{1.402}{0.033} & \pmcell{0.998}{0.025} & \pmcell{1.577}{0.039} & \pmcell{1.104}{0.026} & \pmcell{1.663}{0.049} & \pmcell{1.188}{0.037} \\
\quad $k=32$ & \pmcell{1.388}{0.034} & \pmcell{0.987}{0.025} & \pmcell{1.559}{0.037} & \pmcell{1.094}{0.024} & \pmcell{1.661}{0.049} & \pmcell{1.185}{0.037} \\
\quad $k=64$ (Marmot) & \pmcell{1.378}{0.034} & \bpmcell{0.981}{0.025} & \bpmcell{1.547}{0.036} & \bpmcell{1.085}{0.023} & \pmcell{1.658}{0.049} & \bpmcell{1.183}{0.038} \\
\quad $k=128$ & \bpmcell{1.376}{0.033} & \bpmcell{0.981}{0.025} & \pmcell{1.549}{0.036} & \pmcell{1.088}{0.024} & \bpmcell{1.652}{0.048} & \pmcell{1.185}{0.038} \\
\midrule
\multicolumn{7}{@{}l@{}}{\makebox[0pt][l]{\textbf{(b)} Neighbors dropped at training time}}\\
\midrule
\quad 0\% & \pmcell{1.389}{0.035} & \pmcell{0.986}{0.025} & \pmcell{1.554}{0.036} & \pmcell{1.092}{0.024} & \pmcell{1.666}{0.049} & \pmcell{1.190}{0.038} \\
\quad 10\% & \pmcell{1.380}{0.034} & \bpmcell{0.980}{0.025} & \pmcell{1.551}{0.036} & \pmcell{1.087}{0.024} & \pmcell{1.661}{0.049} & \pmcell{1.188}{0.038} \\
\quad 25\% (Marmot) & \bpmcell{1.378}{0.034} & \pmcell{0.981}{0.025} & \bpmcell{1.547}{0.036} & \bpmcell{1.085}{0.023} & \pmcell{1.658}{0.049} & \pmcell{1.183}{0.038} \\
\quad 50\% & \bpmcell{1.378}{0.034} & \pmcell{0.983}{0.025} & \pmcell{1.548}{0.035} & \pmcell{1.086}{0.023} & \bpmcell{1.657}{0.049} & \bpmcell{1.180}{0.037} \\
\quad 60\% & \pmcell{1.381}{0.034} & \pmcell{0.985}{0.025} & \pmcell{1.551}{0.035} & \pmcell{1.089}{0.023} & \pmcell{1.659}{0.048} & \pmcell{1.181}{0.037} \\
\quad 75\% & \pmcell{1.387}{0.033} & \pmcell{0.992}{0.024} & \pmcell{1.563}{0.035} & \pmcell{1.101}{0.024} & \pmcell{1.664}{0.049} & \pmcell{1.187}{0.038} \\
\midrule
\multicolumn{7}{@{}l@{}}{\makebox[0pt][l]{\textbf{(c)} ERA5 stencil size}}\\
\midrule
\quad 1$\times$1 & \pmcell{1.389}{0.035} & \pmcell{0.986}{0.026} & \pmcell{1.554}{0.036} & \pmcell{1.092}{0.024} & \pmcell{1.663}{0.049} & \pmcell{1.185}{0.038} \\
\quad 3$\times$3 & \bpmcell{1.371}{0.033} & \bpmcell{0.974}{0.025} & \pmcell{1.549}{0.037} & \pmcell{1.088}{0.024} & \pmcell{1.658}{0.049} & \pmcell{1.184}{0.038} \\
\quad 5$\times$5 (Marmot) & \pmcell{1.378}{0.034} & \pmcell{0.981}{0.025} & \bpmcell{1.547}{0.036} & \bpmcell{1.085}{0.023} & \pmcell{1.658}{0.049} & \bpmcell{1.183}{0.038} \\
\quad 7$\times$7 & \pmcell{1.385}{0.033} & \pmcell{0.987}{0.024} & \pmcell{1.560}{0.036} & \pmcell{1.097}{0.025} & \bpmcell{1.657}{0.049} & \pmcell{1.187}{0.038} \\
\quad 9$\times$9 & \pmcell{1.380}{0.033} & \pmcell{0.981}{0.025} & \pmcell{1.551}{0.036} & \pmcell{1.090}{0.024} & \pmcell{1.660}{0.049} & \pmcell{1.185}{0.038} \\
\midrule
\multicolumn{7}{@{}l@{}}{\makebox[0pt][l]{\textbf{(d)} AlphaEarth resolution}}\\
\midrule
\quad 10\,m & \pmcell{1.382}{0.034} & \pmcell{0.982}{0.025} & \pmcell{1.550}{0.036} & \pmcell{1.088}{0.024} & \bpmcell{1.657}{0.048} & \pmcell{1.185}{0.037} \\
\quad 30\,m (Marmot) & \pmcell{1.378}{0.034} & \pmcell{0.981}{0.025} & \bpmcell{1.547}{0.036} & \bpmcell{1.085}{0.023} & \pmcell{1.658}{0.049} & \bpmcell{1.183}{0.038} \\
\quad 100\,m & \bpmcell{1.369}{0.033} & \bpmcell{0.974}{0.024} & \pmcell{1.549}{0.036} & \pmcell{1.089}{0.023} & \pmcell{1.663}{0.049} & \pmcell{1.189}{0.037} \\
\bottomrule
\end{tabular}
\end{table}

\begin{table}[htbp]
\centering
\caption{\textbf{Surface correlates of inferred micro-weather structure.} Fraction of spatial variance ($R^2$) in the inferred micro-weather field of each case study in \Cref{fig:Results:Inference:Phenomena} explained by terrain elevation, land cover, and the two together.}
\label{tab:Appendix:Results:SurfaceCorrelates}
\footnotesize
\begin{tabular}{clccc}
\toprule
Panel & Case & Terrain $R^2$ & Land cover $R^2$ & Together $R^2$ \\
\midrule
a & Diablo Range, California      & \textbf{0.44} & 0.20          & 0.49 \\
b & Great Salt Lake Desert, Utah  & 0.05          & \textbf{0.42} & 0.44 \\
c & Finger Lakes, New York        & \textbf{0.91} & 0.02          & 0.92 \\
d & Antelope Valley, California   & 0.09          & \textbf{0.34} & 0.40 \\
e & Sonoma Valley, California     & \textbf{0.37} & 0.08          & 0.49 \\
f & Hunterdon County, New Jersey  & 0.19          & \textbf{0.56} & 0.59 \\
\bottomrule
\end{tabular}

\end{table}

\end{document}